\PassOptionsToPackage{usenames,dvipsnames,svgnames, table}{xcolor} 
\documentclass[acmsmall]{acmart}

\usepackage{soul}
\usepackage{mdframed}
\usepackage{marginnote}
\usepackage{ifthen}
\usepackage{tabularx} % Better tables
\usepackage{pifont}
\usepackage{xspace}
\usepackage{annotate-equations}
\usepackage{lscape}
\usepackage{graphicx}
\usepackage{subcaption}
\usepackage{caption}
\setcopyright{acmlicensed}
\usepackage{titlesec}
\titlespacing*{\paragraph}{0em}{0em}{0.3333333333em}
\titleformat{\paragraph}[runin]{\normalsize\bfseries}{}{0em}{}[\hspace{0.3333333333em}---]

\renewcommand{\paragraph}[1]{\refstepcounter{paragraph}\noindent\textbf{#1\ ---}\label{par:\theparagraph}}

\usepackage{tikz}
\usepackage[most]{tcolorbox}
\definecolor{lgray}{rgb}{0.8, 0.8, 0.85}
\newcommand*\annotatedFigureBoxCustom[8]{\draw[lgray,thick,rounded corners] (#1) rectangle (#2);\node at (#4) [fill=lgray,text=black, thin,shape=rectangle,inner sep=2pt,scale=0.5] {#3};}
\newcommand*\annotatedFigureBox[4]{\annotatedFigureBoxCustom{#1}{#2}{#3}{#4}{white}{white}{black}{black}}

\newenvironment {annotatedFigure}[1]{\centering\begin{tikzpicture}
\node[anchor=south west,inner sep=0] (image) at (0,0) { #1};\begin{scope}[x={(image.south east)},y={(image.north west)}]}{\end{scope}\end{tikzpicture}}
\usepackage{enumitem}
\setlist[description]{leftmargin=0cm,labelindent=0cm}

\definecolor{CombinedColor}{HTML}{DEEED4}
\definecolor{ModelColor}{HTML}{FFF0C5}
\definecolor{HumanColor}{HTML}{D9E8F6}
\definecolor{DarkCombinedColor}{HTML}{B1BEA9}
\definecolor{DarkModelColor}{HTML}{CCC09D}
\definecolor{DarkHumanColor}{HTML}{ADB9C4}
\definecolor{CustomGrey}{HTML}{888888}
\definecolor{C0}{HTML}{F5F5F5}
\definecolor{C1}{HTML}{E8E8E8}
\definecolor{C2}{HTML}{DCDCDC}
\definecolor{C3}{HTML}{C0C0C0}
\definecolor{C4}{HTML}{A0A0A0}
\definecolor{C5}{HTML}{808080}
\definecolor{C6}{HTML}{606060}
\definecolor{measured}{HTML}{66c2a5}
\definecolor{condition}{HTML}{5e4fa2}
\definecolor{fixed}{HTML}{f46d43}

\setul{0.5ex}{0.3ex}
\newcommand{\humanul}[1]{\setulcolor{HumanColor}\ul{#1}}
\newcommand{\modelul}[1]{\setulcolor{ModelColor}\ul{#1}}
\newcommand{\combul}[1]{\setulcolor{CombinedColor}\ul{#1}}

\newtcolorbox{siderules}[2][]{blanker, breakable, 
     left=3mm, right=3mm, top=0mm, bottom=0mm,
     borderline west={4pt}{0pt}{#2},
     before upper=\indent, parbox=false, before upper={\noindent}, #1}

\newcommand{\shaded}[4]{
    \vspace{0.15em}
    \noindent
    \begin{minipage}{\linewidth}
        \begin{tcolorbox}[skin=bicolor,fonttitle=\bfseries,coltitle=black,colbacktitle=#2,colback=#2!20,colframe=#2,before skip=0pt,after skip=0pt,title=#1,left=3pt, right=3pt, top=3pt, bottom=3pt,boxsep=0pt,colbacklower=#2!10,middle=0.4em,toptitle=2pt, bottomtitle=2pt,sharp corners=all,boxrule=0mm, leftrule=1mm]
            \emph{Definition}: #3
               \tcblower
            \emph{Attributes}: #4         
        \end{tcolorbox}%
    \end{minipage}
    \vspace{0.05em}
    \newline
}

\pgfkeys{
    /dimshaded/.cd,
    title/.store in=\dimshadedtitle,
    color/.store in=\dimshadedcolor,
    definition/.store in=\dimshadeddefinition,
    attrone/.store in=\dimshadedattrone,
    attronecontent/.store in=\dimshadedattronecontent,
    attrtwo/.store in=\dimshadedattrtwo,
    attrtwocontent/.store in=\dimshadedattrtwocontent,
    attrthree/.store in=\dimshadedattrthree,
    attrthreecontent/.store in=\dimshadedattrthreecontent,
    attrfour/.store in=\dimshadedattrfour,
    attrfourcontent/.store in=\dimshadedattrfourcontent,
}

\newcommand{\dimshaded}[1]{  % 11 parameters: title, color, definition, (title1,content1), (title2,content2), (title3,content3), (title4,content4)
    \pgfkeys{/dimshaded/.cd,
        attrone={},
        attronecontent={},
        attrtwo={},
        attrtwocontent={},
        attrthree={},
        attrthreecontent={},
        attrfour={},
        attrfourcontent={}
    }
    \pgfkeys{/dimshaded/.cd,#1}
    \vspace{0.15em}
    \noindent
    \noindent
    \begin{minipage}{\linewidth}
        \begin{tcolorbox}[
            skin=enhanced,
            fonttitle=\bfseries,
            coltitle=black,
            colbacktitle=\dimshadedcolor,
            colback=\dimshadedcolor!50,  % Definition background
            colframe=\dimshadedcolor,
            before skip=0pt,
            after skip=0pt,
            title=\dimshadedtitle,
            left=3pt,
            right=3pt,
            top=3pt,
            bottom=0pt,
            boxsep=0pt,
            sharp corners=all,
            boxrule=0mm,
            leftrule=1mm,
            toptitle=2pt, 
            bottomtitle=2pt
        ]
        \emph{Definition}: \dimshadeddefinition
        \tcbline
        \emph{Attributes}:
        \tcbline
        \vspace{-0.5em}
        \begin{tcolorbox}[
            colback=\dimshadedcolor!20,
            colframe=\dimshadedcolor!50,
            left=0pt,right=-3pt,top=0pt,bottom=0pt,
            boxrule=0mm,
            before skip=0pt,
            after skip=0pt,
            width=\linewidth,
            enlarge left by=-3pt,
            sharp corners=all,
        ]
            \noindent\begin{minipage}[t]{0.15\linewidth}
                \raggedright
                \textbf{\dimshadedattrone}
            \end{minipage}%
            \begin{minipage}[t]{0.83\linewidth}
                \raggedright
                \dimshadedattronecontent
            \end{minipage}
        \end{tcolorbox}
        \ifx\dimshadedattrtwo\empty\else
            \begin{tcolorbox}[
                colback=\dimshadedcolor!40,
                colframe=\dimshadedcolor!50,
                left=0pt,right=0pt,top=0pt,bottom=0pt,
                boxrule=0mm,
                before skip=0pt,
                after skip=0pt,
                width=\linewidth,
                enlarge left by=-3pt,
                sharp corners=all,
            ]
                \noindent\begin{minipage}[t]{0.15\linewidth}
                    \raggedright
                    \textbf{\dimshadedattrtwo}
                \end{minipage}%
                \begin{minipage}[t]{0.83\linewidth}
                    \raggedright
                    \dimshadedattrtwocontent
                \end{minipage}
            \end{tcolorbox}
        \fi
        \ifx\dimshadedattrthree\empty\else
            \begin{tcolorbox}[
                colback=\dimshadedcolor!20,
                colframe=\dimshadedcolor!50,
                left=0pt,right=0pt,top=0pt,bottom=0pt,
                boxrule=0mm,
                before skip=0pt,
                after skip=0pt,
                width=\linewidth,
                enlarge left by=-3pt,
                sharp corners=all,
            ]
                \noindent\begin{minipage}[t]{0.15\linewidth}
                    \raggedright
                    \textbf{\dimshadedattrthree}
                \end{minipage}%
                \begin{minipage}[t]{0.83\linewidth}
                    \raggedright
                    \dimshadedattrthreecontent
                \end{minipage}
            \end{tcolorbox}
        \fi
        \ifx\dimshadedattrfour\empty\else
            \begin{tcolorbox}[
                colback=\dimshadedcolor!40,
                colframe=\dimshadedcolor!50,
                left=0pt,right=0pt,top=0pt,bottom=0pt,
                boxrule=0mm,
                before skip=0pt,
                after skip=0pt,
                width=\linewidth,
                enlarge left by=-3pt,
                sharp corners=all,
            ]
                \noindent\begin{minipage}[t]{0.15\linewidth}
                    \raggedright
                    \textbf{\dimshadedattrfour}
                \end{minipage}%
                \begin{minipage}[t]{0.83\linewidth}
                    \raggedright
                    \dimshadedattrfourcontent
                \end{minipage}
            \end{tcolorbox}
        \fi
        \end{tcolorbox}%
    \end{minipage}
    \vspace{0.05em}
    \newline
}

\newcommand{\requirementbox}[4]{
    \vspace{0.25em}
    \noindent
    \begin{minipage}{\linewidth}
        \begin{tcolorbox}[skin=bicolor,fonttitle=\bfseries,coltitle=black,colbacktitle=#2,colback=#2!20,colframe=#2,before skip=0pt,after skip=0pt,title=#1,left=3pt, right=3pt, top=3pt, bottom=3pt,boxsep=0pt,colbacklower=#2!10,middle=0.4em,toptitle=2pt, bottomtitle=2pt,sharp corners=all,boxrule=0mm, leftrule=1mm]
            #3  
            \tcblower
            #4
        \end{tcolorbox}%
    \end{minipage}
    \vspace{0.2em}
    \newline
}

\newcommand{\oppbox}[2]{
    \vspace{0.25em}
    \noindent
    \begin{minipage}{\linewidth}
        \begin{tcolorbox}[
              enhanced,
              boxrule=0pt,
              toprule=1pt,
              boxsep=0pt,
              left=2pt,
              right=2pt,
              top=2pt,
              sharp corners=all,
              bottom=2pt,
            ]%%
  \emph{Opportunity {#1}:} #2
\end{tcolorbox}
    \end{minipage}
    \newline
}

\newtcolorbox{coloredbar}{
    enhanced,
    boxrule=0pt,
    colback=white,
    colframe=white,
    sharp corners,
    leftrule=3mm,
    coltitle=black,
    fonttitle=\bfseries,
    colbacktitle=white,
    colupper=black,
    attach boxed title to top left={},
    boxed title style={
        colback=white,
        colframe=white,
    },
}

\makeatletter
\newtcbox{\kwColorBox}[1][]{on line,fontupper=\footnotesize\sffamily\bfseries\small,boxrule=0.5pt,arc=2pt,coltext=#1,colback=#1!10!white,colframe=#1,boxsep=0pt,left=1.5pt,right=1.5pt,top=1.5pt,bottom=1.5pt}
\makeatother

\newcommand{\kw}[2]{%
    \begin{kwColorBox}[#2]%
    {#1}%
    \end{kwColorBox}%
    \xspace%
}

\newcommand{\rawuserdonotuseh}[1]{\kw{#1}{DarkHumanColor}}
\newcommand{\defhuman}[1]{\rawuserdonotuseh{\phantomsection\label{interface:#1}#1}}
\newcommand{\refhuman}[1]{\hyperref[human:#1]{\rawuserdonotuseh{#1}}}

\newcommand{\rawuserdonotusem}[1]{\kw{#1}{DarkModelColor}}
\newcommand{\defmodel}[1]{\rawuserdonotusem{\phantomsection\label{model:#1}#1}}
\newcommand{\refmodel}[1]{\hyperref[model:#1]{\rawuserdonotusem{#1}}}

\newcommand{\rawuserdonotuse}[1]{\kw{#1}{DarkCombinedColor}}

\newcommand{\reffbp}[1]{\hyperref[fbp:#1]{\rawuserdonotuse{#1}}}

\newcommand*\rot{\rotatebox{-90}}
\newcommand*\OK{\ding{51}}
\newcommand*\OKGREY{\color{gray}\ding{51}}
\newcommand*\OKBLUE{\color{blue}\ding{51}}

\begin{document}

%%
%% The "title" command has an optional parameter,
%% allowing the author to define a "short title" to be used in page headers.
%\title[A Framework of Human Feedback for RL]{A Conceptual Model of Human Feedback\\ for Reinforcement Learning in Human-AI Interaction}
\title[Mapping out the Space of Human Feedback for Reinforcement Learning]{Mapping out the Space of Human Feedback\\ for Reinforcement Learning: A Conceptual Framework}

\author{Yannick Metz}
\email{yannick.metz@uni-konstanz.de}
\orcid{0000-0001-5955-4487}
\affiliation{%
  \institution{University of Konstanz}
  \streetaddress{Universität Konstanz}
  \city{Konstanz}
  \country{Germany}
  \postcode{78457}
}
\affiliation{%
  \institution{ETH Zurich}
  \city{Zurich}
  \country{Switzerland}}

\author{David Lindner}
\orcid{0000-0001-7051-7433}
\affiliation{%
  \institution{ETH Zurich}
  \city{Zurich}
  \country{Switzerland}}
%\email{larst@affiliation.org}

\author{Raphaël Baur}
\orcid{0009-0005-2629-0784}
\affiliation{%
  \institution{ETH Zurich}
  \city{Zurich}
  \country{Switzerland}}
%\email{rabaur@ethz.ch}

\author{Mennatallah El-Assady}
\orcid{0000-0001-8526-2613}
\affiliation{%
  \institution{ETH Zurich}
  \city{Zurich}
  \country{Switzerland}}
%\email{elassady@ethz.ch}

%%
%% By default, the full list of authors will be used in the page
%% headers. Often, this list is too long, and will overlap
%% other information printed in the page headers. This command allows
%% the author to define a more concise list
%% of authors' names for this purpose.
\renewcommand{\shortauthors}{Y. Metz et al.}

%%
%% The abstract is a short summary of the work to be presented in the
%% article.
\begin{abstract}
Reinforcement Learning from Human feedback (RLHF) has become a powerful tool to fine-tune or train agentic machine learning models. Similar to how humans interact in social contexts, we can use many types of feedback to communicate our preferences, intentions, and knowledge to an RL agent.
However, applications of human feedback in RL are often limited in scope and disregard human factors.
In this work, we bridge the gap between machine learning and human-computer interaction efforts by developing a shared understanding of human feedback in interactive learning scenarios.
We first introduce a taxonomy of feedback types for reward-based learning from human feedback based on nine key dimensions. Our taxonomy allows for unifying human-centered, interface-centered, and model-centered aspects.
In addition, we identify seven quality metrics of human feedback influencing both the human ability to express feedback and the agent’s ability to learn from the feedback.
Based on the feedback taxonomy and quality criteria, we derive requirements and design choices for systems learning from human feedback. We relate these requirements and design choices to existing work in interactive machine learning. In the process, we identify gaps in existing work and future research opportunities.
We call for interdisciplinary collaboration to harness the full potential of reinforcement learning with data-driven co-adaptive modeling and varied interaction mechanics.
\end{abstract}

%%
%% The code below is generated by the tool at http://dl.acm.org/ccs.cfm.
%% Please copy and paste the code instead of the example below.
%%


%%
%% Keywords. The author(s) should pick words that accurately describe
%% the work being presented. Separate the keywords with commas.

\received{20 June 2023}
%\received[revised]{12 March 2009}
%\received[accepted]{5 June 2009}

\begin{teaserfigure}
  \centering
  \includegraphics[width=1.0\linewidth]{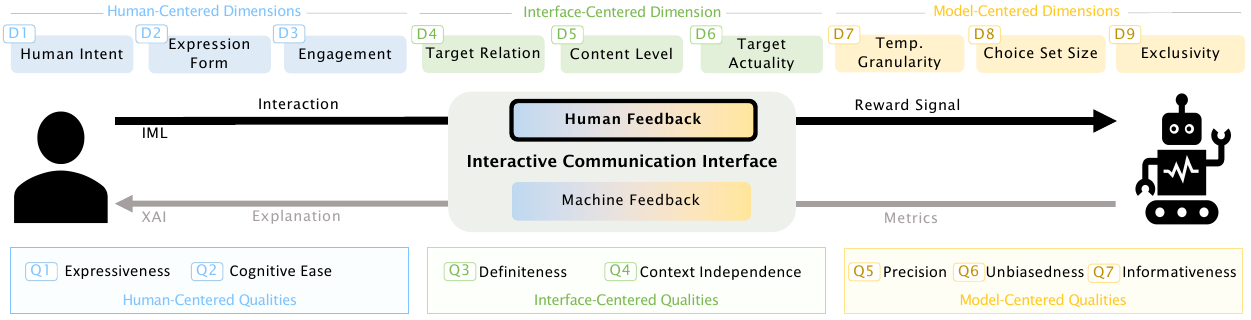}
  \caption{%
    In the context of \textit{Reinforcement Learning}, we present a \textit{conceptual framework of human feedback} based on nine dimensions: We identify \humanul{human-centered}, \combul{interface-centered} and \modelul{model-centered} dimension. For different types of feedback, we also need to consider \textit{human-centered, interface-centered and model-centered qualities} which influence how feedback can be collected and processed.
  }
  \label{fig:teaser}
\end{teaserfigure}

%%
%% This command processes the author and affiliation and title
%% information and builds the first part of the formatted document.
\maketitle

\section{Introduction}
\label{sec:introduction}

The development of intelligent agents that can execute tasks, align with human values, and adhere to our preferences represents a significant long-term goal in artificial intelligence. Reinforcement Learning from Human Feedback (RLHF) and related approaches have emerged as promising tools for training agents, shaping them to follow instructions and align with user preferences~\cite{christiano2017deep, ibarz2018reward, ouyang2022training}.

While research on training intelligent agents with human input has a rich history—including approaches like imitation learning~\cite{hussein2018}, inverse RL~\cite{ng1999policy}, and interactive RL~\cite{cruz2021interactive} — RLHF specifically has gained prominence due to its success in fine-tuning language models~\cite{ouyang2022training}. 
Past research has explored diverse interaction modalities for human feedback, including demonstrations, evaluative and comparative feedback, and corrections. However, the field lacks a common classification of design dimensions for human feedback and a unified perspective that bridges human-computer interaction and machine learning.

Human feedback in real-world scenarios encompasses a wide array of communication strategies adapted to various contexts, tasks, and recipients~\cite{cherry1966human}. In contrast, feedback commonly used in RLHF, in particular for the fine-tuning of generative models such as LLM, i.e., pairwise comparisons~\cite{christiano2017deep, wirth2017survey} of outputs, represents just a very limited way to provide feedback. We therefore argue that the space for possible feedback in human-AI communication should be broadened. This paper aims to address two key research questions:

\begin{itemize}
    \item What different kinds of human feedback exist?
    \item What constitutes good human feedback?
\end{itemize}

To answer the first question, we propose a taxonomy that structures human feedback along nine dimensions, linking these to the broad spectrum of existing work in human-AI communication. For the second question, we discuss seven quality metrics to evaluate the effectiveness of human feedback.
The paradigm shift towards more expressive, targeted, and dynamic human feedback enables a flexible, multi-faceted training process for intelligent agents. In this new paradigm, humans may interact with agents at various stages of training and deployment while agents learn from diverse data sources, environments, and feedback types.

%Addressing these challenges requires carefully considering how users communicate feedback to learning systems—a crucial aspect in designing RL-based agents and systems.
%Contributions
%This work focuses on key questions about using human feedback in RL, exploring how to effectively utilize different feedback sources, what design choices to consider, and what lessons can be drawn from existing work. 

\paragraph{Contributions} Our contribution is threefold:

\textbf{(i)}~We introduce a \textbf{taxonomy of nine dimensions} of human feedback towards AI agents, unifying model-centered, interface-centered, and user-centered aspects. This taxonomy provides a foundation for classifying existing feedback patterns and includes a formalization of human feedback.

\textbf{(ii)}~We present \textbf{seven quality metrics} that summarize reported quality criteria from a broad range of literature, discussing their manifestations and measurement methods.

\textbf{(iii)}~Based on the taxonomy and feedback qualities, we \textbf{derive requirements and design choices} for reinforcement learning systems integrating human feedback. We relate these to existing work across interactive machine learning and visual analytics, \textbf{identifying research opportunities} for dynamic multi-type human feedback.

%\clearpage
\section{Problem Statement and Background}
\label{sec:related_work}
We start by establishing the problem space and key concepts and review related work.

\subsection{Problem Characterization}
\label{subsec:problem_characterization}

\paragraph{Human-AI Communication Space}
\begin{figure}
	\centering
  \includegraphics[width=0.8\linewidth]{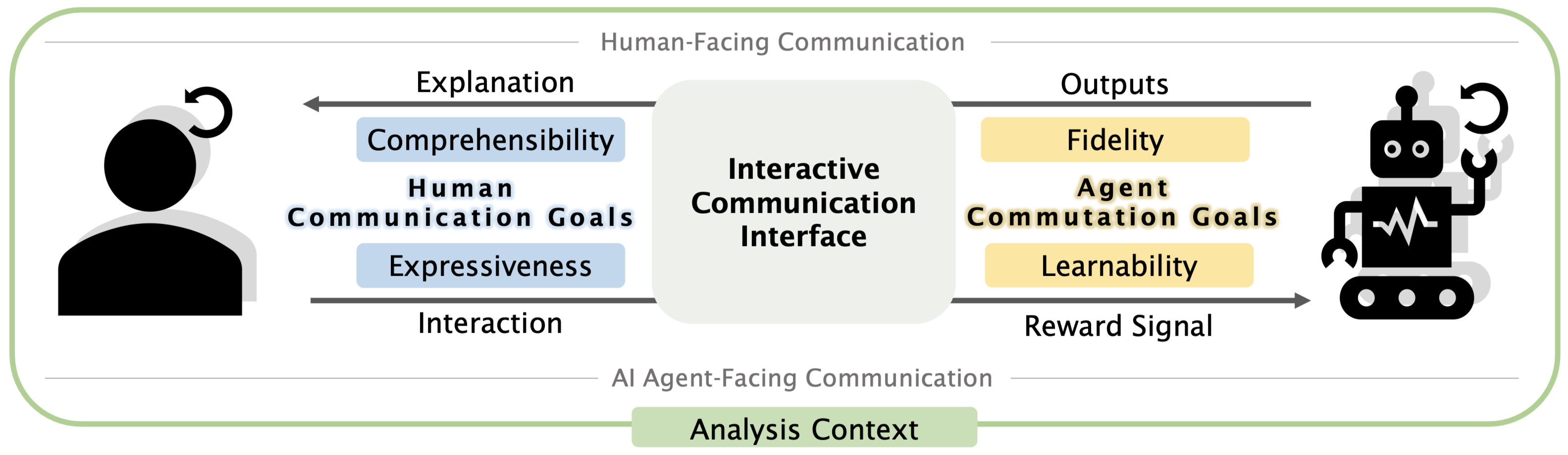}
   \caption{\label{fig:communcation_loop} Human-AI Interaction via Interactive Communication Interfaces: Modal Outputs are communicated to the humans via the interface and presented as comprehensible explanations that should preserve high fidelity. Human feedback is communicated via interactions that must be mapped to a reward signal suitable for RL training.
    }
\end{figure}
In order to effectively bridge different fields like human-computer interaction (HCI) and machine learning, we develop our conceptual framework from first principles. We identify the key actors and goals in the human-AI communication process.

Throughout the paper, we will use two running examples: Due to its ubiquity in current RLHF research~\cite{ouyang2022training, bai2022constitutionalaiharmlessnessai, kaufmann2023survey}; we will use the example of human feedback for a text generation model and outline potential feedback in such a use case. Especially beyond simple, verifiable, and single-objective generation tasks, including multi-step reasoning or domain-dependent task execution, more expressive feedback can be highly desirable.\\ Additionally, we consider a more advanced, future-oriented use case of a household robot deployed in a new household environment. Even if the robot has been pre-trained in general environments, e.g., via supervised~\cite{hussein2018} or fully-autonomous learning~\cite{kober2013reinforcement} in simulation, it needs to learn to act according to the personal preferences of its new owners. This includes stowing objects in desired locations, avoiding certain behaviors, and using location-specific tools. Ideally, we want intuitive ways to dynamically express our goals and preferences to the agent, such as demonstrating desired behavior or commenting on observed actions.

To achieve this type of teaching, expressive human-to-AI communication is crucial. Humans intuitively use various forms of feedback in social contexts~\cite{lin2020review}, but these feedback utterances need to be translated for AI learning algorithms. Effective teaching also requires the learning agent to communicate back to the human teacher. We can, therefore, identify two directions for \textit{Human-AI interaction}.

In \autoref{fig:communcation_loop}, we provide an overview of the \textit{Human-AI communication} space~\cite{el2022biases}. The key elements are:
\begin{itemize}[noitemsep]
\item \humanul{\textbf{Humans}} -- Involved as teachers and potential users of the agent's capabilities, humans have preferences about the agent's behavior.
\item \modelul{\textbf{AI Agents}} -- Agents optimize their behavior using a reward model trained from human feedback.
\item \combul{\textbf{H-AI Communication Interfaces}} -- A variety of interfaces exist, ranging from visual and numerical to natural language, gestures, or mimics~\cite{Li2021}.
\item \combul{\textbf{Analysis Context}} -- The human-AI interaction is embedded in a surrounding context, determining available information and feedback communication methods. We must decide on a specific analysis context, which restricts the contextual information and stakeholders under consideration.
\end{itemize}

\textit{Human feedback} is transmitted from human to agent. Humans benefit from high \textit{expressiveness} (see~\autoref{sec:qualities}), allowing effective communication of internal knowledge, goals, and preferences. Conversely, agents require \textit{learnable} input that can reasonably translate into model updates. Traditionally, the research field of interactive machine learning (IML) addresses this direction~\cite{amershi2014power, review_interactive_ml_1}. In the opposite direction, agents provide \textit{machine feedback} via model outputs or auxiliary data. These metrics are presented as explanations to humans. Machine feedback must balance \textit{fidelity} and \textit{comprehensibility}. High-fidelity outputs accurately describe model behavior~\cite{gunning2019xai}, but this information must be comprehensible to humans via abstractions or filtering, which limits fidelity. This direction is addressed by explainable AI (XAI)~\cite{SurveyXAI}. We highlight these four high-level goals in \autoref{fig:teaser}. In this work, we focus on the first direction.

The context of human-AI interaction directly influences human feedback \cite{lindner_humans_2022}. Different types of context include the \textit{(sociotechnical) personal context} (i.e., the human's background) \cite{lindner_humans_2022}, the \textit{task context} (the basis for the human-AI interaction, e.g., text generation vs. household tasks), and the \textit{utterance context} (the immediate context within a potentially longer interaction process). Current RLHF approaches often fail to consider these contextual factors by not modeling them appropriately. Rich, diverse interactions via expressive human feedback can enable the capture of additional contextual information.

We use this space of \humanul{human}, \combul{interface}, and \modelul{model} to categorize expressions of feedback and discuss feedback quality. We posit that human feedback can be categorized from a human perspective ("How can we categorize the human experience when giving different feedback?"), from an interface perspective ("Which types of interfaces accommodate different feedback?"), Moreover, from a model perspective ("How do we classify feedback for different reward models?"). Besides classifying feedback, we must also address the question "What makes for high-quality feedback?" from human, interface, and model perspectives.

\subsection{Related Work}

\paragraph{Human-in-the-Loop Reinforcement Learning}
Human-in-the-loop reinforcement learning (HITL RL), also known as collaborative RL~\cite{Li2021} or interactive RL~\cite{arzate_cruz_survey_2020}, centers on incorporating human knowledge and feedback. Reinforcement learning, particularly in challenging environments, often suffers from high sample inefficiency and difficult credit assignments when reward signals are sparse. Human experts can leverage their prior knowledge about the environment to enhance training efficiency by guiding RL agents during exploration in large state spaces or when agents are stuck in local optima~\cite{li_human-centered_2019}.
In HITL RL, human teachers use feedback to accelerate or improve an agent's learning process. For instance, a human trainer might observe the learning agent's behavior in an environment and provide positive or negative reward signals to inform the quality of its actions~\cite{maclin2005giving, knox2009interactively, li_human-centered_2019}. Existing surveys have primarily focused on input modalities and learning methods~\cite{lin2020review, arzate_cruz_survey_2020}. We draw from work in this area and expand it to the domain of RLHF.

\paragraph{Reinforcement Learning from Human Feedback}
In scenarios where no environment reward function is available, Reinforcement Learning from Human Feedback (RLHF), using pairwise preferences, has emerged as a viable learning technique~\cite{christiano2017deep}. Pairwise preferences have become so ubiquitous that RLHF is often synonymously with ``RL from human \textit{preferences}''. Existing surveys~\cite{casper2023open, lindner_humans_2022} have highlighted limitations of current RLHF approaches, such as overly simplistic modeling of humans, misspecification, and limitations in expressiveness. Our paper lays out a framework to study and address these limitations by broadening the space of human feedback for RL systems.

\paragraph{Existing Classifications of Feedback}
Another classification separates feedback into three categories~\cite{Li2021, lin2020review}:

\textit{Explicit} feedback: Directly usable as reward model input and consciously given by a human observer.
\textit{Implicit} feedback: Indirectly processed information humans communicate, such as facial expressions or non-directed natural language.
\textit{Multi-modal} feedback: A combination of different feedback types, though this concept is not well-defined beyond using multiple methods simultaneously. 

This classification, while helpful, is rather broad and lacks specificity in defining multi-modality.

Jeon et al.~\cite{jeon_reward-rational_2020} present a unifying formalism for reward learning from various feedback sources. Their reward-rational (implicit) choice model represents different types of feedback—including demonstrations, corrections, comparisons, or direct rewards—as the human choosing an option from a choice set. This allows for a standard implementation of reward models across different feedback types. We extend this work by incorporating HCI aspects and developing a structured conceptual feedback framework.

A survey by Fernandes et al.~\cite{fernandes2023bridging} discusses a series of common formats for feedback in the context of natural language generation, including numerical and ranking feedback and natural language. The differentiation of feedback formats is only a minor part of the survey and is not integrated into a common shared framework or design space. 

Concurrent work by Kaufmann et al.~\cite{kaufmann2023survey} presents a feedback classification along several dimensions but does not focus on HCI aspects and uses a more straightforward classification overall. Their classification is inspired by preliminary work~\cite{metz2023rlhf}.

\section{Mapping out the Space of Human Feedback in Reinforcement Learning}
While existing work in human-in-the-loop RL raises some crucial points, it does not provide a consistent view of feedback types, conflating different concepts and levels. This section proposes a novel, exhaustive categorization of human feedback types. The following sections discuss feedback quality criteria and derive actionable requirements for human-AI interaction systems.

%\subsection{A Conceptual Framework of Human Feedback}
%\label{sec:conceptual_model}

%Let us revisit our running example. In a social context, we intuitively use various methods to provide feedback to a household robot, such as commenting on a specific action (\textit{"You put the object where I wanted to"}), comparing to other behavior (\textit{"Your previous trial was better than this one"}), stating a goal (\textit{"I want the table to be covered with a green tablecloth"}), or even via non-verbal interactions like gestures of approval or disapproval.

%As designers of Human-to-AI communication, we face two main challenges: (1) designing expressive user interactions, including in non-physical learning environments, and (2) ensuring the given feedback is learnable by the agent, i.e., can be interpreted as a reward function by an RL agent. Structuring the feedback space is crucial for RL and interactive machine learning.

\subsection{Methodology and Process} 
We propose a conceptual framework to structure and unify the space of different feedback mechanisms in related work and guide future exploration and research.

\begin{figure}
	\centering
  \includegraphics[width=\linewidth]{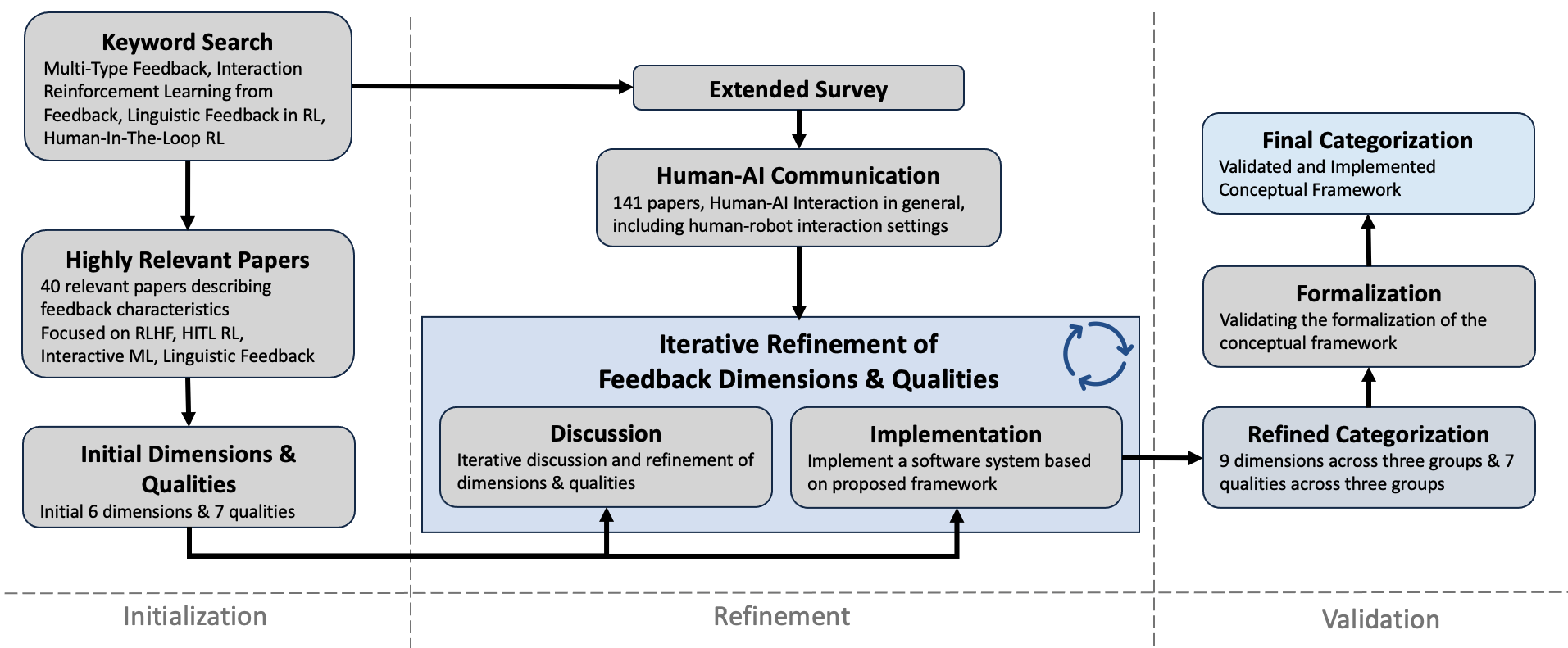}
   \caption{\label{fig:survey_methodology} The methodology used to create the conceptual framework: During the \textit{initialization}, we extract a series of initial dimensions and qualities that exhaustively classify a set of highly relevant surveyed papers. In a second \textit{refinement} phase, we iteratively refine the conceptual framework based on additional papers, external input, and internal discussions. Importantly, the framework is also implemented as a software system and validated to handle multiple use cases. Finally, we \textit{validate} the refined conceptual framework via the implementation and in expert interviews.
    }
\end{figure}

We designed the conceptual framework through an iterative survey and refinement process. In multiple iterations, we surveyed work on possible types of human feedback for training intelligent agents and then discussed potential dimensions for classifying human feedback. Discussions were conducted in a core group of four contributors, with an additional open-group session involving six experts. Based on these discussions, we defined dimensions and quality criteria. For the chosen dimensions and quality criteria, we specified the following requirements:

\begin{itemize}
    \item Exhaustiveness: All surveyed papers describing types of human feedback for agent training must be classifiable with the given dimensions.
    \item Orthogonality: Significantly different approaches should be classified into separate dimensions.
    \item Relevance: The chosen dimensions should relate to the elements of the human-to-AI communication process and enable deriving quality metrics and requirements.
\end{itemize}

We begin with a limited survey of 18 papers on natural language feedback. We identified five initial types of human feedback (equivalent to the types specified in~\autoref{subsec:types_of_feedback}) and defined a first candidate set of possible dimensions to classify feedback types. We then extended the candidate set with an additional 22 papers describing approaches for human feedback in reinforcement learning. We defined a set of 6 dimensions and seven quality criteria to classify the surveyed papers accurately. 

Finally, we performed an exhaustive survey of papers in relevant adjacent fields based on a keyword search. We searched for relevant papers matching the following keywords: \textit{Human Feedback Reinforcement Learning}, \textit{Human-In-The-Loop Reinforcement Learning}, \textit{Interactive Reinforcement Learning}, \textit{Human-Agent Interaction}, and \textit{Agent Machine Teaching}. We considered papers from 2008 to 2024. For the final survey, 141 papers were surveyed in journals and conferences, including \textit{AAAI}, \textit{AAMAS}, \textit{ACL}, \textit{CHI}, \textit{CoRL}, \textit{HAI}, \textit{HRI}, \textit{IJCAI}, \textit{ICRA}, \textit{NeurIPS}. We restricted the search to papers describing the use of human feedback in human-agent interactions. To identify dimensions for future research, we did not limit ourselves to papers from core \textit{RLHF}, i.e., learning a reward model from human feedback. Instead, we considered various approaches utilizing feedback in human-AI interaction. Based on the final survey, we decided on nine dimensions and seven quality criteria summarized from the surveyed literature.\\
For the formalization introduced in the following section, we aimed to find a common abstraction of terms and concepts described in the surveyed work beyond a specific scenario that builds a foundation for the formalization of identified dimensions,

Beyond a systematic review, the taxonomy of feedback types serves two primary purposes: (1) We can specify a set of human-, interface- and model-centered qualities that must be considered when using human feedback in a system (\autoref{sec:qualities}). (2) It guides the design of RLHF systems, including user interfaces, feedback processing, and reward modeling (\autoref{sec:deriving_requirements}).

\subsection{Formalization of Human Feedback}
Before introducing the identified dimensions on a conceptual level, we also want to discuss a formalization of the human feedback space to foster the development of machine learning algorithms. Therefore, we build upon previous efforts to formalize the notion of reinforcement learning from human feedback (RLHF) \cite{christiano2017deep, kaufmann2023survey}, in order to combine perspectives from machine learning and human-computer interaction.

\begin{figure}[tbh]
    \centering
    \includegraphics[width=0.9\linewidth]{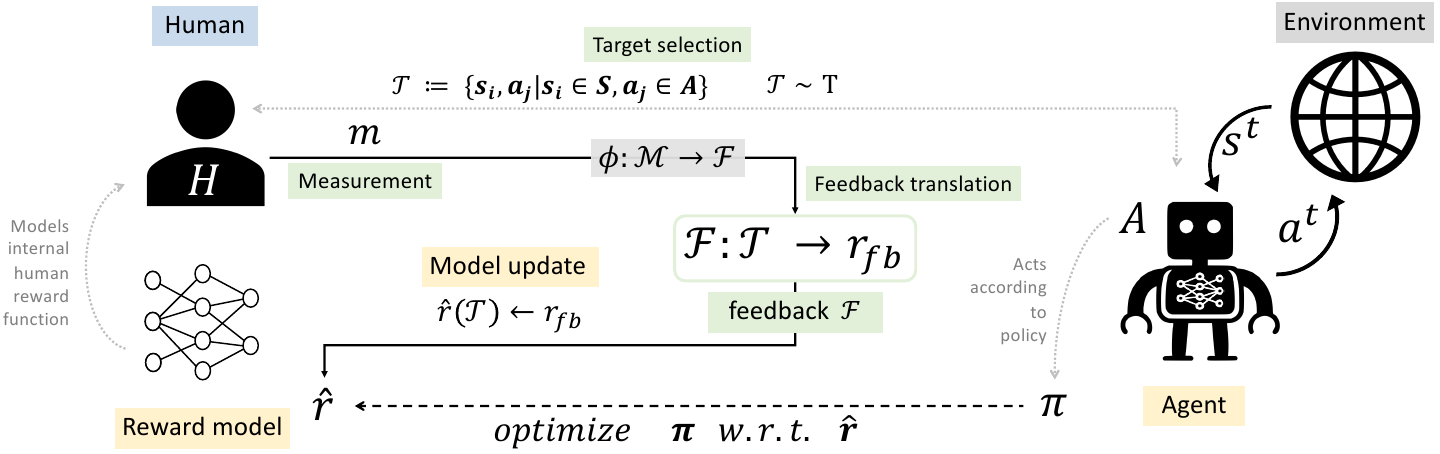}
    \caption{The feedback process formalized: Humans generate feedback for agents that they observe acting in an environment. A feedback instance is based on a target and measured values. It is translated into a form that can be used to update a reward model. The policy of the agent, in turn, is optimized with respect to the updated reward model.}
    \label{fig:feedback_process}
\end{figure}
 As in traditional reinforcement learning (RL), we consider an agent that acts in an environment over a sequence of steps. Concretely, at a timestep $t$, the agent chooses an action $a^t \in \mathcal{A} \subseteq \mathbb{R}^M$ given an observation $o^t \in \mathcal{O}$ of its current state $s^t \in \mathcal{S} \subseteq \mathbb{R}^N$. Here, we assumed, without the loss of generality, that we can map all actions $a^t$ and states $s^t$ to a real-valued vector. Contrary to traditional RL, however, we lack access to the \textit{true}\footnote{In many scenarios, a true reward function might not exist, especially if the desired output can only be judged based on subjective, potentially inconsistent, preferences, such as aesthetic criteria.} {\textbf{reward function}} $r: \mathcal{S} \times \mathcal{A} \to \mathbb{R}$, that provides a reward $r^t = r(s^t, a^t)$ for a state-action pair. In many real-world problems, defining a valid reward function is as hard as solving the problem itself. Instead, RLHF resorts to learning an estimate $\hat{r}: \mathcal{S} \times \mathcal{A}$ of the true reward function based on human preferences transmitted via human feedback. The goal is to learn an optimal policy $\pi^\ast: s^t \to a^t$ that maximizes the expected discounted rewards according to $\hat{r}$. In our example, the observations can be the robot's sensor readings and camera input to capture the state of the home environment. Actions can be low-level (motor controls of robot joints) or high-level (moving to a location, picking/placing objects, etc.). In this work, we do not assume a certain action granularity. However, the task of aligning available feedback with actions is a challenge in itself, aided by the availability of varied feedback.\\

In this paper, we assume human feedback to be based on a \textbf{measurement} $m$, obtained in the human-AI interaction process. A measurement is defined for a set of measurement variables $m := {v_1, ..., v_k}$, which depend on the design of the human-AI interaction interface. Variables can be based on diverse interactions, including explicit inputs like buttons or sliders and sensors measuring implicit interactions. Language feedback is a versatile type of feedback that can be interpreted in different ways.

Any state information can be interpreted as a measurement as well. We may consider a measurement as a "snapshot" of the interaction process that can be arbitrarily complex based on the chosen analysis context—for example, a "goal state" like a desired arrangement of objects on shelve.

\begin{figure}[tbh]
    \centering
    \includegraphics[width=1\linewidth]{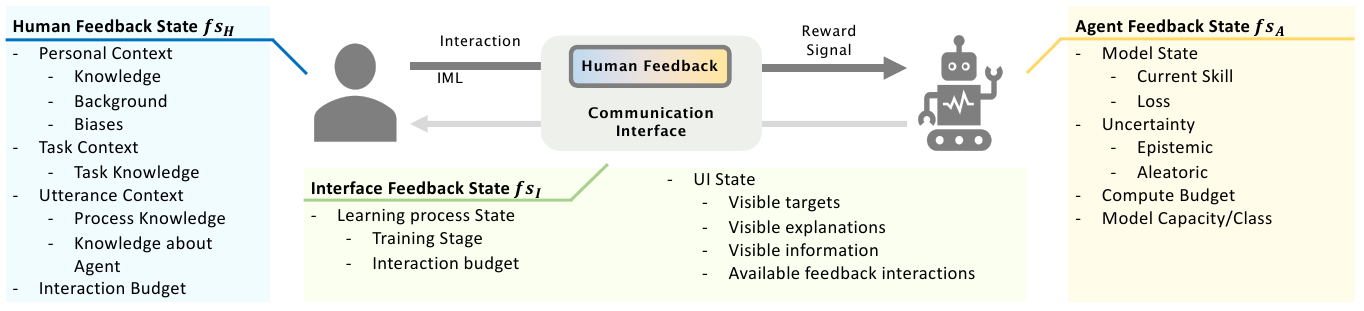}
    \caption{The feedback state consists of three different sub-states: (1) The \humanul{human feedback state}, (2) the state of the \combul{interface}, and (3) the \modelul{model state}.}
    \label{fig:feedback_state}
\end{figure}

Based on the Human-AI communication space~\autoref{subsec:problem_characterization}, we define a \textbf{feedback state} $fs$ in which the interaction takes place and that a measurement is conditioned on. The feedback state can be decomposed into \textit{three sub-states} mirroring the actors: (1) A \humanul{human feedback state} $fs_h$, which might contain information like the knowledge state or exhaustion level of the human, (2) the \combul{interface feedback state} $fs_I$, i.e. in which state is the learning process, which information and interactions are available to the human, and which measurements are used, (3) an \modelul{agent feedback state} $fs_a$, e.g. containing the model state and computation budget. The feedback state is summarized in~\autoref{fig:feedback_state}. The use of a human model $H$ and an agent/robot model $R$ to capture such aspects of human-AI communication has already been proposed in existing work~\cite{hadfield2016cooperative}. However, in this work, we go beyond already proposed modeling approaches.

As part of the reward learning process, "raw" measurements need to be translated to a usable reward signal that can be used to optimize a reward model. To achieve this, a vector of measurements $m$ must be assigned to a \textbf{target} $\mathcal{T} \subseteq \Xi$ from the space of possible trajectories $\Xi$, i.e., all possible behaviors. On a high level, a target $\mathcal{T}$ is nothing more than a set of features from the observation or action spaces (which are the input to the reward model): $\mathcal{T} := \{s_i,a_j | s_i \in , a_j \in \mathbb{A}\}$. As a special case of $s_i=s,a_j=a$, we can write a target as a set of state-action trajectories $\mathcal{T} = \{[(s^0,a^0), (s^1,a^1),...,(s^i,a^i)],...\}$, for example, a target can be an entire episode or even a set of episodes. This is related to the concept of grounding functions found in the reward rational (implicit) choice framework \cite{jeon_reward-rational_2020}. Special cases include a single state-action pair $\mathcal{T} = (s, a)$ and the entire set of possible trajectories $\mathcal{T} = \Xi$. In our example, a target of feedback can be a single state-action pair, like placing a single object in a particular location based on the observation of the existing locations, or feedback for a whole trajectory of behavior, like cleaning an entire room. 

Secondly, as a usable reward signal, the target $t$ must be mapped to the inputs in the domain of the reward function $r$. The content of the utterance must be translated to a feedback reward value $r_{fb} \in R(r)$, with $R(r)$ being the range of the reward function $r$, commonly the real numbers $R(r) = \mathbb{R}$. We define a processed \textbf{human feedback instance} $\mathcal{F}$ as a mapping from a target $\mathcal{T}$ to a feedback value $r_{fb}$:
$$
 \mathcal{F}: \mathcal{T} \rightarrow r_{fb} \in \mathbb{R}
$$
To generate processed feedback, we need to design a \textbf{translation algorithm} $\phi: m \rightarrow (\mathcal{T}, r_{fb})$. A translation algorithm generates or predicts targets and \textbf{feedback values}, i.e., a feedback value encoding, from raw measurements. In the case of simple explicit measurements, like a slider value or buttons indicating preferences, the translation function might be the identify function $\phi(m): m$. This setup is assumed in most existing work in machine learning~\cite{jeon_reward-rational_2020}.

Assuming more complex, implicit feedback, e.g., language-based feedback as in our running example, a simple translation algorithm mapping from language feedback to a value is a sentiment classifier~\cite{krening2016learning}, which maps the raw utterance, contained in $m$ into a feedback value $r_{fb} = {-1,0,+1}$.

Based on our previous discussions, a robust translation algorithm/reward modeling approach should take the feedback state into context~\cite{lindner_humans_2022}. Based on our definition, we may identify two types of measurement variables: (1) \textit{Intrinsic} measurement variables, e.g., the primary user interaction, designed to receive a feedback value $r_{fb}$ for a specified target $\mathcal{T}$, $m_{int}$, (as specified above), and (2) \textit{contextual} measurements $m_{ctx}$, that contain information about the feedback state $fs$. Again, a translation algorithm translates these raw measurements $m_{ctx}$ into a \textbf{context encoding}: $\phi(m_{ctx}) \rightarrow  \mathcal{C}$, which can be utilized for reward model training, e.g., by training multiple rewards models for different contexts (e.g., user groups) or condition a reward model om the context.

Lastly, a key aspect is the selection of targets that are shown to the human user:  We assume that the selection of targets follows a probability distribution $\mathsf{T}$, i.e., $\mathcal{T} \sim \mathsf{T}$, which we call the \textbf{target distribution}. $\mathsf{T}$ may simply be randomly sampled from the agent's behavior or follow more sophisticated strategies based on the feedback state, as found in many querying strategies~\cite{zhan2023query, christiano2017deep}.

The entire process, including key terms, is shown in~\autoref{fig:feedback_process}.

To summarize, the elements we need to consider for human feedback are the choice of \textit{measurement variables and domains}, models of \textit{feedback state}, set of available \textit{targets}, and \textit{translation algorithm} $\phi: m \rightarrow \mathcal{F}$. The dimensions we present in this chapter enable the appropriate choice of elements. We can define the entire process as follows:
\vspace{1.5em}
\begin{equation*}
    \eqnmarkbox[NavyBlue]{meseq1}{\bigl(m_{int}, m_{ctx} | fs_h, fs_a, fs_i)} \xrightarrow[]{\eqnmarkbox[WildStrawberry]{phieq1}{\phi}} \eqnmarkbox[Plum]{feedbeq1}{\bigl( \mathcal{F}: \mathcal{T} \rightarrow r_{fb} | \mathcal{C} \bigl)} \quad \text{with} \eqnmarkbox[Goldenrod]{rewmodel}{\hat{r}_{\mathcal{C}}(\mathcal{T}) = r_{fb}} \quad \text{,} \quad \eqnmarkbox[OliveGreen]{tsampleeq1}{\mathcal{T} \sim \mathsf{T}}
\end{equation*}
\annotate[yshift=0.5em]{above,left}{meseq1}{Measurement, Context}
\annotate[yshift=0.5em]{above,right}{phieq1}{Translation algorithm}
\annotate[yshift=-0.5em]{below,left}{feedbeq1}{Processed Feedback}
\annotate[yshift=0.5em]{above,right}{rewmodel}{Reward Model}
\annotate[yshift=-0.5em]{below,right}{tsampleeq1}{Target distribution}
\vspace{1em}

This basic model opens up an extensive design space of possible human feedback. In the following sections, we introduce nine dimensions that help to structure the space of possible design choices and help us to think about diverse human feedback in a systematic way.

\subsection{Dimensions of Human Feedback}

As shown in \autoref{fig:teaser}, we classify human feedback along nine dimensions split into three categories: \humanul{human-centered dimensions}, \combul{interface-centered \& structural dimensions}, and \modelul{model-centered dimensions}. Despite this grouping, the full set of dimensions influences all design and interpretation stages. 

\begin{figure}
	\centering
  \includegraphics[width=\linewidth]{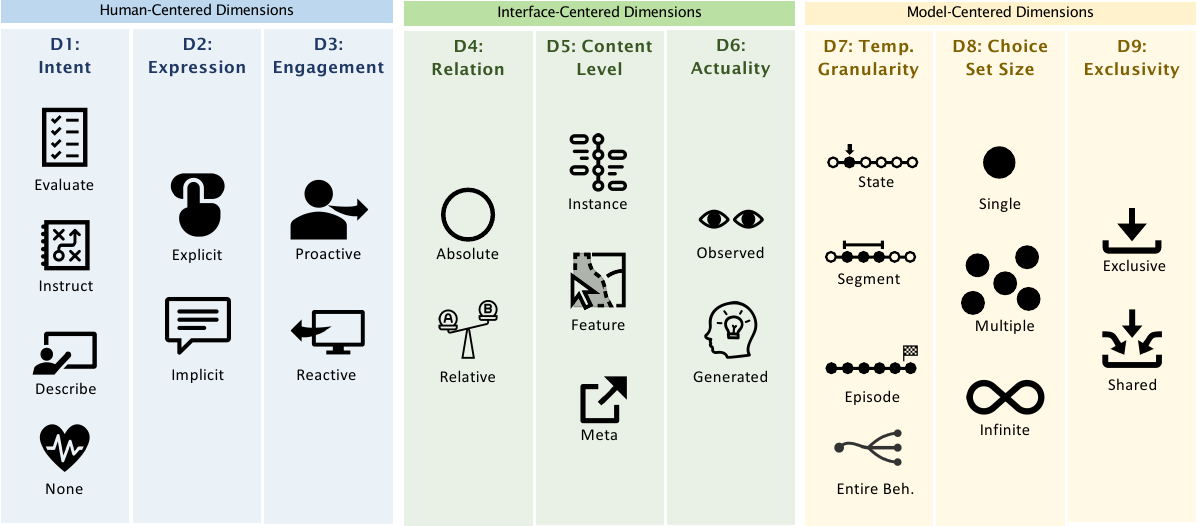}
   \caption{\label{fig:dimensions} In the context of \textit{Reinforcement Learning}, we present a \textit{conceptual framework of human feedback} based on nine dimensions: We identify \humanul{human-centered}, \combul{interface-centered} and \modelul{model-centered} dimension.
    }
\end{figure}

\subsection{Human-Centered Dimensions}
\label{subsec:human-centered-dimensions}
These dimensions are directly connected to the design of human-centered interfaces. As user interface designers, we must ensure that the interfaces match the dimensions of the feedback we want to be able to collect from humans. The interfaces should allow humans to (1) provide feedback according to their \textit{intention}, (2) \textit{expressed} in a way that fits the ability and preferences of the human, (3) with the correct level of \textit{engagement}.\\
%\subsubsection{Human Intent}
%\shaded{Intent \refhuman{D1}}{HumanColor}{\label{human:D1}The intentions or the ``goal'' of human feedback in a given situation.}{\textbf{Evaluate} (give a rating) | \textbf{Instruct} (state what to do)  |  \textbf{Describe} (qualitative account)   |  \textbf{None} (no particular intention)}

\dimshaded{
    title={Intent \refhuman{D1}},
    color={HumanColor},
    definition={\label{human:D1}The intentions or the "goal" of human feedback in a given situation.},
    attrone={Evaluate},
    attronecontent={Evaluative feedback can be in the form of \textit{binary critique} \cite{knox2012reinforcement, millan2019human, harnack2023quantifying, yu2023thumbs}, \textit{scalar ratings} \cite{thomaz2007asymmetric, warnell2018deep, wilde2021learning, yu2023thumbs, kreutzer2018neural, liu2018dialogue}, \textit{preferences} \cite{christiano2017deep, wirth2017survey, lee2021b, askell2021general, bai2022constitutionalaiharmlessnessai}, etc. \cite{arzate_cruz_survey_2020}},
    attrtwo={Instruct},
    attrtwocontent={Instructive feedback can be given via \textit{action advice} \cite{frazier2019improving, Amir2016, krening2018newtonian}, \textit{demonstrations} \cite{taylor2011integrating, ng2000algorithms}, \textit{physical} \cite{losey2022physical, mehta2024strol, chetouani2021interactive}or \textit{verbal corrections} \cite{liu2023interactive}},
    attrthree={Describe},
    attrthreecontent={Descriptive feedback can be in the form of \textit{rules} \cite{, maclin1996creating, kuhlmann2004guiding, bignold2021persistent} or \textit{principles}~\cite{bai2022constitutionalaiharmlessnessai}, \textit{constraints} \cite{hayes2013advice, de2019accelerating, wilde2020improving, sharma2022correcting}, \textit{subgoal generation} \cite{okudo2021subgoal, xu2023dexterous}, among others \cite{sumers_linguistic_2022}.},
    attrfour={None},
    attrfourcontent={Feedback can be given without intention, by \textit{facial expressions} \cite{arakawa2018dqn, Wang2019interactive, cui2021empathic, li2020facial}, \textit{error} \textit{potentials in the brain} \cite{kim2017intrinsic, xu2021accelerating}, \textit{gaze} \cite{zhang2019leveraging, aronson2022gaze, zhang2020atari} and other subconscious reactions.}
}
While there is ample space for possible intentions in general, we focus on four attributes that are exhaustive in the human-AI context: \humanul{evaluating} the agent, i.e., assessing how well it is doing, \humanul{instructing} the agent, i.e., communicating a desired behavior or outcome, \humanul{describing}, i.e., providing supplementary information to complement other feedback. Finally, feedback can be provided \humanul{without} a clear interaction intention. 

In existing work for LLMs, simple evaluative feedback (in particular preferences) are the dominant ways a human can express their intent~\cite{fernandes2023bridging}. However, we can imagine interactions for instructive feedback (e.g., providing a fragment of desired or corrected text output to the model). 
In the case of our household robot, we might want to evaluate the observed behavior (\textit{"I did or did not like the way the robot cleaned some furniture"}), instruct it (\textit{"Clean it using a different type of cleaner"}), give high-level descriptive feedback (\textit{"These type of wooden surfaces should always be cleaned with a special cleaner"}). We might also provide feedback unintentionally (e.g., by grimacing when watching the robot cleaning incorrectly, but, e.g., when manually cleaning something after the robot is done without a clear intention to teach the robot).

\paragraph{Formal Definition: Intent}
Recall the basic definition of the reward function $r: (s, a) \rightarrow r$.  We can interpret feedback of different intentions to update our knowledge about the reward estimate $r$:
$$
m \xrightarrow[]{\phi} \begin{cases}
(\mathcal{T}, r_{fb}) \quad (evaluative)\\
(\mathcal{T}:[(s^0,a_H^1),...], r_{fb} = (0,1] ) \quad(instructive)\\
(\mathcal{T} \subseteq [(s_0,a_1),...], r_{fb}) \quad(descriptive)\\
\{(\mathcal{T}, r_{fb}), \mathcal{C} \} \quad(no~intention)\\
\end{cases}
$$
Here, we might assume that instructive feedback is optimal $(r_{fb} = 1)$ or has a certain optimality, e.g., confidence, assigned $(r_{fb} < 1)$. Feedback from other types, such as evaluative, can also contribute to the context $\mathcal{C}$; we omit it here for simplicity.
For unintentional feedback, we receive a measurement $m$, which needs to be translated into a well-defined value range.\\

\dimshaded{
    title={Expression Form \refhuman{D2}},
    color={HumanColor},
    definition={\label{human:D2}Whether the human provides the feedback deliberately.},
    attrone={Explicit},
    attronecontent={Feedback can be directly articulated as numeric or ordinal values, via \textit{buttons} \cite{knox2012reinforcement, li2013using, christiano2017deep}, \textit{sliders} \cite{thomaz2005real, cakmak2011human, macglashan2017interactive, wilde2021learning} and can be directly interpreted.},
    attrtwo={Implicit},
    attrtwocontent={Involves more subtle forms of expression, such as \textit{gestures} \cite{cruz2018multi, trick2022interactive, lukowicz2023sa}, \textit{natural language} \cite{sumers_linguistic_2022, lin2022inferring, lukowicz2023sa, wu2023fine, liu2023interactive} \textit{movement} \cite{wilson2023drone, chetouani2021interactive} or \textit{physiological signals} \cite{kim2017intrinsic, xu2021accelerating}},
    %attrthree={},
    %attrthreecontent={},
    %attrfour={},
    %attrfourcontent={}
}
Feedback can be communicated either \humanul{explicitly}, i.e., via a deliberate action by a human, or \humanul{implicitly}, i.e., through indirect interactions \cite{Li2021}. The contrast between explicit and implicit feedback has already been established in a series of previous works~\cite{lin2020review, arzate_cruz_survey_2020, lindner_humans_2022}. Explicit feedback is generated by direct interaction, e.g., pressing a button. Implicit feedback includes, e.g., gestures or mimics. Implicit feedback can but does not have to be unconscious. Unintended feedback is related to the concept of information leakage~\cite{jeon_reward-rational_2020}, i.e., information that can be inferred from observing human actors without direct interaction.

When interacting with our robot, we could, e.g., have an app that lets us rate the observed behavior of the robot explicitly (\textit{"How would I rate the performance of the robot on a scale from 1 to 5?}), but we could also choose implicit communication channels, like verbal feedback ("I was not satisfied with the behavior"). Unintentional feedback is, by definition, also a type of implicit feedback and can be a type of leaked information as described above (e.g., correcting the execution of the robot).

\paragraph{Formal Definition: Expression}
We have introduced the translation function $\phi: m \rightarrow \mathcal{F}$. For explicit feedback, we receive a measurement $\hat{m}$, containing measurement variables that can be directly translated into usable feedback, i.e., via a deterministic, simple translation function $\phi$. For implicit feedback, we require an intermediate mechanism to process the measurement into a simplified measurement via a (statistical) model to interpret the measurement as a reward value.
$$
\phi: m \rightarrow \mathcal{F} \quad = \begin{cases}
\phi(m) = m \quad(explicit)\\
\phi_{\theta}(m) \neq m \quad(implicit)\\
\end{cases}
$$
The reward model itself, in particular, if coupled with (deep) representation learning, can potentially map a measurement to a reward value directly.\\

\dimshaded{
    title={Engagement \refhuman{D3}},
    color={HumanColor},
    definition={\label{human:D3}How the human is engaged when giving the feedback},
    attrone={Proactive},
    attronecontent={Feedback is offered proactively by the user, without direct solicitation, via selection or manual intervention \cite{maclin1996creating, tenorio2010dynamic, newman2023towards, li2021learning}.},
    attrtwo={Reactive},
    attrtwocontent={Feedback is provided in response to a query, initiated by the system or agent \cite{mindermann2018active, fitzgerald2023inquire, zhang2023good}.},
    %attrthree={},
    %attrthreecontent={},
    %attrfour={},
    %attrfourcontent={}
}
Feedback can be provided \humanul{proactively}, i.e., when the human decides to give feedback for a specific state, action, or feature out of a potentially large set of other states. On the other hand, feedback can be provided \humanul{reactively}, i.e., when specifically requested or being queried by the system. Proactive feedback can include giving it only at specific points in time, but otherwise, not giving any feedback towards agent behavior. On the other hand, providing feedback on request also includes standard labeling of given samples, e.g., by crowd workers. One significant implication of user engagement is the interpretation of non-feedback. When not being able to give feedback, non-engagement can be interpreted as silent agreement, and in turn, providing feedback must be weighted appropriately.

When interacting with generative AI systems, humans may be able to give optional feedback, i.e., \textit{thumbs up}/\textit{thumbs down}, for a response to a query. When interpreting this feedback, we should consider how often a user decides to use this feedback, as it indicates either a very positive or very negative response to a query, compared to feedback received due to a direct query.
When observing the household robot, we might either choose to pro-actively engage with the robot (\textit{"Please stop the current action and use the vacuum cleaner first"}) or reactively, e.g., by being queried for feedback (\textit{"Please rate the robot's behavior on a scale from 1 to 5")}. 

\paragraph{Formal Definition: Engagement} For reactive/queried feedback, we can assume that feedback is sampled according to the known target distribution $\mathsf{T}$. For proactive feedback, we cannot assume this fact. Instead, we assume that human feedback is a result of a human target distribution $\mathsf{T}_H$. Considering this fact can be important when updating the reward model. Also, this type of "information leakage" might help us more accurately model the human's internal objectives.

$$
\mathcal{T} \sim \quad \begin{cases}
\mathsf{T} \quad(reactive)\\
\mathsf{T}_H \neq \mathsf{T} \quad(proactive)\\
\end{cases}
$$

\subsection{Interface-Centered Dimensions}
\label{subsec:structural-dimensions}
The following dimensions are relevant when processing human feedback as they determine which information can be encoded. The feedback can contain different \textbf{relations} between the selected targets, it can be given at different \textbf{levels of content}, and finally, it can contain information about an \textbf{actual} or a non-existing target.\\

\dimshaded{
    title={Target Relation \reffbp{D4}},
    color={CombinedColor},
    definition={\label{fbp:D4}The relation of the target of feedback to other targets.},
    attrone={Absolute},
    attronecontent={Feedback that is not comparative, focusing on a single action or behavior at a time, like \textit{binary critique} \cite{knox2012reinforcement, millan2019human}, \textit{scalar feedback} \cite{thomaz2007asymmetric, warnell2018deep}, or \textit{demonstrations} \cite{taylor2011integrating, ng2000algorithms}.},
    attrtwo={Relative},
    attrtwocontent={Feedback like \textit{preferences} \cite{christiano2017deep, wirth2017survey},\textit{rankings} \cite{myers2022learning, Zhu2023}, or \textit{corrections} \cite{losey2022physical, liu2023interactive, tandon2022learning, elgohary2021nl}, comparing multiple actions or behaviors to guide choice or improvement.},
    %attrthree={},
    %attrthreecontent={},
    %attrfour={},
    %attrfourcontent={}
}
Feedback can target a single unit, i.e., a state, episode, or input feature, which is an \combul{absolute} way, i.e., standing on its own. For example, when we say if a state is good or bad, the relationship to other states is only given implicitly. Alternatively, we can give feedback on relations between units, e.g., comparisons or corrections. In these cases, the target relation is \combul{relative}. In sequential settings, where we observe or interact with multiple instances after another, we could consider all feedback relative to previous instances. However, for our taxonomy, we only consider explicitly contrasted examples in an immediate shared context as relative feedback.

When interacting with the robot, we might comment on a single action (\textit{"The way you handled the duster was not good, you could have damaged the delicate glasses"}), but potentially also in relation to another action or state ("The way you used the duster was way better compared to the previous time"). The same preferences can also be stated for states or features analogous to absolute feedback (\textit{"I like the placement of pillows much better than the last time"}, \textit{"In general, I like blue pillows more than green ones"}).

\paragraph{Formal Definition: Target Relation} Absolute feedback contains information about a single target associated with a scalar feedback value, whereas relative feedback contains information about multiple targets, with the feedback value being a relation like inequality or order.

$$
(\mathcal{T}, r_{fb}) \quad \text{with} \begin{cases}
|\mathcal{T}| = 1, r_{fb} \in {\mathbb{N}, \mathbb{R}, \{0,1,-1\}} \quad(absolute)\\
|\mathcal{T}| > 1, r_{fb} \in \{\succ, \prec, =\}  \quad(relative)\\
\end{cases}
$$
with $|\cdot|$ meaning the cardinality of a set.\\

\dimshaded{
    title={Content Level \reffbp{D5}},
    color={CombinedColor},
    definition={\label{model:fbp5}The levels of abstraction the feedback is given at.},
    attrone={Instance},
    attronecontent={Specific feedback related to particular instances or actions, such as \textit{critique} \cite{knox2012reinforcement, millan2019human}, \textit{preferences} \cite{christiano2017deep, wirth2017survey} or \textit{action advice/demonstrations} \cite{ng2000algorithms, frazier2019improving}.},
    attrtwo={Feature},
    attrtwocontent={Targeting detailed aspects of behavior, e.g, \textit{feature preferences} \cite{shek2023learning, basu2018learning}, \textit{feature selection} \cite{zhang2019leveraging, aronson2022gaze}, \textit{constraints} \cite{hayes2013advice, wilde2020improving}, or \textit{subgoal generation} \cite{okudo2021subgoal}.},
    attrthree={Meta-Level},
    attrthreecontent={Feedback revealing contextual influence factors, e.g., distinguishability queries~\cite{feng2024comparing}, skill assessment questions~\cite{singhal2024scalable}.},
    %attrfour={},
    %attrfourcontent={}
}
Feedback can be targeted at specific \modelul{instances} of behavior, e.g., environment states the agent encountered, or at more abstract \modelul{features} of the behavior. For example, a human actor might label a specific action as "good" or "bad", or comment on a particular feature of the action being desirable.
Finally, we may receive \modelul{meta-level} feedback, which can be information that cannot be directly translated into a reward signal but instead reveals information about the feedback context and can, therefore, be used to improve the contextual translation and modeling of feedback.

In our running example, feedback can be given on an instance level (\textit{"I rate this particular action with a score from 1 to 5"}) or on a feature level (\textit{"I like to use the yellow table cloth in the summer"}). Meta-level feedback could be a statement like \textit{My eyesight is bad, so I might have trouble distinguishing small objects from each other}). Finally, measurements such as thinking time, etc., can be interpreted as meta-feedback, which indicates, e.g., the knowledge state of the human.

\paragraph{Formal Definition: Content Level} 
For instance-level feedback, we assume that the target contains whole states or actions. For feature-level feedback, a target is a set of features out of the state or action space that generally do not correspond to single states or actions. Meta-level feedback is translated to contextual information.
$$
\phi(m) = \quad \begin{cases}
(\mathcal{T}: \{[(s^0,a^1),...] | s^k=(s_1,...s_N), a^k=(a_0,...,a_M)\}, r_{fb}) \quad(instance)\\
(\mathcal{T}: \{s_i,a_j | s_i \subset s_i \subseteq \mathcal{S}, a_i \subseteq \mathcal{A}\}, r_{fb}) \quad(feature)\\
\mathcal{C} \quad(meta)\\
\end{cases}
$$
%\shaded{Target Actuality \reffbp{D6}}{CombinedColor}{\label{fbp:D6}Whether the feedback is targeted at something the human has observed.}{\textbf{Observed} (states from AI agent rollouts) | \textbf{Hypothetical} (not observed in the data, generated from human imagination, counterfactuals)}

\dimshaded{
    title={Target Actuality \reffbp{D6}},
    color={CombinedColor},
    definition={\label{fbp:D6}Whether the feedback is targeted at something the human has observed.},
    attrone={Observed},
    attronecontent={Feedback or instructions are based on actual observed behavior or actions, such as \textit{critique} \cite{knox2012reinforcement, millan2019human}, \textit{preferences} \cite{christiano2017deep, wirth2017survey} or \textit{corrections} \cite{losey2022physical, liu2023interactive}.},
    attrthree={Hypothetical},
    attrthreecontent={\quad Feedback for scenarios that haven't occurred but are conceivable, such as \textit{demonstrations} \cite{taylor2011integrating}, \textit{generalizable rules} \cite{kuhlmann2004guiding, bignold2021persistent}, or \textit{imagined examples} \cite{yeh2018bridging, okudo2021subgoal}.},
    %attrfour={},
    %attrfourcontent={}
}
When giving feedback, a human actor needs to choose a target to which the given feedback is directed. The target can be something the human \combul{observed}, e.g., a step or episode of agent behavior. Alternatively, the target can be \combul{hypothetical}, e.g., when imagining a preferable sequence of actions an agent should have taken or a goal state an agent should have reached. Feedback can also be targeted at the training process or overall workflow, e.g., expressions of fatigue or commenting on an interaction mechanic. In a later section, we comment on using such meta feedback, e.g., for user modeling. In the context of observation actuality, we consider such feedback to have an observed target.

Our feedback can target a behavior observed from robot behavior (\textit{"You made the carpet wet when placing the mop over it"}) or be based on a hypothetical or even counterfactual scenario (\textit{"You should walk around it to avoid dripping water"}). Descriptive feedback might both target observed and hypothetical scenarios (\textit{"In case you have mop in your hand, avoid walking over the carpet"}). Given the language model use case, observed feedback is targeted at the generations of the language model, e.g., rating or correcting a text fragment generated by the LLM in response to a prompt. Hypothetical feedback can incorporate human-written answers to a prompt, as well as more high-level rules, such as in \textit{constitutional AI}~\cite{bai2022constitutionalaiharmlessnessai}, that are not directly related to the immediate generations of an LLM-based agent.

\paragraph{Formal Definition: Target Actuality} 
As the name suggests, we are interested in the origin of feedback targets; we can assume that observed feedback is \textit{in-distribution} with respect to the agent's policy, i.e., it is part of rollouts that result from following an agent's policy during training. In contrast, hypothetical feedback cannot be assumed to be in-distribution. Moreover, hypothetical feedback might be completely independent of a specific policy and instead describe general rules and preferences. 
$$
\mathcal{T} \sim \quad \begin{cases}
\mathbb{E}_{\mathcal{T} \sim \pi}[(s^0,a^0),...] \quad(observed)\\
\{s_i,a_j | s_i \in s, a_j \in a\}, \Xi \quad(hypothetical)\\
\end{cases}
$$
Observed feedback is sampled from policy rollouts, whereas hypothetical feedback can incorporate all trajectories from the trajectory space $\Xi$ and all possible features from the state and action spaces.

\subsection{Model-Centered Dimensions}
\label{subsec:model-centered-dimensions}
The model-centered dimensions directly influence the design of the reward model, for example, for which \textbf{temporal granularity} the feedback is available, which \textbf{value granularity} it has, or if it is \textbf{exclusive} or given alongside other feedback.\\

%\subsubsection{Temporal Granularity} 
%\shaded{Temporal Granularity \refmodel{D7}}{ModelColor}{\label{model:D7}The temporal granularity of states, actions, or features the feedback is given at.}{\textbf{State} | \textbf{Segment} | \textbf{Episode} | \textbf{Entire Behavior}}
\dimshaded{
    title={Temporal Granularity \refmodel{D7}},
    color={ModelColor},
    definition={\label{model:D7}The temporal granularity of states, actions, or features the feedback is given at.},
    attrone={State},
    attronecontent={Feedback can be given for single states or state-action pairs \cite{judah2010reinforcement, Amir2016, frazier2019improving, sheidlower2022keeping}.},
    attrtwo={Segment},
    attrtwocontent={Feedback is also often assigned to segments of various length \cite{christiano2017deep, arakawa2018dqn, arzate2020mariomix, torne2023breadcrumbs}.},
    attrthree={Episode},
    attrthreecontent={Here, the reward is given to entire episodes in episodic RL tasks},
    attrfour={Entire Behavior},
    attrfourcontent={Feedback might be directed at overall behavior, not at a specific state or action \cite{mindermann2018active, wilde2020improving, lin2022inferring}.}
}
Feedback can have targets at different levels of granularity. Very granular feedback can be given about \modelul{individual states, actions, or features}, e.g., if a human recommends an alternative action in a particular state. Less granular is the typical scenario where humans rate segments or episodes. We can give even less granular feedback, e.g., about an agent's (current) \modelul{overall behavior}. Importantly, this granularity is a spectrum, and many different levels between individual actions and the agent's overall behavior are possible.

For generated text, we may give feedback for single words or even tokens~\cite{wu2023fine}, sentences and segments (e.g., single reasoning steps, phrases in a generated poem, etc.), as well as entire task executions (e.g., whether a LLM-based agent has solved a complex task correctly). When communicating feedback to our household robot, we could either comment on single actions (\textit{"I mark the correctness of actions in this sequence via the app"}), on a segment level (\textit{"The way you folded this was great"}), per episode if we define it as an episodic task (an episode might be an entire task) (\textit{"Your performance today folding the clothes was better than last week overall"}). Finally, there are different ways to give feedback not directed at particular states (\textit{"You should not fold shirts, they go directly to the coathanger"},\textit{"I have no complaints about your work"}).

\paragraph{Formal Definition: Temporal Granularity} 
Temporal granularity influences the type of targets available for a feedback instance, i.e., the set of targets to which a feedback value is assigned.
$$
\mathcal{T} \in \quad \begin{cases}
(s,a) \quad(state)\\
[(s_k,a_k),(s_{k+1},a_{k+1}),...(s_{k+M},a_{k+M})] \quad(segment)\\
[(s_0,a_0),(s_1,a_1),...(s_T,a_T)] \quad(episode)\\
\{s_i,a_j | s_i \in s, a_j \in a\}, \Xi \quad\text{(entire behavior)}\\
\end{cases}
$$
with $M$ being the length of a segment, $s_0$ being a start state and $T$ being a terminal state. The first three cases can be easily extended to sets of either single states/state-action pairs, segments, or episodes (e.g., for relative feedback).\\
%\subsubsection{Value Granularity} 
%\shaded{Choice Set Size \refmodel{D8}}{ModelColor}{\label{model:D8}The value granularity of states, actions, or features the feedback is given at.}{\textbf{Binary} | \textbf{Discrete} | \textbf{Infinite (Continuous space)}}

\dimshaded{
    title={Choice Set Size \refmodel{D8}},
    color={ModelColor},
    definition={\label{model:D8}The value granularity of states, actions, or features the feedback is given at.},
    attrone={Binary},
    attronecontent={Simple feedback indicating a binary choice, like good/bad \cite{knox2009interactively, sheidlower2022keeping}, binary preferences \cite{christiano2017deep, wirth2017survey}, stop execution \cite{jeon_reward-rational_2020}, hard constraints or rules in first-order logic \cite{kuhlmann2004guiding, de2019accelerating}.},
    attrtwo={Discrete},
    attrtwocontent={Feedback can be given on whole number (Likert) scales \cite{gao2018april}, ranking of k elements \cite{myers2022learning, shrestha2022fairfuse}, choosing the best action out of a set \cite{frazier2019improving}.},
    attrthree={Infinite},
    attrthreecontent={Feedback that exists on a continuous scale, such as demonstrative and corrective feedback in continuous action spaces or implicit feedback (e.g., gaze or physiological signals).},
    %attrfour={Entire Behavior},
    %attrfourcontent={Feedback might be directed at overall behavior, not at a specific state or action \cite{mindermann2018active, wilde2020improving, lin2022inferring}.}
}
Feedback can have different value granularities determined by the interface that the user is provided with. For example, giving a thumbs up or down is interpreted as binary feedback. Similarly, an intervention can also be interpreted as a binary value (either intervening or not intervening). Similarly, a pairwise comparison is a type of binary-valued feedback. Additionally, we can interpret certain types of descriptive feedback, like hard constraints, as binary feedback. Feedback from a multi-step ranking or on a multi-point scale can be interpreted as discrete/integer-valued feedback. Continuous feedback might either be based on a very fine-grained rating scale or, more often, come from implicit modalities, e.g., physical corrections of a robot, or feedback based on human facial expressions might be available on a continuous scale.

When interacting with commercial LLM-based chat systems, we are often restricted to binary feedback, such as "thumbs up"/"thumbs down" or the choice between two alternative generations. However, we might extend these choices, e.g., selecting words to generate out of a set of multiple possible choices~\cite{spinner2024generaitor}, or freely generating desired output, which in practice coincides with a virtually infinite choice set.
When giving binary feedback to our robot, we give a positive or negative signal (\textit{"Thumbs up/down for this"}), binary preference (\textit{"I like this placement better"}), or fixed rule (\textit{"The remote should always be put on the armrest"}). Integer-valued feedback could be on a scale ("I rate this execution on a 5 out of 10") or choose out a set of alternatives (\textit{"I like this arrangement best, this arrangement second best, and the third one the least"}). Finally, continuous feedback can correspond to a target out of an infinite set, e.g., when applying a physical force to correct the movements of a robot. Also, we might choose from a "virtually" infinite space of possible alternative state-action trajectories when generating a demonstration.

\paragraph{Formal Definition: Choice Set Size} 
The choice set size indicates how many different options for a measured variable exist, i.e., from how many options a human can choose a feedback value.
$$
r_{fb} \in \quad \begin{cases}
{0,1,\succ, \prec} \quad(binary)\\
\mathbb{N} \quad(discrete)\\
\mathbb{R} \quad(infinte)\\
\end{cases}
$$
%\subsubsection{Exclusivity} 
%\shaded{Feedback Exclusivity \refmodel{D9}}{ModelColor}{\label{model:D9} Whether the utterance is the exclusive source of feedback or available besides other sources.}{\textbf{Single} (Only reward source) | \textbf{Mixed} (Alongside other sources)}

\dimshaded{
    title={Feedback Exclusivity \refmodel{D9}},
    color={ModelColor},
    definition={\label{model:D9} Whether the utterance is the exclusive source of feedback or available besides other sources.},
    attrone={Single},
    attronecontent={Scenarios in which a type of human feedback is the sole source of a learning signal, e.g. in RLHF fine-tuning for LLMs \cite{ouyang2022training, Ziegler2019finetuning}.},
    attrtwo={Mixed},
    attrtwocontent={Feedback that supplements other sources of feedback,  e.g., multi-faceted feedback~\cite{wu2023fine}, shaping rewards \cite{knox2010combining} or simultaneous multi-modal feedback \cite{fitzgerald2023inquire, li2020facial}.},
    %attrthree={Infinite},
    %attrthreecontent={Feedback that exists on a continuous scale, such as demonstrative and corrective feedback in continuous action spaces or implicit feedback (e.g., gaze or physiological signals).},
    %attrfour={Entire Behavior},
    %attrfourcontent={Feedback might be directed at overall behavior, not at a specific state or action \cite{mindermann2018active, wilde2020improving, lin2022inferring}.}
}
When providing feedback, human feedback can serve as the exclusive single source of reward in a situation, e.g., when tuning an agent with personal preferences. However, feedback for the same target might also be given from multiple sources. These other sources can be environment reward signals (like in reward shaping), feedback from other human users for the same target, or even other AI models. Finally, feedback might also be provided by the same user via a different feedback channel, e.g., an additional feedback type.

When not provided with another reward source, the household robot purely relies on personal feedback to optimize its behavior. A different source can be other humans, e.g., other persons living in the household with potentially differing opinions (\textit{"I want the book to be put on a different shelf")}, but also environment signals (\textit{e.g., not damaging objects could be an objective irrespective of human supervision}) or even other AI models (\textit{Besides the human feedback, the robot also uses a cloud-based reward model to update its policy}).

\paragraph{Formal Definition: Exclusivity} 
In our framework, the non-exclusivity of feedback means that multiple feedback values can be assigned to the same target, not caused by variance in repeated measurement but by expressing an actual underlying difference in feedback value.
$$
\exists \mathcal{T} \quad \begin{cases}
\exists! \hat{m}_A \quad (exclusive)\\
\exists \{\hat{m}_A, \hat{m}_B, ...\} \ \quad (mixed)
\end{cases}
\quad \text{with $A,B,...$ different sources} $$
So $\hat{m}_A, \hat{m}_B$ are measurements from different sources which can be potentially different, i.e., measurements from different sources can provide different values for the same target. We want to differentiate this from repeated measurements from the same source, which can be different (e.g., by changing preferences throughout training). We will comment on this  in~\autoref{sec:qualities}.

% Established feedback table
\begin{table*}[tbp]
    \centering
    \caption{Classifying Established Feedback Types: We structure feedback along the nine dimensions we introduce in~\autoref{subsec:human-centered-dimensions} to \autoref{subsec:model-centered-dimensions}. (\OK) Black checkmarks refer to exclusive attributes (\OKGREY). Grey checkmarks indicate that feedback can have characteristics across different work. We can classify existing types of feedback according to our dimensions.}
    \definecolor{blueish}{HTML}{F4F7F8}
    \rowcolors{2}{white}{blueish}
    \setlength{\tabcolsep}{1.5pt}
    \renewcommand\arraystretch{1.20}
    \resizebox{\textwidth}{!}{%
    \begin{tabular}{r | *{4}{c} | *{2}{c} | *{2}{c} | *{2}{c} | *{2}{c} | *{2}{c} | *{4}{c} | *{3}{c} | *{2}{c} | *{1}{c} }
        & \multicolumn{4}{|c|}{\textcolor{DarkHumanColor}{\textbf{D1}}} 
        & \multicolumn{2}{c|}{\textcolor{DarkHumanColor}{\textbf{D2}}} 
        & \multicolumn{2}{c|}{\textcolor{DarkHumanColor}{\textbf{D3}}} 
        & \multicolumn{2}{c|}{\textcolor{DarkCombinedColor}{\textbf{D4}}} 
        & \multicolumn{2}{c|}{\textcolor{DarkCombinedColor}{\textbf{D5}}} 
        & \multicolumn{2}{c|}{\textcolor{DarkCombinedColor}{\textbf{D6}}}
        & \multicolumn{4}{c|}{\textcolor{DarkModelColor}{\textbf{D7}}}
        & \multicolumn{3}{c|}{\textcolor{DarkModelColor}{\textbf{D8}}}
        & \multicolumn{2}{c|}{\textcolor{DarkModelColor}{\textbf{D9}}}\\
        & \multicolumn{4}{|c|}{\textcolor{DarkHumanColor}{\textbf{Intent}}} 
        & \multicolumn{2}{c|}{\textcolor{DarkHumanColor}{\textbf{Expres.}}} 
        & \multicolumn{2}{c|}{\textcolor{DarkHumanColor}{\textbf{Engag.}}} 
        & \multicolumn{2}{c|}{\textcolor{DarkCombinedColor}{\textbf{Rela.}}} 
        & \multicolumn{2}{c|}{\textcolor{DarkCombinedColor}{\textbf{Content}}} 
        & \multicolumn{2}{c|}{\textcolor{DarkCombinedColor}{\textbf{Tgt.Act.}}} 
        & \multicolumn{4}{c|}{\textcolor{DarkModelColor}{\textbf{Temp.Gra.}}} 
        & \multicolumn{3}{c|}{\textcolor{DarkModelColor}{\textbf{Choi.Set}}} 
        & \multicolumn{2}{c|}{\textcolor{DarkModelColor}{\textbf{Exclus.}}} 
        &\\
        \cmidrule{2-24}
        Feedback Type 
        & \rot{Evaluate} & \rot{Instruct} & \rot{Describe} & \rot{None} 
        & \rot{Explicit} & \rot{Implicit} 
        & \rot{Proactive} & \rot{Reactive} 
        & \rot{Absolute} & \rot{Relative} 
        & \rot{Instance} & \rot{Feature} 
        & \rot{Actual} & \rot{Hypothetical} 
        & \rot{Step/Action} & \rot{Segment} & \rot{Episode} & \rot{Ent.Beh.}
        & \rot{Binary} & \rot{Discrete} & \rot{Continuous} 
        & \rot{Exclusive} & \rot{Augmenting} & Publications \\
        
        \midrule
        Critique &
        % Intent
        \OK & & & &
        % Expression
        \OK & &
        % Engagement
        \OKGREY & \OKGREY &
        % Target Relation
        \OK & &
        % Content Level
        \OK & &
        % Target Actuality
        \OK & &
        % Temp. Granularity
        \OKGREY & \OKGREY & & &
        % Value Granularity
        \OK & \OK & &
        % Exclusivity
        \OKGREY& \OKGREY &
        % Publication
        \cite{guan2021widening}\\
        
        Shaping &
        % Intent
        \OK & & & &
        % Expression
        \OKGREY & \OKGREY &
        % Engagement
        \OKGREY & \OKGREY &
        % Target Relation
        \OK & &
        % Content Level
        \OK & &
        % Target Actuality
        \OK & &
        % Temp. Granularity
        \OKGREY & \OKGREY & \OKGREY & &
        % Value Granularity
        \OKGREY & \OKGREY & \OKGREY &
        % Exclusivity
        & \OK &
        % Publication
        \cite{knox2009interactively, warnell_deep_2017, chuang2021using} \\

        Behavior Pref. &
        % Intent
        \OK & & & &
        % Expression
        \OK & &
        % Engagement
        \OKGREY & \OKGREY &
        % Target Relation
        & \OK &
        % Content Level
        \OK & &
        % Target Actuality
        \OK & &
        % Temp. Granularity
        & \OKGREY & \OKGREY & &
        % Value Granularity
        \OKGREY & \OKGREY & &
        % Exclusivity
        \OK & &
        % Publication
        \cite{christiano2017deep, brown_extrapolating_2019, ibarz2018reward} \\

        Outcome Rating &
        % Intent
        \OK & & & &
        % Expression
        \OK & &
        % Engagement
        & \OK &
        % Target Relation
        \OK & &
        % Content Level
        \OK & &
        % Target Actuality
        \OK & &
        % Temp. Granularity
        & & \OK & &
        % Value Granularity
        \OKGREY & \OKGREY & &
        % Exclusivity
        \OKGREY & \OKGREY &
        % Publication
        \cite{Pahic2018, THOMAZ2008716teachablerobots} \\

        \midrule

        Action Advice &
        % Intent
        & \OK & & &
        % Expression
        \OK & &
        % Engagement
        & \OK &
        % Target Relation
        \OK & &
        % Content Level
        \OK & &
        % Target Actuality
        \OK & &
        % Temp. Granularity
        \OK & & & &
        % Value Granularity
        \OK & & &
        % Exclusivity
        & \OK &
        % Publication
        \cite{frazier2019improving, da2020uncertainty, Sharath2019} \\
        
        Demos &
        % Intent
         & \OK & & &
        % Expression
        \OK & &
        % Engagement
        \OK & &
        % Target Relation
        \OK & &
        % Content Level
        \OK & &
        % Target Actuality
        & \OK &
        % Temp. Granularity
        \OK & & & &
        % Value Granularity
        \OKGREY & & \OKGREY &
        % Exclusivity
        \OKGREY & \OKGREY &
        % Publication
        \cite{ng2000algorithms, Puri2020Explain, ibarz2018reward} \\

        Demos w/o acts. &
        % Intent
         & \OK & & &
        % Expression
        & \OK &
        % Engagement
        \OK & &
        % Target Relation
        \OK & &
        % Content Level
        \OK & &
        % Target Actuality
        & \OK &
        % Temp. Granularity
        \OKGREY & \OKGREY & & &
        % Value Granularity
        & \OK & &
        % Exclusivity
        \OKGREY & \OKGREY &
        % Publication
        \cite{liu2018imitation, Torabi2018}\\

        Correction &
        % Intent
         & \OK & & &
        % Expression
        \OKGREY & \OKGREY &
        % Engagement
        \OK & &
        % Target Relation
        & \OK &
        % Content Level
        \OK & &
        % Target Actuality
        \OK & &
        % Temp. Granularity
        & \OKGREY & \OKGREY & &
        % Value Granularity
        & \OKGREY & \OKGREY &
        % Exclusivity
        & \OK &
        % Publication
        \cite{mehta2022unified, sharma2022correcting, losey2022physical} \\
        
        \midrule

        Feat. Selection &
        % Intent
         & & \OK & &
        % Expression
        \OKGREY & \OKGREY &
        % Engagement
        \OKGREY & \OKGREY &
        % Target Relation
        \OK & &
        % Content Level
        & \OK &
        % Target Actuality
        \OK & &
        % Temp. Granularity
        \OKGREY & \OKGREY & & \OKGREY &
        % Value Granularity
        \OKGREY & \OKGREY & &
        % Exclusivity
        & \OK &
        % Publication
        \cite{sumers2021learning, THOMAZ2008716teachablerobots} \\

        Feat. Saliency &
        % Intent
         & & \OK & &
        % Expression
        \OKGREY & \OKGREY &
        % Engagement
        \OKGREY & \OKGREY &
        % Target Relation
        \OK & &
        % Content Level
        & \OK &
        % Target Actuality
        \OK & &
        % Temp. Granularity
        \OKGREY & \OKGREY & & &
        % Value Granularity
        \OK & & &
        % Exclusivity
        & \OK &
        % Publication
        \cite{guan2021widening} \\

        Goal Spec. &
        % Intent
         & & \OK & &
        % Expression
        \OKGREY & \OKGREY &
        % Engagement
        \OKGREY & \OKGREY &
        % Target Relation
        \OK & &
        % Content Level
        & \OK &
        % Target Actuality
        & \OK &
        % Temp. Granularity
        \OK & & & &
        % Value Granularity
        \OK & & &
        % Exclusivity
        & \OK &
        % Publication
        \cite{okudo2021subgoal, koert2020multi}\\

        Goal Pref. &
        % Intent
         & & \OK & &
        % Expression
        \OKGREY & \OKGREY &
        % Engagement
        \OKGREY & \OKGREY &
        % Target Relation
        & \OK &
        % Content Level
        & \OK &
        % Target Actuality
        & \OK &
        % Temp. Granularity
        \OK & & & &
        % Value Granularity
        \OKGREY & \OKGREY & &
        % Exclusivity
        & \OK &
        % Publication
        \cite{wirth2017survey}\\

        \midrule

        Gaze &
        % Intent
         & & & \OK &
        % Expression
        & \OK &
        % Engagement
        & \OK &
        % Target Relation
        \OK & &
        % Content Level
        \OK & &
        % Target Actuality
        \OK & &
        % Temp. Granularity
        & \OKGREY & \OKGREY & &
        % Value Granularity
        & \OK & &
        % Exclusivity
        & \OK &
        % Publication
        \cite{zhang2020atari} \\

        Reactions &
        % Intent
        & & & \OK &
        % Expression
        & \OK &
        % Engagement
        \OK & &
        % Target Relation
        \OK & &
        % Content Level
        \OK & &
        % Target Actuality
        \OK & &
        % Temp. Granularity
        & \OKGREY & \OKGREY & &
        % Value Granularity
        & \OK & &
        % Exclusivity
        & \OK &
        % Publication
        \cite{veeriah2016face} \\

        \midrule
        
        \bottomrule
    \end{tabular}
    }
    %\me{where did these types come from? are these the only ones?}}
    \label{tab:feedback_types}
    \vspace{-1.5em}
\end{table*}

\subsection{Classifying Established Feedback Types}
\label{subsec:types_of_feedback}
First, we want to relate the taxonomy to commonly used types of feedback. Here, we select feedback types that are either commonly used in practice~\cite{arzate_cruz_survey_2020, metz2023rlhf} or at least clearly described in theory~\cite{jeon_reward-rational_2020}.
We summarize the identified feedback types in \autoref{tab:feedback_types}. As mentioned above, in this work, we focus on feedback types for which algorithmic exploitation has been established in the literature to ensure applicability~\cite{jeon_reward-rational_2020, lindner_humans_2022, ibarz2018reward, sumers_how_2022}.

\paragraph{Quantitative Feedback} Humans can give agents numerical performance \humanul{evaluation}. Examples include a binary or scalar reward about an episode (sometimes called as \textit{critique}\cite{arzate_cruz_survey_2020}), or reward shaping, which relies on assigning individual scalar values to single steps or trajectory segments~\cite{knox2009interactively, ng1999policy}. It is expressed \humanul{explicitly}, i.e., via user interactions like buttons or sliders. We might call \humanul{implicit} feedback targeted at absolute observed instances as \textit{reactions}. In many cases, quantitative feedback is \textit{reactive}, so queried from a human. However, when combined with the interactive selection of targets, it can also be given \textit{proactively} \cite{zhang_time-efficient_2022}. This feedback type always targets \combul{observed} behavior and is not given at the \combul{instance} level (i.e.,  does not comment on the merit of particular features). It is \combul{absolute}, i.e., not referring to another target. In terms of \modelul{temporal granularity}, we can observe the use of step-wise, segment-wise, and episode-wise feedback, as the assignment of a quantitative value to a selected target is equivalent among granularities. Most often, we see \modelul{binary} or scale, i.e., \modelul{multiple} possible feedback values. Quantitative feedback can be used both \modelul{exclusively} and \modelul{non-exclusively}. 

\paragraph{Comparative Feedback} This type of feedback provides the agent with information about how its performance compares to a reference, such as an alternative execution, benchmark, or a human expert. As such, it is still evaluative. The most known example of comparative feedback is rankings~\cite{brown_extrapolating_2019} or pairwise comparisons~\cite{christiano2017deep}. Existing forms of comparative feedback are generally targeted at \combul{observed} instances, i.e., by contrasting the replay of agent behavior against one another. As such, this type of feedback represents \combul{relative target relations}. We often see these comparisons between sequence \modelul{segments or full episodes}. Rankings give a \modelul{discrete} choice set, while pairwise preferences present a \modelul{binary} choice. Preferences have been used as the \modelul{exclusive} source of feedback to train reward models.

\paragraph{Corrective Feedback} Corrective feedback provides agents with specific information about what they did wrong and how they can correct their behavior and is, therefore, instructive. A main component of this type of feedback is the necessity of generative human processing, i.e., the human needs to imagine and demonstrate or describe a \humanul{hypothetical} sequence of actions or a hypothetical preferred goal state. For example, when recommending an alternative action compared to the one taken by agents (often referred to as \textit{action advice}~\cite{frazier2019improving, lindner_humans_2022, lin2020review}), we must imagine a hypothetical/counterfactual future in which the alternative action leads to a more desired outcome. For corrections, we provide feedback with a \combul{relative target relation}, i.e., in direct reference to an \humanul{observed} instance. Corrective feedback is generally combined with other types of feedback, i.e., \modelul{non-exclusive}.

\paragraph{Demonstrative Feedback} Here, humans provide agents with information about the desired behavior by demonstrating the sequence of corrective actions in a particular context. Compared to corrective feedback, demonstrations are \combul{absolute} feedback. If provided \humanul{explicitly}, e.g., by directly inputting the sequence of required actions, we call this type of feedback \textit{demonstrations}. If only provided \humanul{implicitly}, e.g., by showing body movements to an observing embodied robot, we might call them \textit{demonstrations without actions}~\cite{liu2018imitation, Torabi2018}. Further, compared to corrective feedback, demonstrative feedback almost always has a higher \modelul{granularity}, e.g., an entire episode, compared to just a small number of steps.

\paragraph{Descriptive Feedback} is a weakly defined term in the literature~\cite{sumers_how_2022}. In our taxonomy, we relate it to feedback at the \modelul{feature level}, compared to all previous forms of feedback, which are primarily given at the \modelul{instance level}. Descriptive feedback generally refers to certain aspects of either the \humanul{observed} behavior or (\humanul{hypothetical}) outcomes. We do not explicitly distinguish between process-level descriptive feedback (e.g., a particular joint position of a robot at a certain point in time is undesirable) and outcome-level descriptive feedback (e.g., a goal state of the environment). Descriptive feedback is given as complementary \modelul{non-exclusive} feedback.

%\clearpage
\section{Qualities of Human Feedback}
\label{sec:qualities}
We have now established a conceptual framework for feedback types. So, having answered the question \textit{"What kind of human feedback is there?"}, we now want to look at the question \textit{"What makes for good human feedback?"}.\\
To tackle this question, we choose a similar approach to identify relevant dimensions. From the surveyed set of papers, we identify quality criteria discussed in the literature. We find a large set of different quality measures and raised issues about human feedback in interactive RL and RLHF~\autoref{tab:quality-terms}. We posit that the large variety of terms is because we can look at feedback qualities from three different view angles as outlined in~\autoref{subsec:problem_characterization}: We ask the question \textit{"What is high-quality feedback from ..?"}
\begin{itemize}
    \item a \humanul{human} perspective: How good is the human experience of giving feedback?
    \item an \combul{interface} perspective: How well can the feedback be collected and processed?
    \item a \modelul{model} perspective: How well can a model be optimized based on the feedback?
\end{itemize}
In the following, we present seven quality criteria that can be used to evaluate the quality of human feedback. Based on these three perspective, we group the quality metrics into three categories equivalently to dimensions: \humanul{human-centered}, \combul{interface-centered}, and \modelul{model-centered} feedback quality criteria.

\begin{table}[htbp]
\centering
\caption{Human feedback qualities used in the literature: We summarize a large set of quality measures from surveyed literature into seven main quality criteria to give a more condensed set of relevant quality criteria.}
\definecolor{blueish}{HTML}{F4F4F4}
\rowcolors{2}{white}{blueish}
\resizebox{\textwidth}{!}{%
\begin{tabular}{lllllll}
\humanul{\textbf{Expressiveness}} & \humanul{\textbf{Ease}} & \combul{\textbf{Definiteness}} & \combul{\textbf{Context Independence}} & \modelul{\textbf{Precision}} & \modelul{\textbf{Unbiasedness}} & \modelul{\textbf{Informativeness}} \\ \hline
Richness \cite{sumers2021learning, casper2023open} & Fatigue \cite{freedman2021choice, koppol2023interactive} & Uncertainty \cite{scherf2022learning} & \begin{tabular}[c]{@{}l@{}}Choice set \\ dependence\end{tabular} \cite{freedman2021choice} & Accuracy \cite{bignold2021evaluation, scherf2022learning, faulkner2020interactive, bignold2023human} & Reward bias \cite{bignold2021evaluation, knox2012reinforcement, thomaz2007asymmetric} & Information gain \cite{bignold2023human, koppol2023interactive} \\
Open-Endedness \cite{sumers2021learning, luketia2019} & Availability \cite{bignold2021evaluation} & Confidence \cite{freedman2021choice, scherf2022learning} & Policy dependence \cite{macglashan2017interactive} & Repeatability \cite{bignold2021evaluation} & Asymmetry \cite{thomaz2007asymmetric} & Stability \cite{mehta2024strol} \\
& Time requirements \cite{bignold2023human, kaufmann2023survey} & Security \cite{scherf2022learning} & Knowledge level \cite{bignold2021evaluation} & Consistency \cite{bignold2021evaluation, cruz2018multi, lin2017explore, scherf2022learning} & Cognitive bias \cite{bignold2021evaluation} & Performance \cite{koppol2023interactive} \\
& Latency \cite{bignold2021evaluation} & \begin{tabular}[c]{@{}l@{}}Decision problem \\ uncertainty\end{tabular} \cite{laidlaw2021uncertain} & Independence \cite{li2021learning}  & Variation \cite{bignold2021evaluation} & Concept drift \cite{bignold2021evaluation} & \begin{tabular}[c]{@{}l@{}}Retrospective \\ optimality\end{tabular} \cite{scherf2022learning} \\
& Response time \cite{scherf2022learning} & Understandability \cite{kaufmann2023survey} & Ambiguity \cite{Lopes2011}  & Reliability \cite{zhang2023self, kaufmann2023survey} & Scale difference \cite{lin2017explore} & Completeness \cite{faulkner2020interactive, bignold2021evaluation}  \\
& \begin{tabular}[c]{@{}l@{}}Budgetary \\ efficiency\end{tabular} \cite{fachantidis2017learning} & Verifiability \cite{bignold2023human, macglashan2017interactive} & Hidden Context \cite{siththaranjan2023distributional}  & Noise \cite{faulkner2020interactive, mehta2024strol, koppol2023interactive} &  &  \\
& Cognitive load \cite{kaufmann2023survey, koppol2023interactive} & \begin{tabular}[c]{@{}l@{}}Perception \\ quality\end{tabular} \cite{cruz2018multi}  &  & \begin{tabular}[c]{@{}l@{}}Mislabeling \\ probability\end{tabular} \cite{ibarz2018reward} &  &  \\ \hline
\end{tabular}%
}
\label{tab:quality-terms}
\end{table}
We discuss how these seven qualities can be evaluated. We also discuss how we can measure each quality, i.e., which metrics exist.

\subsection{Human-Centered Qualities}
From a human perspective, we identify two key desiderata with their associated qualities: feedback should be \textbf{expressive}, and it should require minimal \textbf{effort} to provide. These qualities directly relate to the human-centered dimensions of \humanul{intent} \refhuman{D1}, \humanul{expression form} \refhuman{D2}, and \humanul{engagement} \refhuman{D3}.\\

\shaded{Expressiveness \refhuman{Q1}}{HumanColor}{\label{human:Q1}How expressive is the feedback in communicating human intentions?}{\textbf{Low Expressiveness} (Limited, unable to express according to intentions) -- \textbf{High Expressiveness} (Flexible, able to express intentions)}
Expressiveness can be understood as \textit{adequacy} from the human perspective: Does the available feedback mechanism match user expectations? Is it appropriate for the task complexity? The perceived \textit{expressiveness} of feedback heavily depends on the user's current state and their intended feedback goals.\\
\paragraph{How to evaluate \textit{Expressiveness}?} \textit{Expressive} feedback must align with the user's \humanul{intention} in any given situation. Users should be able to provide open \humanul{implicit} feedback when needed. As a human-centered quality, expressiveness is evaluated through \textit{human-centered evaluation} methods \cite{iml_review_sperrle}.\\
\emph{Subcategories:}
\begin{itemize}[noitemsep,topsep=0pt,parsep=0pt,partopsep=0pt]
\item \textbf{Open-Endedness:} tasks may be either \emph{open-ended}, i.e., not having a clearly defined outcome (e.g., "Write a story..") or close-ended ("put the dirty dishes into the dishwasher"). Open-ended tasks are well served by expressive, diverse feedback to give users ways to express complex preferences. Simple, explicit feedback can be the right choice for close-ended, clearly defined tasks.
\item \textbf{Satisfaction:} Users can rate their satisfaction with the feedback's expressiveness during the interaction process~\cite{koppol2023interactive}.
\item \textbf{Chosen Feedback:} When multiple feedback types are available, users' choices can reveal their preferences~\cite{metz2023rlhf}. These choices often indicate feedback appropriateness, though care should be taken to minimize confounding factors to obtain meaningful preference data.
\item \textbf{Engagement Metrics:} Retention rates, frequency of additional user comments, and similar behavioral indicators can serve as proxy measures for perceived expressiveness~\cite{koppol2023interactive}.
\end{itemize}
\emph{Optimizing expressiveness:} To optimize expressiveness, users must be able to effectively communicate their \humanul{intent} \refhuman{D1} in any given state. We hypothesize that \humanul{instructive} or \humanul{descriptive} feedback will be perceived as more expressive, especially when agents exhibit low skill levels or the human has expert knowledge. Additionally, \humanul{open-ended}, \humanul{implicit} or \humanul{multi-modal} feedback \refhuman{D2} options may enhance the user's sense of expressiveness. Similarly, the opportunity for users to act \humanul{proactively} \refhuman{D3} for targets has the potential to increase perceived expressiveness. We will discuss further research opportunities in~\autoref{sec:conclusion}.\\

%\subsubsection{Cognitive Demand}
\shaded{Ease \refhuman{Q2}}{HumanColor}{\label{human:Q2}The level of cognitive effort and time required to provide feedback.}{\textbf{No Conscious Effort} (unconscious or reactive) -- \textbf{Low Effort} (simple and brief interactions) -- \textbf{High Effort} (requires concentration or creativity)}
Different feedback types impose varying levels of cognitive demand on humans. \humanul{Implicit} feedback typically requires minimal or no conscious effort, making it well-suited for extended sessions. In contrast, \humanul{explicit} feedback generally demands higher cognitive engagement, especially for demonstrations and descriptions. \humanul{Reactive} feedback, initiated by user queries, enhances speed and ease of use. \humanul{Pro-active} feedback, however, requires greater user attention and additional effort to identify appropriate feedback opportunities.

\paragraph{How to evaluate \textit{Ease}?} Ease is evaluated through \textit{human-centered approaches}. The field of human-computer interaction offers numerous experimental tools for measuring users' cognitive workload~\cite{kosch2023survey, iml_review_sperrle}.\\
\emph{Subcategories:}
\begin{itemize}[noitemsep,topsep=0pt,parsep=0pt,partopsep=0pt]
\item \textbf{Cognitive Load:} This can be assessed through surveys or ratings that measure experienced cognitive demand and exhaustion~\cite{kaufmann2023survey, koppol2023interactive}. Regular assessments during the interaction process can track exhaustion levels. Response time measurements can serve as proxy indicators for cognitive demand, particularly when analyzing how interaction times evolve throughout the process~\cite{scherf2022learning}.
\item \textbf{Temporal Demand:} This metric quantifies the time required to provide feedback~\cite{bignold2023human}. Temporal demand typically correlates with cognitive load~\cite{bignold2021evaluation} and has significant implications for budgetary efficiency~\cite{fachantidis2017learning}.
\item \textbf{Knowledge Level:} A key consideration is the level of knowledge expected to give feedback for a task. Tasks might be either highly \emph{domain-specific}, requiring rare expert knowledge, or be \emph{domain-general}, i.e., can be evaluated or even solved by large parts of the population. We must collect feedback from domain experts for domain-specific tasks, e.g., coding tasks. Consequently, we both must be efficient with feedback, i.e., focusing on highly informative feedback and giving domain experts ways to express their knowledge by providing interactions for instructive or descriptive feedback. For domain-general tasks, simplicity of interactions and cost-effectiveness might be more important considerations.
\end{itemize}
\emph{Optimizing ease:} User experience can be enhanced by aligning feedback mechanisms with human \humanul{intent} \refhuman{D1}, avoiding ineffective or repetitive feedback. Leveraging \humanul{implicit feedback} \refhuman{D2} where appropriate, and striking an optimal balance between proactive and reactive feedback approaches \refhuman{D3}.\\

\subsection{Interface-Centered Qualities}
From an interface or system perspective, good feedback needs to fulfill two criteria mainly: It should be \textit{definite}, i.e., a faithful representation of the human expression and its associated uncertainty, and it ideally has a tractable \textit{context}, i.e., the context necessary to correctly interpret the context is available and transparent. Alternatively, the influence of external context should be minimized. These qualities can be mainly influenced by choices in the design dimensions of \combul{relation} \reffbp{D4}, \combul{content level} \reffbp{D5}, and \combul{actuality} \reffbp{D6}.\\

\shaded{Definiteness \reffbp{Q3}}{CombinedColor}{\label{fbp:Q3}How definite is the measurement of feedback, its associated context, and uncertainty is.}{\textbf{Low Definiteness} (Measurement quality unknown and user uncertainty unknown) -- \textbf{High definiteness} (Measurement uncertainty is known and tractable)} 
Uncertainty of human feedback comes in various forms. We distinguish between \humanul{user uncertainty}, \combul{systematic uncertainty}, and \modelul{model uncertainty}. We define user uncertainty as something that the user is either aware of (\humanul{explicit}) and could quantify or which is communicated via \humanul{implicit} interactions, e.g., hesitation or correcting feedback (\humanul{implicit}). Systematic uncertainty is inherent in the system; we refer to it as \textit{aleotoric} or irreducible uncertainty of the reward learning process. Lastly, model uncertainty is the uncertainty of the reward model concerning human feedback. This uncertainty has a reducible and irreducible component: Some part of it, the \textit{epistemic} uncertainty of the reward learning process, is reducible by acquiring more data about the human via feedback. However, the inherent uncertainty from the human and systematic uncertainty can not be reduced with more information.

\paragraph{How to evaluate \textit{Definiteness}?} As an interface-centered quality, \textit{definiteness} is evaluated more systematically, i.e., more akin to software testing.\\
\emph{Subcategories:}
\begin{itemize}[noitemsep,topsep=0pt,parsep=0pt,partopsep=0pt]
    \item \textbf{Measurement Complexity:} Which measurement variables are tracked (i.e., how well defined the \textit{feedback state} $f_s$ is) and whether uncertainties, inherent to the user or introduced in the measurement process (i.e., when translating implicit feedback), are quantified.
    \item \textbf{Completeness:} As part of the feedback measurement process, tracking a wide set of user interactions (such as mouse events, timings, interaction with visual interactions, etc.) enables more comprehensive modeling~\cite{faulkner2020interactive, bignold2021evaluation}.
    \item \textbf{Sensor Uncertainty:} In particular for implicit feedback, the quality of feedback might rely on the quality of sensor reading~\cite{cruz2018multi}.
    \item \textbf{User Uncertainty:} As part of the feedback process, users might be able to quantify their uncertainty for a given target (e.g., by a slider or by expressing uncertainty in natural language). We might also estimate uncertainty based on additional contextual measurements.
    \item \textbf{Task Verifiability:} An agents attempt at a task can be \emph{objectively} or \emph{subjectively} evaluated. We may also call certain tasks \emph{verifiable} or \emph{non-verifiable}. Objective tasks, such as mathematical calculations, with a clear objective solution or measurable outcome, are a good fit for automated validation strategies and require less or no human evaluation. Here, human feedback might be better suited as a source to give feedback on the process or intermediate steps, i.e., descriptive or instructive.
\end{itemize}
 Feedback with a high level of \textit{definiteness} enables the system to accurately model the feedback state and uncertainty associated with a feedback instance.\\
\emph{Optimizing definiteness:} \combul{Relative} feedback, or \combul{absolute} feedback \reffbp{D4} with ways to communicate confidence/uncertainty generally has higher definiteness. Measuring additional metrics beyond the explicit feedback interactions, including contextual factors like expertise level, background, and task familiarity, improves definiteness. We summarize many of these measurements as \combul{meta-level} \reffbp{D5} feedback. Finally, feedback for \combul{observed} \reffbp{D6} feedback has a significantly higher degree of definiteness because we can directly control the selection and definition of samples. On the flip side, \combul{generated} feedback might reflect the background and user context in a way that would be hard to uncover with pre-defined targets.\\

\shaded{Context Independence \reffbp{Q4}}{CombinedColor}{\label{fbp:Q4}To what extent does the feedback depend on the context of the interaction?}{\textbf{Low Context Independence} means that feedback is highly contextual, so specific to a human or task -- \textbf{High Context Independence} means that feedback values are less dependent on influencing factors}
A related aspect is context dependency, which means that feedback has to be interpreted either in relation to other elements inside the analysis context, e.g., other states, feedback, or the interfaces, which we call \textit{internal context dependency}. On the other hand, \textit{external context dependency} encompasses factors that are not explicitly modeled as part of the analysis context. For example, \combul{hypothetical absolute} feedback, e.g., demonstrations, are often generated \humanul{implicitly} in response to the current state of training, e.g., a desire to instruct agent behavior based on recent observation. Similarly, isolated ratings of episodes might also be influenced by previous internal contexts. In \combul{relative} feedback, we provide a reference as a controlled internal context. External context, which encompasses, e.g., the cultural or sociological context of the human and the surrounding of the interaction environment, can influence how certain types of feedback are communicated. In general, the less formalized feedback, e.g., \combul{hypothetical} \humanul{descriptive} and \humanul{implicit} feedback, the higher the degree of external context-dependency can be expected. Gestures, mimics, or natural language expressions may vary significantly between humans, whereas external factors influence simple scores or comparisons less.

\paragraph{How to evaluate \textit{Context Independence}?} 
Context-independence of feedback can be evaluated by tracking contextual factors alongside feedback values. These allow us to relate the feedback and identify both confounding factors and whether the feedback is robust to these.\\
\emph{Subcategories:} 
\begin{itemize}[noitemsep,topsep=0pt,parsep=0pt,partopsep=0pt]
    \item \textbf{Task Dependency:} The solution for a given task might be highly \emph{context-dependent}, or \emph{context-independent}. In context-dependent tasks, correct agent behavior must be adapted to specific circumstances. For example, some tasks must consider a user's background, adapt behavior, and interpret feedback accordingly. Based on this, we may adapt the feedback to track contextual factors or ignore them during translation and modeling.
    \item \textbf{User Dependency:} Feedback may be heavily dependent on the user's knowledge (both in terms of global domain knowledge or the knowledge during a given moment of the training). Tracking the knowledge of a user when measuring a feedback instance allows us to measure these dependencies.
    \item \textbf{Interface Dependency:} Feedback values depend on design choices in the implementation of the feedback process, e.g., choice set dependency~\cite{freedman2021choice} ("From how many different choices can a user select?"). Thus, measuring the effects of design variations on human feedback allows us to model these effects.
    \item \textbf{Model Dependency:} Feedback depends not only on the agent behavior and task but also the current policy of an agent~\cite{macglashan2017interactive}. In particular, humans adapt their feedback to the policy they observe during training. Measuring feedback values during the training process can uncover these dependencies.
    \item \textbf{Feedback Interdependency:} Feedback may depend on previous feedback instances, i.e., the magnitude of values is influenced by previous values~\cite{li2021learning}. One must treat sequential feedback as interdependent instead of viewing each instance as isolated to measure this bias.  
\end{itemize}
\emph{Optimizing context independence:} From an interface perspective, feedback is more independent of external context if it contains internal contextual information, i.e., \combul{relative} \reffbp{D4} feedback already induces a context between multiple targets, whereas \combul{absolute} feedback is potentially more context-dependent (this comparative stability of relative  (preference) feedback is generally accepted in the literature \cite{wirth2017survey}). However, other contextual factors might influence feedback, particularly if it is formulated more openly \cite{lindner_humans_2022}. Similar to the previous quality, additional measurements (meta feedback \reffbp{D5}) can be used to determine dependencies better. Enabling users to provide \combul{generated} \reffbp{D6} feedback beyond actually \combul{observed} targets or choices also allows them to learn more about context dependencies, as the choice of generated feedback by the human user can also reveal contextual information.\\

\subsection{Model-Centered Qualities}
From a model perspective, we need feedback that is \textit{learnable}, i.e., feedback that enables us to learn a faithful and successful reward model. Two qualities can be derived directly from machine learning fundamentals: Assuming there exists an internal, i.e., not directly observable, human reward model \footnote{There is extensive discussion why this assumption is not generally valid \cite{casper2023open}}, we want the feedback to have \textit{low variance/high precision} and \textit{low bias}, i.e., representing the human internal rewards accurately. Additionally, we want feedback to be \textit{informative}, i.e., revealing as much as possible about the internal human reward model. Model-centered qualities should be evaluated quantitatively by analyzing the feedback values and model performance.\\

\shaded{Precision \refmodel{Q5}}{ModelColor}{\label{model:Q5}If feedback stays consistent over multiple measurements, i.e. has low variance}{\textbf{Low Precision} (high variation or change over time) -- \textbf{High Precision} (stable and temporally coherent)}
Given a similar state to \humanul{evaluate} or a similar correction to propose, we ideally expect a high agreement between the given ratings or instructions. However, for different types of feedback, we expect varying levels of consistency over time. For example, we might expect feedback with a \combul{hypothetical} component to be relatively inconsistent because generating a demonstration without a reference, e.g., \combul{absolute hypothetical} feedback, requires choosing from a vast \humanul{implicit} space of possible state-action sequences. We, therefore, expect a high degree of variation. The feedback that is generated \combul{relative} to an \combul{observed} state, for example, a correction, can already be considered more consistent. Furthermore, feedback without a \combul{hypothetical} component, like evaluation, has fewer degrees of freedom, which should increase consistency over time. Consequently, \combul{relative} \humanul{evaluative} feedback for \combul{observed} states, for example, rankings or comparisons, will likely have the highest degree of consistency.

\paragraph{How to evaluate \textit{Precision}?} High precision/low variance of feedback values means that the measured values are close to an underlying "correct" feedback value (which is not directly observable). We can query feedback for the same targets multiple times to receive estimates of feedback variance. The observed variance might serve as an approximate measure of overall feedback variance.\\
\emph{Subcategories:} 
\begin{itemize}[noitemsep,topsep=0pt,parsep=0pt,partopsep=0pt]
    \item \textbf{Variation:} Stable feedback for similar targets indicates high precision, i.e., low variance. This means that for similar, or even the same, targets, we expect similar or equivalent feedback values instead of a widely spread distribution of values~\cite{bignold2021evaluation}.
    \item \textbf{Rates of Human Error:} Rates of misslabeling or human error are indicative of undesired variance~\cite{christiano2017deep, ibarz2018reward, faulkner2020interactive}.
    \item \textbf{Translation Accuracy:} Translating implicit feedback might introduce additional variance into feedback values, e.g., if the translation is also based on probabilistic modeling~\cite{cruz2018multi}.
\end{itemize}
\emph{Optimizing precision:} Because labeling error is a key factor in low precision, reducing error must be a key objective. Secondly, reducing the \modelul{choice set} \refmodel{D8} might improve precision/consistency because choices are more distinct. Pairwise preference might, therefore, have a lower variance than score feedback. Generally, reducing ambiguity in both the task and available choices for feedback values reduces error probability and variance. Although multiple \modelul{non-exclusive} reward sources~\refmodel{D9} have advantages (see the following quality criteria), they might introduce additional variance, as different sources potentially disagree with each other. T counteract the increased variance, the total number of samples must be increased.\\

\shaded{Unbiasedness \refmodel{Q6}}{ModelColor}{\label{model:Q6}Different biases that are inherent in the feedback and negatively influence its quality.}{\textbf{Unbiased} (feedback has low systematic error)  -- \textbf{Biased} (feedback values exhibit systematic error) }
Existing work has established multiple biases that might impact human feedback in reinforcement learning. Psychology has found that human judgment and extension feedback are highly susceptible to biases, such as base rates, priming, asymmetry, etc. We expect that these biases are reflected in human feedback. These biases become problematic if they hinder the model's learning of a correct reward estimate.\\
\paragraph{How to evaluate \textit{Unbiasedness}?} Without any reference, i.e., if human feedback is the only reward source, it is difficult to measure and estimate biases. We can specify a reference set of pre-labeled target data (in experimental settings, we could use ground-truth reward data) and then compare the human feedback with this reference data to estimate reward bias. In future cases, we can apply these learned bias estimates to new human feedback.\\
\emph{Subcategories:}
\begin{itemize}[noitemsep,topsep=0pt,parsep=0pt,partopsep=0pt] 
    \item \textbf{Distribution Bias:} A simple bias that has been described in the literature is \textit{asymmetry bias}, specifically humans tending to give more positive than negative feedback for rating-based feedback \cite{thomaz2007asymmetric, knox2012reinforcement}. The pretense of individual biases, e.g., in questionnaires~\cite{choi2004catalog}, is long-established. We should, therefore, expect such \textit{distributional} biases for evaluative feedback in RLHF.
    \item \textbf{Contextual Bias:} As mentioned previously (see~\reffbp{Q4}), the context can introduce biases in feedback. The context can act as a confounding factor in feedback. By defining the context, we can measure its effect on the feedback \cite{bignold2021evaluation}.
    \item \textbf{Complexity Bias:} A specific type of bias, prominently observed for LLMs is a bias towards more complex, e.g., lengthy or expertly seeming outputs or behaviors \cite{singhal2023long, park2024disentangling, hosking2024human}. 
    \item \textbf{Temporal Bias:} Finally, feedback in a particular situation can be biased by temporal dependencies, e.g., by other proceeding targets, ordering of targets, or learning/adaption processes during the interaction  \cite{li2021learning, saito2023verbositybiaspreferencelabeling}. Humans might give different feedback for the same target at different stages of the human-AI process.
\end{itemize}
\emph{Optimizing unbiasedness:} To improve unbiasedness, we must control for the influence of contextual factors on feedback values. As outlined above, tracking of contextual measurements can be used to adapt the translation or reward model to the estimated bias of the human user. In general, using a combination of \modelul{non-exclusive} reward sources~\refmodel{D9}, e.g., from different intentions, granularities~\refmodel{D7} or choice set sizes~\refmodel{D8} can be used to correct for specific biases of single sources.\\

%\subsubsection{Informativeness}
 \shaded{Informativeness \refmodel{Q7}}{ModelColor}{\label{model:Q7}The amount of information the feedback provides about how to update the model.}{\textbf{Uninformative} (Introduces redundant or irrelevant information)  -- \textbf{Informative} (Introduces novel and relevant information)  }
The agents update their reward model based on the given feedback. Informative feedback, therefore, leads to better model updates. Different types of feedback carry different amounts of information; for example, \modelul{instance-level} feedback for a \modelul{single step}, e.g., action advice, can potentially update the reward function to a small degree (especially if the feedback aligns well with the previous estimate). \modelul{Feature-level} information, e.g., evaluating the value of a feature's presence, might generalize more broadly and could, therefore, lead to a more significant update of the reward model. This quality is essential for reward models; however, it might be difficult to judge if a human is giving feedback to an agent.

Finally, we have established that human feedback is assigned to a particular \modelul{target granularity}, i.e., a subset of the state of potential state-action pairs. However, given that, human feedback may only target a part of the subset under observation. For example, a human might evaluate an entire episode but direct the feedback only to a particularly salient sub-sequence of an episode. We call this \modelul{partial coverage}. Ideally, we have a one-to-one match between feedback and state-action-sequence. On the other hand, humans could give feedback applicable beyond the immediate context in mind, e.g., \humanul{descriptive} feedback about the quality of particular \combul{features}.

\paragraph{How to evaluate \textit{Informativeness}?} 
We can evaluative informativeness by considering the reward model state.\\
\emph{Subcategories:}
\begin{itemize}[noitemsep,topsep=0pt,parsep=0pt,partopsep=0pt]
    \item \textbf{Information gain:} For some types of problems, we can directly compute the information gain of an instance of feedback given the model parameters \cite{ghosal2022effect, bignold2021evaluation, koppol2023interactive}.
    \item \textbf{Epistemic uncertainty:} We can estimate uncertainty for a reward model, e.g., by quantifying the disagreement between sub-models of a model ensemble. In that case, we can estimate if a state-action pair is in or out of distribution. Thus, feedback for targets with high associated uncertainty is generally more informative. Similarly, we can analyze the drop of uncertainty for the reward model in response to a model update based on a feedback instance~\cite{ji2024reinforcementlearninghumanfeedback, christiano2017deep, zhan2023query}.
    \item \textbf{Learning Impact:} We can analyze learning speed/update strength during reward model training. If we can evaluate the trained reward model or agent learning with the reward model, we could estimate the effectiveness of feedback by looking at the learning curve~\cite{koppol2023interactive}.
    \item \textbf{Coverage:} Properties like the coverage, i.e., which part of the state-action space is covered by a feedback instance, could also be interpreted as measures of informativeness. The redundancy/non-redundancy of information relies on the state of reward model.
\end{itemize}
\emph{Optimizing informativeness:} Based on metrics such as uncertainty or expected information gain, we can actively query the user for targets and feedback types~\cite{jeon_reward-rational_2020, ghosal2022effect, zhan2023query}. Being able to adapt \modelul{granularity}~\refmodel{D7} or \modelul{choice set size}~\refmodel{D8} of a query improves the possibility of receiving targeted and varied feedback. Access to multiple \modelul{non-exclusive} reward sources~\refmodel{D9} also opens up the space for informative queries and feedback.

\clearpage
\section{Designing Systems for Diverse Human Feedback: Existing Approaches and Opportunities}
\label{sec:deriving_requirements}
In the previous sections, we have presented a conceptual framework of nine dimensions and seven qualities for human feedback. Now, we want to outline how we can operationalize the framework.

We begin by briefly introducing a basic architecture, i.e., the main components of a system learning from expressive reinforcement learning from human feedback. Then, we outline requirements and design choices for systems spanning the space of dimensions. We reference existing work in human-AI interaction and interactive RL for each dimension. Additionally, the framework allows us to identify research gaps and opportunities.

\subsection{Basic System Design}
\begin{figure}
	\centering
  \includegraphics[width=0.65\linewidth]{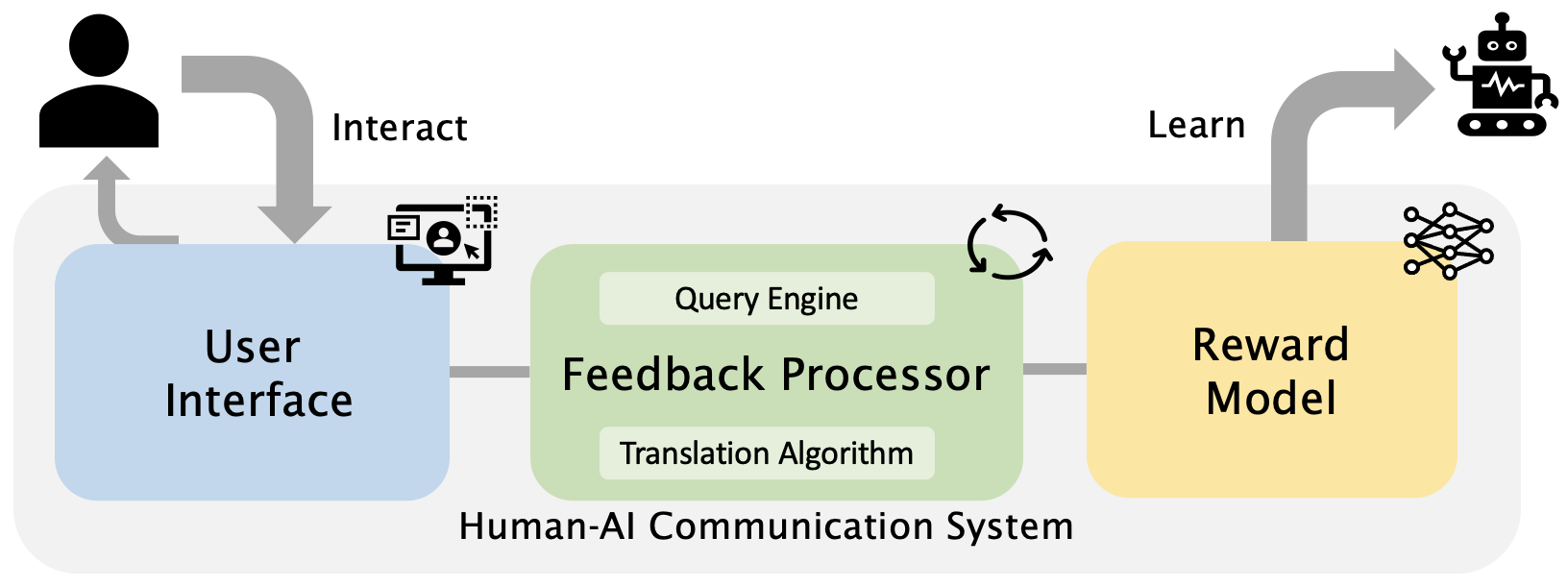}
   \caption{\label{fig:basic_system_design} The basic system components for reward learning: A \humanul{user interface}, \combul{feedback processor}, and \modelul{reward model}.
    }
\end{figure}

In \autoref{fig:basic_system_design}, we summarize the identified core components of an RL-based human-AI interaction setup: A \humanul{user interface}, implementing feedback interactions, as well as providing a view of the two-sided communication, i.e., the targets feedback can be provided for and potentially additional information about the agent or training process. The \combul{feedback processor} takes the data recorded in feedback interactions, potentially unstructured and highly diverse, and translates it into a standard format that can be passed onto the reward modeling stage. The \modelul{reward model} is optimized based on human feedback and must capture human preferences accurately and allow the AI agent to optimize its behavior to maximize the predicted reward. 

The choice of feedback generally influences the whole stack of system components, i.e., implementing feedback of different types requires adaptations to the UI, feedback processor, and reward model. In the following chapter, we discuss the influence of feedback on all system components.

\paragraph{User Interface}
How can humans communicate with the agent? Depending on the setting, we have specific interaction mechanics, like verbalization, movement, or using a GUI. Interactive visual user interfaces can enable expressive and situation-aware multi-type feedback. The user interface is responsible for recording measurements, intrinsic and contextual ones. Therefore, an ideal interface might go beyond just implementing the narrow explicit interactions and provide additional inputs, different types of feedback, and even multi-modal inputs like camera, voice, etc.

\requirementbox{User Interface Requirements}{HumanColor}{The user interface serves as the interface for communication between the human and the agent. Here, the human inputs feedback via interactions and receives information about the system and the environment. We infer the user interface requirements on human-centered qualities $\refhuman{Q1}$, $\refhuman{Q2}$, as well as interface-centered qualities $\reffbp{Q3}$, $\reffbp{Q4}$. However, UI design has huge implications for the reward model training as well:}
%Based on the human-centered dimensions D1 and D2, structural dimensions D3 and D4, as well as the human-centered quality measures Q1, Q2, and Q7 we identify the following requirements for user interfaces:}
{(UI.R1) Give users the ability to express feedback matching their intentions. Implement versatile input interactions for evaluative, instructive, or descriptive feedback should be available. \refhuman{Q1} (UI.R2) User interfaces should consider cognitive demand and user fatigue, i.e., interactions should be efficient and the use of human input minimized. \refhuman{Q2} (U1.R3) User interfaces should enable the measurement of uncertainty or additional metrics to verify the quality of feedback. \reffbp{Q3} (U1.R4) User interfaces should track contextual factors. \reffbp{Q4}}

\paragraph{Feedback Processor}
When going beyond simple, explicit feedback, which can be fed to a reward model directly, we must integrate a processing step that translates the measurements into "learnable" inputs for the reward model. The feedback processor implements the \textbf{translation algorithm} $\phi$ specified above.

In our framework, we assign a second role to the feedback processor. The processor also serves as the \textbf{query engine}, which handles \textit{target selection} $(\mathcal{T} \sim T)$, i.e., determines which targets are shown to the human user. This querying can be combined with varying levels of human involvement, i.e., manual or assisted selection of targets by the human user.

\requirementbox{Feedback Processing Requirements}{CombinedColor}{The feedback processor parses the different expressions of human feedback. The processor has access to all relevant data in the analysis context and mediates bi-directional processes such as querying and user-aware modeling. We identify the following requirements, considering dimensions, based on qualities \reffbp{Q3}, \reffbp{Q4}, \refmodel{Q5}, \refmodel{Q6}, and \refmodel{Q7}:}
{(FP.R1) Define a consistent protocol to translate different forms of human feedback to a format that can be passed onto a reward model. \reffbp{Q3} (FP.R2) Collect and process necessary information for reward learning, like user and task context, temporal dynamics, or uncertainty, to build the basis to adapt to the user and previous feedback. \reffbp{Q3} \reffbp{Q4} (FP.R3) Apply possible corrections for variance and biases based on measurements. \refmodel{Q5} \refmodel{Q6} (FP.R4) Implement a target querying strategy that facilitates high feedback quality. \refhuman{Q2} \reffbp{Q3} \reffbp{Q4} \refmodel{Q5} \refmodel{Q7}}

\paragraph{Reward Model}
As introduced in \autoref{subsec:problem_characterization}, the RL-based learner needs access to a reward function that assigns state-action(-state) pairs(triplets) a scalar-valued reward. This reward function can be implemented in a variety of ways: It can be explicitly supplied by the environment, e.g., by being implemented as a table stating the reward for a fixed set of state-action trajectories~\cite{Sutton2018} or an in-game score~\cite{Bellemare2013}. However, this means the reward function is fixed or pre-specified. A \textit{reward model} is a dynamic reward function that can be updated, e.g., with human feedback. A reward model is used to train the agent asynchronously. Because a model can predict reward for arbitrary state-action sequences, asking for human feedback after each step or episode is unnecessary. The choice of function class determines the expressiveness of the reward model, i.e., the complexity of factors that can be considered. Here, we identify a trade-off between different models: We may use complex high-fidelity models that can capture a large degree of variation and potential nuance, e.g., by modeling the human as a separate agent with a partially observable state~\cite{hadfield2016cooperative, lindner2021learning, shah2020benefits}. However, such models may require large amounts of data to provide ``learnable'' reward estimation, particularly in complex scenarios. On the other hand, we can choose simpler models that cannot capture certain aspects, e.g., fatigue or updating knowledge in humans. Such a model may be significantly more straightforward to train because there are fewer free parameters than need to adapt based on data~\cite{jeon_reward-rational_2020, ghosal2022effect}.

\requirementbox{Reward Model Requirements}{ModelColor}{The reward model learns a reward function that is used to perform internal training of the agent. Considering \refhuman{Q1}, \refmodel{Q5}, \refmodel{Q6}, and \refmodel{Q7}, we identify the following requirements:}{(RM.R1) Learning joint reward models from diverse and expressive feedback \refhuman{Q1} (RM.R2) Consider feedback bias and variance in the estimation \refmodel{Q5} \refmodel{Q6} (RM.R3) Use human feedback efficiently and provide a mechanism to estimate informativeness \refmodel{Q7}}

\subsection{Systems for Feedback of Different Human-Centered Dimensions}
In the following, we discuss proposed solutions for feedback on different intents, expression forms, and engagement. To move beyond existing work, we highlight research opportunities for human-centered human feedback.\\

\requirementbox{Human-Centered}{HumanColor}
{Relevant Dimensions: Human Intent \refhuman{D1},  Expression \refhuman{D2}, Engagement \refhuman{D3}}
{Relevant Qualities: Expressiveness \refhuman{Q1}, Ease \refhuman{Q2}} 
\vspace{-2em}
\subsubsection{Human Intent~\refhuman{D1}}
We can design systems to enable humans to express their feedback in a way that matches their intent, namely to evaluate, instruct, or describe. Furthermore, we can also implement mechanics that capture unintentional human feedback.

\paragraph{Evaluate}
\noindent \refhuman{VP} \label{human:VP} Human feedback is often obtained in experimental settings with simplistic user interfaces, e.g., where users elicit their preferences between pairs of trajectories shown in short videos by clicking a button~\cite{christiano2017deep, torne2023breadcrumbs}. Often, it is also possible for humans to state no preference. This interface is simple, easy to understand, and therefore suited for large-scale collection of human preference data as is common for fine-tuning of LLMs \cite{ouyang2022training}.\\
\refhuman{PI} \label{human:PI} A first notable example of interfaces going beyond simple interactions is by Thomaz and Breazeal~\cite{THOMAZ2008716teachablerobots}. Users provided feedback by clicking and dragging their mouse, where the length of the dragging motion determined the magnitude, and its direction decided the sign of the corresponding reward.\\
\refhuman{TC} \label{human:TC} Recent examples of data labeling also approach fine-tuning large language models to generate more appropriate responses. Often, this feedback is comparative in that, e.g., users indicate which of two suggested answers is more harmful~\cite{ganguli2022red}.
Recently,  Zhang et al.~\cite{zhang_time-efficient_2022} introduced \textit{Time-Efficient Reward Learning} via visually assisted cluster ranking. This work utilizes scatterplot interactions with cluster candidates but is limited in scope and restricted to ranking-based feedback.\\
In another example, \textit{Human Feedback in Continuous Actor-Critic Reinforcement Learning} employs a slider interface for approving or disapproving agent actions, enabling real-time preference indication.\\ 
\cite{chetouani2021interactive} provides an overview of evaluative feedback methods tailored for robotics, demonstrating effectiveness in dynamic environments.

\paragraph{Instruct}
There has been some work enhancing user interaction and feedback for demonstration generation. 
\refhuman{CF} One example, \textit{CueTip}, introduces a mixed-initiative tool for handwriting-to-text translation, where users can provide corrective feedback through multiple interface options~\cite{5.4_quetip}. Given the sequence-like nature of the text, some of these interface choices might translate well into episode editing.\\
\refhuman{GG} \label{human:GG} Furthermore, \textit{GestureScript} is a user interface designed for single-stroke gesture classification, allowing users to create example gestures and provide structural information about the gesture drawing~\cite{gesturescript}. The sequence-like nature of Gesture Script makes it comparable to episodes, and its interface offers an exciting way to structure feedback, potentially applicable for annotating demonstrations.
\refhuman{GA} \label{human:GA} In \textit{ML-Agents}, users can record demonstrations directly in the game-like environment~\cite{juliani2020mlagents}. To that end, the developers implement a function that maps user input to agent actions.\\
\refhuman{PC} \label{human:PC} Finally, an example of correcting the physical behavior of robots is to modify/correct a movement trajectory via physical interactions. For many physical problems, such direction interactions can be significantly more efficient than the use of a virtual user interface.\\
Instructional feedback involves interfaces where users guide agents by specifying actions directly. 
Studies like \cite{Amir2016} employ a student-teacher framework, enabling a teacher to direct actions for optimized training outcomes.\\
In a similar approach, \cite{krening2018newtonian} maps linguistic instructions to actions within game environments, enhancing flexibility in human-robot communication.

\paragraph{Describe}
\refhuman{IL} \label{human:IL} One example for pixel-wise annotations, is the ``Crayon method''~\cite{5.5_crayon}, that enables to annotates image regions for pixel classification. Users are interactively queried to provide more annotations if needed.\\
\refhuman{SL} \label{human:SL} \textit{FeatureInsight} is a tool that allows users to classify websites based on their content in a process the authors dub \textit{structured labeling}~\cite{featureinsight}. Users label websites by dragging them into self-created categories or opening a new category if needed. With that, it offers more flexibility and empowers the users' decision-making.\\
\refhuman{SI} \label{human:SI} \textit{ReGroup} is a user interface that allows users to interactively create custom social networks by assigning members to groups~\cite{5.4_regroup}. The system simultaneously learns a probabilistic model that suggests new members to the user for inclusion, thus pruning irrelevant samples. RL training generates large amounts of unlabeled data, but the users' attention and motivation are limited.\\
\refhuman{FA} \label{human:FA} Annotation was not only for labeling instances but also to specify meaningful features for machine learning models. One such system is \textit{Flock}, which uses human feedback to select a set of relevant features and then algorithmically weighs these features.

\paragraph{Evaluate + Instruct}
Some interfaces allow both evaluative and instructional feedback. For example, \cite{fitzgerald2023inquire} uses a multi-type feedback approach where users can both evaluate and instruct.\\
Similarly, \cite{mehta2022unified} provides a robotics interface where users can demonstrate or correct behaviors. Another notable example, \cite{chi2022instruct}, allows users to alternate between instructions and evaluations, facilitating flexible user-agent interactions.

\paragraph{Evaluate + Instruct + Describe}
In cases where multiple types of feedback are needed, interfaces incorporate evaluative, instructional, and descriptive intents. An example is \cite{sumers2021learning}, which employs a game-like setting where users provide linguistic feedback that includes preferences, instructions, and descriptive elements, creating a robust, multi-faceted learning environment.

\paragraph{Unintentional}
Unintentional feedback can be derived from spontaneous reactions or subconscious signals, providing insights that are indirectly informative. Studies like \cite{veeriah2016face} leverage facial expression analysis to gather implicit satisfaction levels, which are used to enhance agent adaptation. Additionally, physiological signals, such as EEG potentials, can indicate user discomfort or attention focus, as explored in \cite{kim2017intrinsic}. Eye-tracking is another modality used for unintentional feedback, where gaze patterns reveal regions of interest, aiding agents in refining state representations.

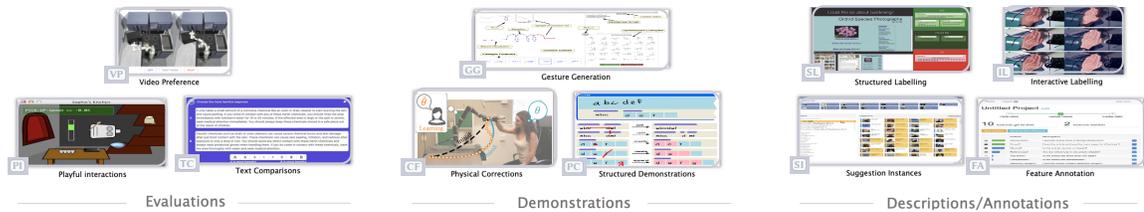
\begin{figure*}[tbh]

\begin{annotatedFigure}
	{\includegraphics[width=1.0\linewidth]{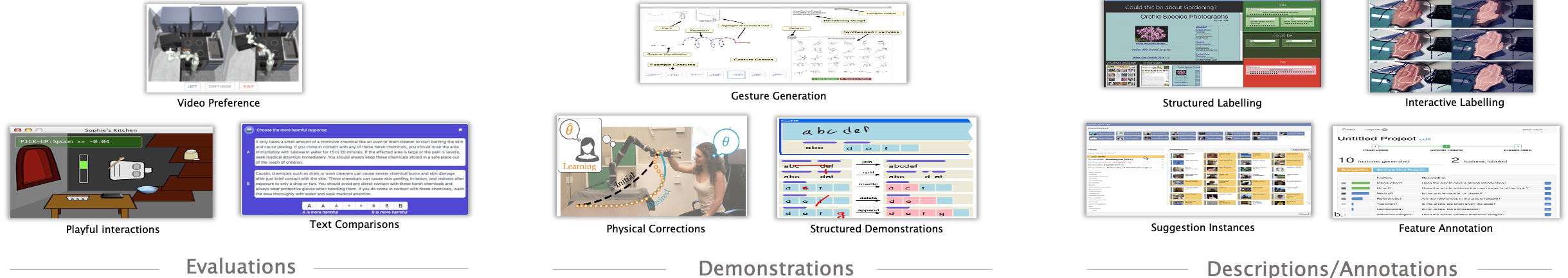}}
	\annotatedFigureBox{0.093,0.6677}{0.194,0.997}{\refhuman{VP}}{0.093,0.6677}%bl
	\annotatedFigureBox{0.006,0.217}{0.1372,0.5478}{\refhuman{PI}}{0.006,0.217}%bl
	\annotatedFigureBox{0.155,0.2283}{0.2995,0.552}{\refhuman{TC}}{0.155,0.2283}%bl
	\annotatedFigureBox{0.406,0.6902}{0.5785,0.9872}{\refhuman{GG}}{0.406,0.6902}%bl
	\annotatedFigureBox{0.354,0.2114}{0.4795,0.5985}{\refhuman{CF}}{0.354,0.2114}%bl
	\annotatedFigureBox{0.495,0.2114}{0.6235,0.5816}{\refhuman{PC}}{0.495,0.2114}%bl
	\annotatedFigureBox{0.705,0.6789}{0.8425,1.0041}{\refhuman{SL}}{0.705,0.6789}%bl
	\annotatedFigureBox{0.8725,0.6789}{0.9775,1.0041}{\refhuman{IL}}{0.8725,0.6789}%bl
	\annotatedFigureBox{0.693,0.2227}{0.8365,0.5534}{\refhuman{SI}}{0.693,0.2227}%bl
	\annotatedFigureBox{0.85,0.2171}{0.9965,0.5478}{\refhuman{FA}}{0.85,0.2171}%bl
\end{annotatedFigure}

\caption{\textbf{A selection of proposed solutions to match human intentions} -- There have been several proposed solutions for user interfaces, the feedback processor, and the reward model. For user interfaces, we highlight (A) evaluative workflows, (B) demonstrations, and (C) descriptions.}
\label{fig:intent-selection_work}
\end{figure*}

\oppbox{1. Matching Human Intent}{Even though some preliminary studies exist \cite{koert2020multi, mehta2022unified, crochepierre2022interactive, metz2023rlhf}, we see much potential in \textbf{combining feedback of multiple intents}. To enable expressive \refhuman{Q1} feedback, it is worth going beyond simple interactions, e.g., ratings or pairwise preferences, and towards more complex interactions, enabling instructive or descriptive feedback if applicable. Giving the human user the ability to choose an appropriate interaction that matches their intent in a given situation can potentially help greatly improve the expressiveness of feedback.\\
Adapting interactions to match human intent can also help improve ease \refhuman{Q2} by reducing fatigue due to repetitive feedback or expressing feedback in a way that feels comfortable and requires an acceptable amount of cognitive effort in a given moment. 
}

\subsubsection{Expression Form~\refhuman{D2}}
On the topic of expression form, there has been much work in fields like human-robot interaction \cite{chi2022instruct, fitzgerald2023inquire}, as well as interactive RL \cite{lin2020review}. Due to the large body of existing work, we want to limit the discussion of implicit feedback in particular and point at related literature for inspiration \cite{lin2020review, chi2022instruct}.

\paragraph{Explicit Expression}
\noindent \defhuman{TC} Explicit expression methods enable users to convey feedback with clarity and precision, often using textual interfaces or simple selection tools. For example, chat interfaces are commonly employed for preference ranking or qualitative feedback responses, as seen in Ouyang et al.~\cite{ouyang2022training} or Bai et al.~\cite{bai2022training}. Here, users can explicitly express preferences between responses, guiding model fine-tuning. Additionally, fine-grained feedback via the use of interactive text annotation interfaces has been explored~\cite{wu2023fine}, where explicit feedback is given directly on textual content, allowing for targeted improvements.

\paragraph{Implicit Expression}
\noindent Implicit expression captures feedback indirectly, often through involuntary or less direct cues. Interfaces using eye-tracking technology, for example, capture the user's focus points to subtly guide agents in interpreting regions of interest~\cite{aronson2022gaze}. Physiological signals, such as EEG measurements~\cite{xu2021accelerating}, can also provide feedback on user engagement, attention, or comfort, allowing systems to adapt based on real-time emotional states.

\paragraph{Mixed-Modal Expression}
Combining explicit and implicit feedback interactions is still relatively rare. 

\oppbox{2. Expression Form}{Human-computer/human-robot interaction presents a huge opportunity to create novel feedback interactions. \textbf{Dynamic feedback interactions} can improve expressiveness by providing humans with natural and versatile input mechanics \refhuman{Q1}.\\
Both explicit and implicit interactions can be designed with ergonomic principles to reduce cognitive load and stress, i.e., effort overall. \refhuman{Q2}. We see great potential in human-centered studies to investigate implicit user interactions for human feedback and optimize explicit interactions via user interfaces.
}

\subsubsection{Engagement~\refhuman{D3}}
How to engage humans in RLHF is a central topic because we need to balance the quantity and quality of human feedback with human requirements such as fatigue and cognitive demand. As outlined, we can design proactive and reactive feedback strategies, i.e., letting humans select interesting instances or querying for feedback.

However, a promising middle ground is to combine selection and querying effectively:
Instead of just improving querying strategies, users can be instructed to provide better (more AI-appropriate) teaching. Optimal teaching for humans can substantially differ from optimal teaching for AI~\cite{good_teaching_for_ml}\\.
In active learning, a type of semi-supervised learning, the primary goal is to achieve high accuracy with minimal manual labeling effort. This is achieved by a candidate selection strategy identifying the instances that potentially contribute most to the learning progress of the model. Several strategies have been proposed, including \textit{uncertainty sampling}, \textit{relevance-based selection}, \textit{data-dependent sampling} or sampling based on \textit{error reduction estimation}~\cite{bernard2017comparing, settles2009active, olsson2009literature}. 

Reinforcement learning agents learn in a potentially large state space of unknown/unlabeled states, which must be reached by exploration. In active learning~\cite{bernard2017comparing, settles2009active, olsson2009literature}, exploration is undertaken by the human annotators, although often supported by automatic metrics, while it is generally fully automated in RL. Therefore, existing work in human-in-the-loop reinforcement learning has focused on integrating humans into the learning process to effectively guide the exploration of new states or the exploitation of good optimal states. Compared to supervised learning, giving feedback for state-action sequences (similar to unlabeled states) shapes the exploration of an acting agent, influencing the composition of the set of possible states. Selecting such high-impact states is still an unsolved question for on-policy learning and might, therefore, especially profit from human oversight and expertise.

Visual interactive labeling, established by~\cite{bernard2018vial}, describes the unification of labeling across machine learning and visual analytics. The \textit{VIAL} process introduces specific important steps, which we also find for reward-based learning from human feedback: (1) The necessity to visualize a model and its progress during training, (2) A visualization of results, i.e., the classification results. (3) Candidate suggestion, which needs to find a balance between instances that improve the model most while respecting a user's perspective and capacity. (4) A labeling interface appropriate for the used data instance and label types. (5) A way to interpret user feedback.
Multiple works have explored visual interactive labeling, e.g., integrating gamification and explainable AI~\cite{sevastjanova_questioncomb_2021}.

\begin{figure}[h]
    \centering
    \includegraphics[width=1\linewidth]{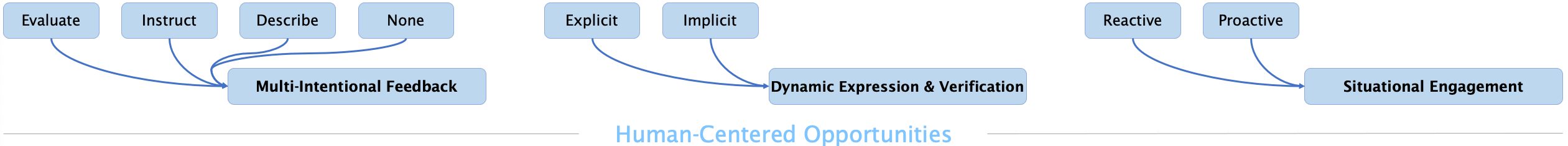}
    \caption{A Summary of Opportunities for Human-Centered Dimensions}
    \label{fig:human_opp_summary}
\end{figure}

\oppbox{3. Engagement}{Query mechanics for single feedback types have already been widely studied. However, querying for multiple feedback types has been explored to a small degree. Investigating this kind of \textbf{situational engagement} has great potential because querying for the right feedback type at the right time can improve expressiveness \refhuman{Q1}, reduce effort \refhuman{Q2} while preserving informativeness \refhuman{Q7}. Finally, mixed strategies like visual interactive labeling could be highly effective for RLHF, and there is significant potential for exploring such approaches further.
}
The opportunities for human-centered human feedback are summarized in~\autoref{fig:human_opp_summary}.

\subsection{Systems for Feedback of Different Interface-Centered Dimensions}
We want to discuss proposed solutions and opportunities for feedback on different relations, content levels, and actuality. Compared to the previous dimension, the choice of interface-centered dimensions is more dependent on the actual type of target (i.e., text, images, numerical values), so we will comment on this difference when appropriate.\\

\requirementbox{Interface-Centered}{CombinedColor}
{Relevant Dimensions: Target Relation \reffbp{D4},  Content Level \reffbp{D5}, Target Actuality \reffbp{D6}}
{Relevant Qualities: Definiteness \refhuman{Q3}, Context Independence \refhuman{Q4}} 
\vspace{-2em}
\subsubsection{Target Relation~\reffbp{D4}}
The move from absolute to relative feedback, i.e., from approaches like interactive shaping to preference-based RL, has been established over the last few years. Preference-based RL is now the standard RLHF approach, particularly for fine-tuning LLMs \cite{Ziegler2019finetuning, ouyang2022training}. Binary preferences strike a favorable combination of simplicity, which is especially relevant for crowd-sourced feedback and stability of feedback.\\
So far, relative feedback, such as preferences or rankings, has been used almost exclusively for evaluations. Here, they are easily implemented, e.g., via simple button interfaces, and work for modalities that can be visualized (text, but also videos of execution). However, some work has, e.g., investigated the use of implicit feedback methods to extract relative feedback.\\
So far, relative feedback for instructions or descriptions has been investigated to a far smaller degree. Here, a potential example would be action advice with multiple actions (i.e., recommending a best, but also second-, third-best action, etc.). Similarly, users could be asked first to generate a few demonstrations and then rank them according to self-perceived quality, resulting in relative demonstrative feedback. Similarly, it might be possible to formulate relative descriptive statements or rules instead of absolute constraints or descriptive statements.

\oppbox{4. Target Relation}{As relative feedback often has significant advantages in terms of context dependency \reffbp{Q4}, a currently under-explored research direction is the use of instructive/descriptive feedback as relative feedback, e.g., by giving humans the ability to express preferences between demonstrations or descriptions like rules and constraints. By varying contextual feedback for the same target, we can use relative relationships between measurements to define influencing factors \reffbp{Q3}.
}

\subsubsection{Content Level~\reffbp{D5}}
A critical dimension to improving feedback's expressiveness is using feature-level feedback. Instance-level is the most dominant type of feedback because it can be implemented via simple user interactions (like sliders, buttons, mimics detection \cite{knox2009interactively, li2018interactive, christiano2017deep}). Even though user interactions for demonstrative feedback are generally more challenging to implement, instance-based feedback requires uni-directional interactions that let human users select or generate actions. On the other hand, feature-level feedback requires more complex user interfaces in order to let users select parts of the feature space (e.g., selecting regions of image-based input as salient features \cite{metz2023rlhf, zhang2019leveraging}). Some feature-level feedback might be collected via implicit, non-conscious interactions such as gaze tracking~\cite{zhang2020atari}. However, we might collect such feedback via explicit \textbf{annotations}, i.e., by allowing a human user to annotate behavior and states, selecting sub-features of the generated state-action sequences~\cite{zhang2023self}. 

The differentiation between instance- and feature-level feedback for language models is more challenging because we could interpret each predicted token as an action of the \textit{agentic} language model. Therefore, annotating single words from a fully generated text could be interpreted as fine-grained feature-level feedback~\cite{wu2023fine}. However, a better interpretation of feature-level feedback in the context of natural language might be the annotation of specific facts or generation patterns that occur over multiple generations (i.e., are not bound to a specific generated instance). User interfaces allowing these feature-level annotations might open novel ways to train language models.

Finally, meta-level feedback is an emerging technique in RLHF. Such feedback can be integrated into the normal interaction process, either before a session or during normal feedback interactions. One example is \emph{skill assessment questions}~\cite{singhal2024scalable}, which uses additional contextual questions not directly related to feedback to asses the domain knowledge of a human labeler. This, in turn, allows us to calibrate the expected quality of reward for certain difficult queries. For such queries, feedback from domain experts should be rated higher than feedback from a lay user.\\
A second example is \emph{distinguishability queries}~\cite{feng2024comparing}, which combines pairwise preferences with queries to measure which segment pairs are easy or difficult to distinguish for labelers. In turn, we can trust preference pairs more highly distinguishable than pairs that are only hard to differentiate. We can also adopt our querying strategy to serve pairs specifically.

\oppbox{5. Content Level}{In our mind, moving beyond instance-level feedback is a key step towards more expressive feedback. Feature-level feedback suits itself to define human values and preferences more fine-grained \reffbp{Q3} and independent of the immediate feedback context \reffbp{Q4}. Exploring ways to collect, encode, and process this feedback is crucial for more powerful human-AI communication. Meta-level feedback can be intensively valuable in understanding contextual factors of human feedback and calibrating the reward learning process accordingly. Combining instance-level feedback with content-level \textbf{annotations} and meta-level interactions is a way to combine learnability \refmodel{Q5}\refmodel{Q6}\refmodel{Q7} with expressiveness \refhuman{Q1}, improving definiteness of feedback \reffbp{Q3} and context dependence~\reffbp{Q4}.
}

\subsubsection{Target Actuality~\reffbp{D6}}
Generative feedback generally requires more complex user interactions compared to feedback for observed states. Feedback for observed states can be simple content-level feedback (such as preferences and scores) and feature-level feedback (annotations, descriptions). Such feedback, therefore, requires user interactions for \textbf{selection}, potentially at different temporal granularity (see~\refmodel{D7}), feature granularity. We need an interaction to select a correct action for demonstrative feedback, such as action advice. 

For generative feedback, we need to provide \textbf{generative} user interactions, i.e., a way to choose and execute actions that generate demonstrations, but also sub-goal generation~\cite{xu2023dexterous}, creation of novel rules or constraints~\cite{kuhlmann2004guiding, wilde2020improving}. For natural language targets, the generation of demonstrative feedback is simple in principle and can be implemented easily as normal text input. However, more complex user interactions can enable more expressive text-based feedback (such as descriptions).

Generating new targets without any reference might be overwhelming to users, especially if they are not experienced with certain AI systems. Providing stronger guidance in the form of references can improve the ability of humans to generate targeted and helpful feedback at the price of potentially biasing generation if the reference influences the user too strongly.

\begin{figure}[h]
    \centering
    \includegraphics[width=1\linewidth]{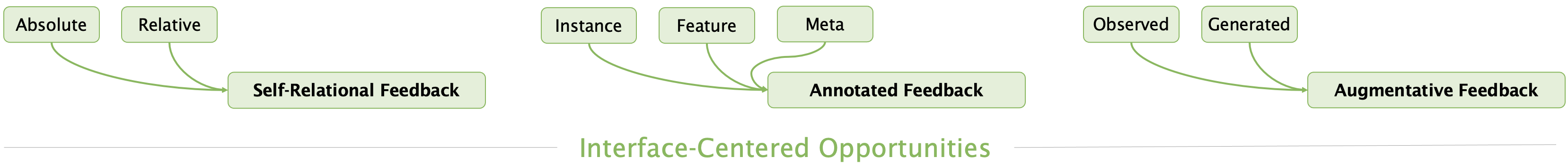}
    \caption{A Summary of Opportunities for Interface-Centered Dimensions}
    \label{fig:interface_opp_summary}
\end{figure}

\oppbox{6. Target Actuality}{Generative feedback, not reliant on observed behavior, can be highly informative and serve learning agents as strong guidance for exploration, especially at lower skill levels. Therefore, implementing systems that enable such generative feedback is a worthwhile challenge, especially in domains where explicit input or demonstrations are difficult (e.g., in robotics, the explicit specification of robot positions is difficult); we need to develop novel user interactions for generated feedback. Not losing information in this process \reffbp{Q3} while maintaining a consistent context \reffbp{Q4} remain important challenges. To not overwhelm the user, we can focus specifically on the \textbf{augmentation} of existing behavior, i.e., giving users a starting point (like an initial state-action sequence or another type of reference) to generate new targets for feedback.}
The opportunities for human-centered human feedback are summarized in~\autoref{fig:interface_opp_summary}.

\subsection{Systems for Feedback of Different Model-Centered Dimensions}
Finally, we want to discuss proposed solutions and opportunities for feedback on different granularities, choice set sizes, and exclusivity.\\

\requirementbox{Model-Centered}{ModelColor}
{Relevant Dimensions: Temporal Granularity \refhuman{D7},  Choice Set Size \refhuman{D8}, Exclusivity \refhuman{D9}}
{Relevant Qualities: Precision \refmodel{Q5}, Unbiasedness \refmodel{Q6}, Informativeness \refmodel{Q7}} 
\vspace{-2em}
\subsubsection{Temporal Granularity~\refmodel{D7}}
An element missing from simple approaches, e.g., in ranking-based reward learning~\cite{Ziegler2019finetuning}, is a way to get a good overview of the candidate set, i.e., the set of possible instances available for feedback. For inspiration, we can look at existing visual interactive labeling approaches~\cite{bernard2018vial, bernard_towards_2018, bernard2018vial} that describe the unification of labeling across machine learning and visual analytics. The \textit{VIAL} process introduces certain important steps, which we also find for reward-based learning from human feedback: (1) Visualizing a model and its progress during training. (2) Visualizing results, i.e., the classification results. (3) Visualizing candidate suggestion, which needs to find a balance between instances that improve the model most while respecting a user's perspective and capacity. (4) A labeling interface appropriate for the used data instance and label types. (5) A way to interpret user feedback. Multiple works have explored visual interactive labeling, e.g., integrating gamification and explainable AI~\cite{sevastjanova_questioncomb_2021}. Instead of instances where we want to assign a discrete or numeric label, we are interested in assigning rewards to state-action sequences of varying lengths. Compared to data types such as images, state-action sequences can have a high degree of heterogeneity, particularly due to different lengths and content.

We examine existing approaches to visualizing state-action sequences in interactive machine-learning systems. As previously outlined, these instances can include states, observations, actions, scalar quantities such as cumulative reward, and sequences thereof, for example, state-action sequences or more abstract objects like policies.

\paragraph{Single States} We can differentiate between two main approaches to state visualizations. In the first, states are visualized directly, and in the second, the visualization of the state is a function of the state itself, which applies further processing or abstraction. Many popular reinforcement learning environments already provide visual representations of states that could convey the agent's state without further processing. 
\defhuman{DS}In the \textit{ATARI} domain~\cite{Bellemare2013}, an observation itself is a raster image of the game state and is thus inherently visual. In other domains, however, visualizing states directly is not possible. 
\defhuman{IS} In the \textit{MuJoCo} domain~\cite{mujoco}, the state corresponds to a configuration of joint angles of robot skeletons, which, even for a small number of joints, is hard to visualize and understand intuitively. Instead, the state is shown by creating a 3D rendering of the robot in its environment.

\paragraph{Single Actions} Actions are often not visualized directly but are left implicit and need to be inferred by the user by comparing consecutive visualizations of states, which might be reasonable for small and discrete action spaces but becomes infeasible for large or continuous spaces, where the size of the action space does not permit this approach. 
\defhuman{AC} A possible approach to visualizing distributions of chosen actions, divorced from the corresponding state, is exemplified with \textit{DQNViz} on the \textit{Breakout} game with a small, discrete action space~\cite{Wang2018-dqnviz}. The user interface provides a summary of chosen, discrete action distributions per episode as a pie chart and the temporal progression of these action distributions throughout training as a stacked bar chart.

\paragraph{State-Action Segments} States, chosen actions, and resulting states are highly interdependent, and as a result, existing visualizations often display them jointly. A widely-used approach to visualizing state-action sequences is to show them as videos, where each frame corresponds to the visualization of a single state in the sequence~\cite{Atrey2020Exploratory, McGregor2015Facilitating}. As alluded to earlier, this approach has its limitations, as it does not explicitly show the actions taken by the agent.

\defhuman{SA} A possible remedy can be found in Luo et al.~\cite{luo_visual_2018}, where state-actions sequences in a visual environment are displayed as grids consisting of visual observations, which have a different tint depending on which of the three possible actions the agent has taken.

\defhuman{ST} \textit{DQNViz}~\cite{Wang2018-dqnviz} also uses joint displays of states and actions in the \textit{Breakout} environment by displaying the agent's state and actions in longitudinal scatter plots, where the x-coordinate corresponds to the time-step in the episode, and the y-coordinate corresponds to the horizontal position in the environment. The color of the marker encodes one of three actions, either \textit{left}, \textit{right}, or \textit{idle}.

Similar approaches have been used for continuous action spaces, e.g., in Verma et al.~\cite{verma2019programmatically}, which plotted continuous actions against time steps.

\paragraph{State-Action Distributions} While visualizing the state or observation of an agent provides an intuitive understanding of single states, understanding the distribution of such states and their relationship to each other is challenging due to their often high-dimensional nature. Dimensionality reduction methods have been used to project state-action pairs into a low-dimensional latent space, providing an overview of a state-action distribution found, e.g., in replay buffers~\cite{vizarel}. The authors suggest that RL users can obtain a quick overview of the state-action diversity of an agent. 
\defhuman{SD} This approach was leveraged to highlight differences in state-action distribution between different RL agents~\cite{such2019atari}.

\paragraph{Policies} Instead of visualizing the artifacts generated by policies, e.g., state-action distributions as described before, some authors have attempted to provide visual summaries of policies directly. 
\defhuman{PO} Lyu et al.~\cite{lyu2019sdrl} use \textit{symbolic deep reinforcement learning} to formulate policies as symbolic plans, which they then draw into the environment.

\begin{figure*}[tbh]

\begin{annotatedFigure}
	{\includegraphics[width=1.0\linewidth]{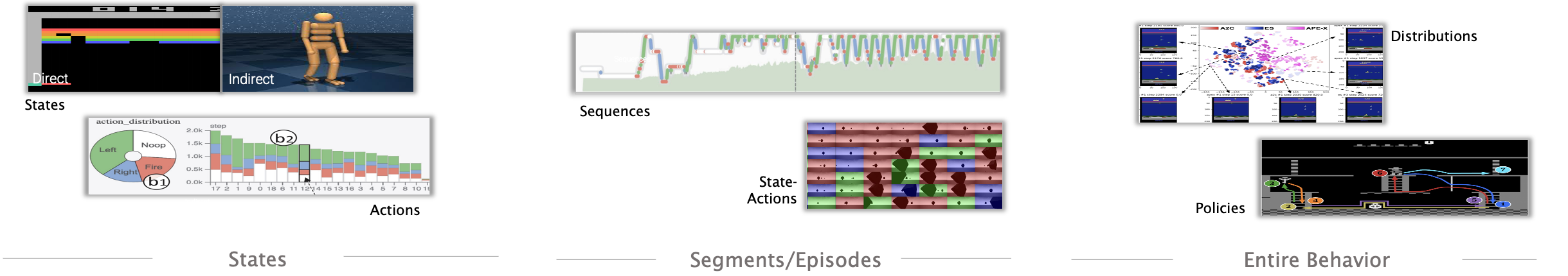}}
 	\annotatedFigureBox{0.018,0.6653}{0.1415,0.9921}{\refhuman{DS}}{0.018,0.6653}%bl
	\annotatedFigureBox{0.141,0.671}{0.267,0.9887}{\refhuman{IS}}{0.141,0.671}%bl
	\annotatedFigureBox{0.054,0.2967}{0.2745,0.578}{\refhuman{AC}}{0.054,0.2967}%bl
	\annotatedFigureBox{0.368,0.6483}{0.6405,0.8785}{\refhuman{ST}}{0.368,0.6483}%bl
	\annotatedFigureBox{0.514,0.2456}{0.6385,0.5633}{\refhuman{SA}}{0.514,0.2456}%bl
	\annotatedFigureBox{0.726,0.5576}{0.8835,0.907}{\refhuman{PO}}{0.726,0.5576}%bl
	\annotatedFigureBox{0.8045,0.2286}{0.9755,0.4985}{\refhuman{SD}}{0.9755,0.2286}%bl
\end{annotatedFigure}

\caption{\textbf{A selection of proposed visualizations for different granularities in RL scenarios} -- several proposed solutions for user interfaces display elements of a reinforcement learning process at different granularities.}
\label{fig:granularity-selection_work}
\end{figure*}
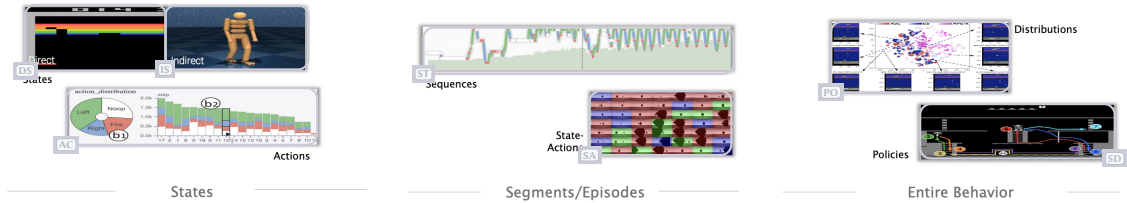

\oppbox{7. Temporal Granularity}{While many approaches for single granularities exist, there is still a gap in the simultaneous visualization of state-action sequences, which restricts effective exploration and selection by humans. While the ability to give feedback at different granularities also positively impacts expressiveness \refhuman{Q1}, }

\subsubsection{Choice Set Size~\refmodel{D8}}
The available choice set is highly dependent on the task domain, but there are choices a designer can make for different feedback.

As with other dimensions, we find a trade-off in the choice set size: A larger choice set is potentially more expressive and informative. Having only a limited set of choices might feel restrictive, i.e., the human user feels like they lack control over the feedback. In terms of information, k-wise ranking compared to binary preferences~\cite{Zhu2023, myers2022learning} can be more informative because a k-wise ranking contains multiple binary preferences. Similarly, the choice of the best action out of a larger set of available actions could be interpreted as ruling out more sub-optimal actions.

Conversely, a larger set of possible choices can lead to user fatigue and higher cognitive demand. Additionally, a larger choice set can also lead to a higher variance in the given values, although it might exhibit a larger bias if the given choices do not match the true desired internal best choice.

\oppbox{8. Choice Set Size}{
The choice set size is partially dependent on the chosen task, and there are apparent designs based on other chosen dimensions (e.g., for a demonstration, the choice set size is dependent on the action space, i.e., from which set of actions I can choose for the demonstration). However, there is potential in varying the choice set to find an optimal setup. For example, rankings might have advantages over pairwise comparisons; similarly, instead of generating a novel trajectory in a continuous space, we might want to select from a set of candidate trajectories instead to improve efficiency \refhuman{Q2}. We might find that a more extensive choice set size might positively influence unbiasedness \refmodel{Q6}, because the human is not forced to provide values in a restricted range. On the other hand, a large choice set size might lead to a higher variance \refmodel{Q6}. Finally, a small choice set size might increase learning speed because the model can be simpler, but choosing from a large choice set could be more precise and, therefore, informative \refmodel{Q7}. A potential remedy is using \textbf{adaptive choice feedback}, which dynamically adapts the choice size during the feedback generation process according to human and model qualities.
}

\subsubsection{Exclusivity~\refmodel{D9}}

Training a new reward model for each human, estimating irrationality or the connection between internal reward and observable behavior is infeasible for most learning scenarios. To overcome this, we can draw inspiration from recommender systems that are challenged with the so-called ``cold-start'' problem, i.e., the initialization of models without any data~\cite{wang2022deep, afsar2022reinforcement}. We can utilize meta-learning \defmodel{ML} strategies to quickly adapt models to the given use case. A way to achieve this would be to train \textit{population} models that, e.g., capture certain feedback types' general irrationality. Based on this model, we could train a \textit{user} model that captures the particularities of the user, e.g., specific ways implicit feedback is communicated. Finally, we might then use a \textit{session} or \textit{task} model that learns a reward function specific to the given task and can also consider session-specific feedback quality.  

\begin{figure}[h]
    \centering
    \includegraphics[width=1\linewidth]{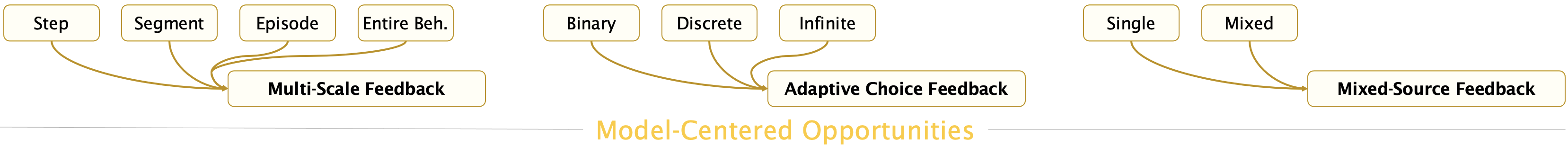}
    \caption{A Summary of Opportunities for Model-Centered Dimensions}
    \label{fig:model_opp_summary}
\end{figure}

\oppbox{9. Exclusivity}{
Experimenting with the use of \textbf{multiple feedback sources} is a promising avenue. When dealing with feedback from multiple human labels, as for large-scale post-training of LLMs \cite{ouyang2022training}, we are natively confronted with non-exclusive reward sources and potential problems like labeler disagreement. This can lead to problems like decreased precision \refhuman{Q5}.\\
Different reward sources might exhibit different biases (like rating behavior based on different internal scales) or different preferences. Modeling different non-exclusive reward sources directly might help to alleviate some issues, actually correcting individual human biases \refhuman{Q6}, and detecting the quality of sources in terms of informativeness \refmodel{Q7}.
}
The opportunities for human-centered human feedback are summarized in~\autoref{fig:model_opp_summary}.

\section{Discussion and Outlook}
The goal of training with tight human-AI interaction is to create powerful, aligned, and trustworthy agents. Considering our running example, Our household robot might now be able to perform various additional tasks, act according to our usual preferences, and be predictable and reliable in its movements and actions. 

As we have outlined in this paper, we think that diverse and complementary feedback types are essential for humans to give expressive feedback, require appropriate effort, enable reasoning about contextual information and uncertainty, and are learnable and informative. Powerful and natural human-AI interactions may only be achieved if diverse interactions and feedback types are blended dynamically and flexibly. Thus, human-AI communication incorporating multiple types of human feedback, expressive reward modeling, querying, and translation are part of a conceptual model for the next steps in RL with human feedback. Future work across different areas can contribute towards the common goal of human-aligned intelligent agents:

\begin{figure}
	%\centering
  \includegraphics[width=.95\linewidth]{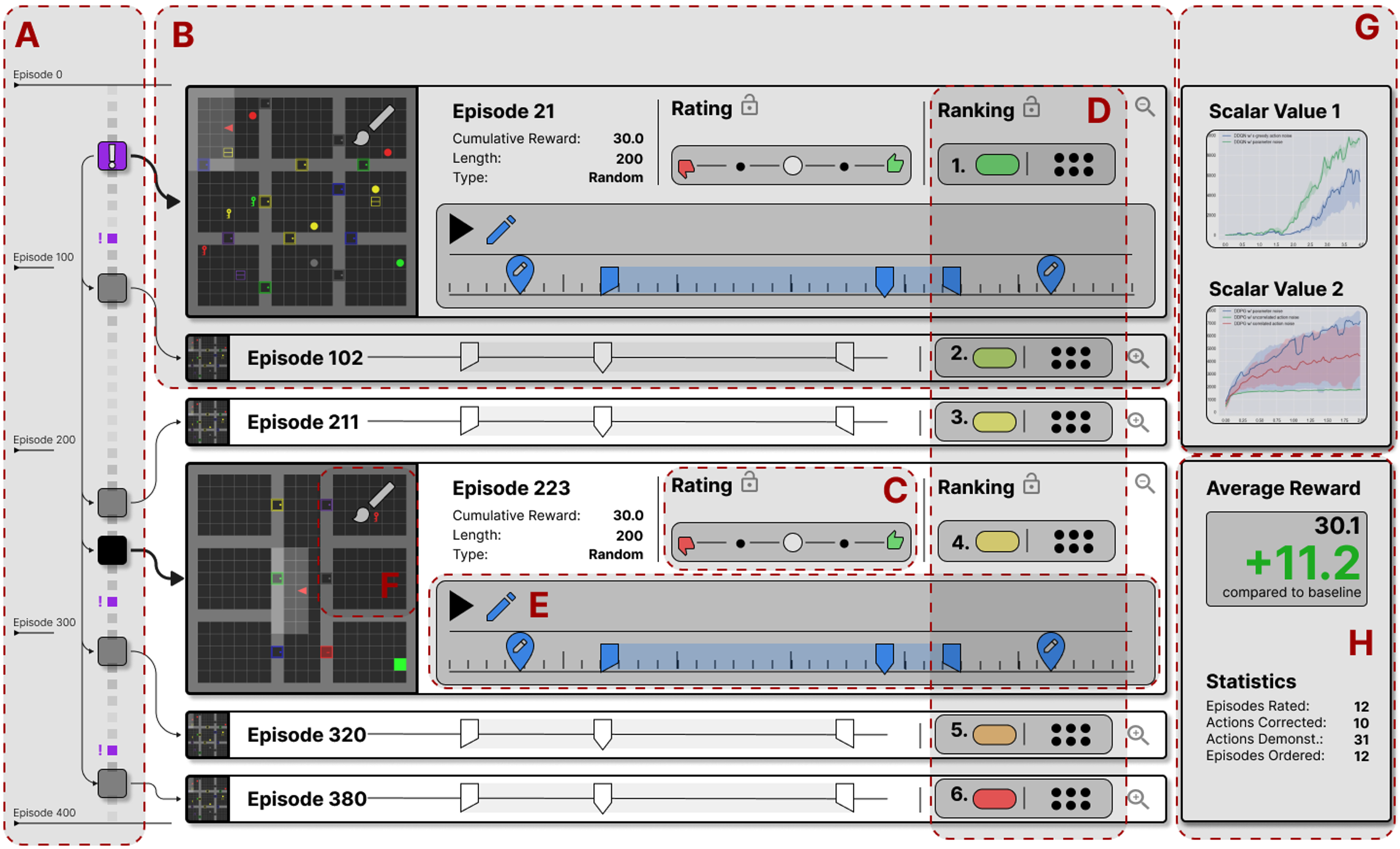}
  \caption{\label{fig:prototype} Prototype implementation of an integrated system for multi-type human feedback in reinforcement learning. (\textit{A}) is a scrollable list of episodes: The user can be queried for feedback by visual indicators, selected episodes are visible, and the user can visit previous episodes to investigate the effect of feedback. In the center column, we have an episode view with multiple feedback interactions (\textit{B}): The user evaluates the episodes via ratings (\textit{C}), ranks via drag \& drop (\textit{D}), select episode segments for correction (\textit{E}), or brush to highlight important features (\textit{F}). Furthermore, the human annotator is informed of global model training via output metrics (\textit{G}). Shown is \textit{BabyAI}~\cite{chevalier-boisvert_babyai_2018}.
    }
\end{figure}

\begin{siderules}[oversize]{HumanColor}
\noindent\textbf{(1)~Explore the Space of Human Feedback}
Widening the space of human feedback types that RL-based systems can use enables more expressive and accessible systems to train AI agents. By using different interactions for feedback, humans have more ways to express their goals and preferences to AI systems, which can improve safety, robustness, and satisfaction on the users' side.
\end{siderules}

\begin{siderules}[oversize]{HumanColor}
\noindent\textbf{(2)~Holistic Visual Interactive Systems}
As we show in our survey, although many isolated solutions exist, there is huge potential in systems that implement a holistic approach for RL from diverse human feedback. Researchers with HCI or visualization backgrounds can contribute systems to evaluate the utility of different feedback types. In \autoref{fig:prototype}, we show a prototype of a possible interface integrating several types of feedback into a common interface.
\end{siderules}

\begin{siderules}[oversize]{ModelColor}
\noindent\textbf{(3)~Towards Expressive Reward Models}
To capture the full fidelity of human feedback, including factors like uncertainty, we need reward models aware of human-centered and structured dimensions of feedback and the user or task context. To achieve this, these models need to receive additional information as input, for example, user and session information, potentially model temporal dynamics, and need the capacity to learn the irrationality or uncertainty of different feedback types. Meta-learning approaches can support adapting expressive models to new users and keep the demand for data and calibration in check.
\end{siderules}

\begin{siderules}[oversize]{CombinedColor}
\noindent\textbf{(4)~Empirical Investigation of Human Feedback}
When estimating the quality of feedback, like informativeness or consistency, we should not rely on gut feeling or scale alone. Instead, we should ground such considerations on resulting modeling decisions on rigorous experimentation with human subjects across diverse backgrounds and tasks~\cite{iml_review_sperrle}. Versatile visual interactive systems that allow for experimentation can be crucial tools to enable this experimentation. We find a critical current research gap in investigating the trade-off between natural language feedback, which was proposed in recent papers~\cite{harrison2017guiding, sumers_linguistic_2022}. However, as has been argued in the space of explainable AI~\cite{sevastjanova2018going}, natural language has its limit, and we might instead view other modalities as complementary. For example, language alone can make it difficult to refer, e.g., to a certain granularity, feature, etc.
\end{siderules}

The given conceptual framework should enable thinking about human feedback in a structured and systematic way and inspire future system design for expressive, efficient, and highly effective learning of reward models. However, there are many learning factors from human feedback that are still not covered, e.g., temporal dynamics, or this framework does not fully discuss co-adaptation. Furthermore, although we developed the framework in an iterative process, there might be other ways to frame the problem space for different use cases. 

\section{Conclusion}
\label{sec:conclusion}
This paper discusses a conceptual framework for human feedback for reinforcement learning in human-AI collaboration. We have presented nine dimensions to classify feedback types, encompassing human-centered, interface-centered, and model-centered dimensions. Additionally, we have presented seven quality criteria for human feedback. We outline the design requirements for interactive systems using reinforcement learning from human feedback for the three core components. We argue that systems using a rich set of possible feedback with expressive reward models, supported by targeted visual interactive interfaces, can lead toward human-aligned intelligent agents.

\bibliographystyle{ACM-Reference-Format}
\bibliography{bibliography}

\newpage

\appendix

\section{Full Survey Results}

\begin{figure}[h]
    \centering
    \includegraphics[width=0.92\linewidth]{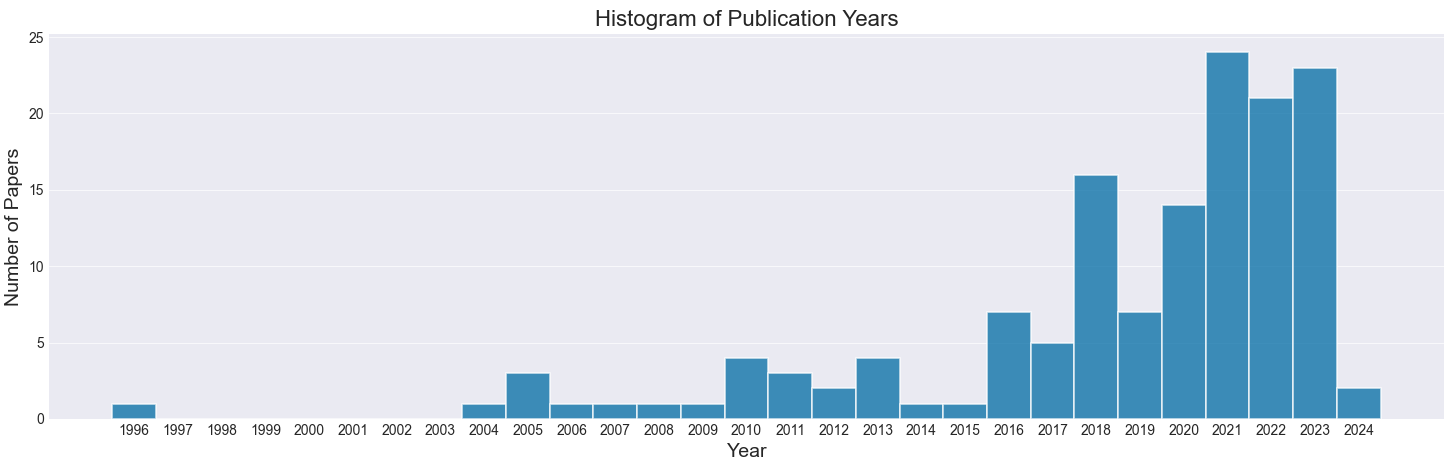}
    \caption{The publication years of surveyed publications.}
    \label{fig:year_distribution}
\end{figure}

\begin{figure}[ht]
    \centering

    % D1: Intent
    \begin{subfigure}[b]{0.3\textwidth}
        \centering
        \includegraphics[width=\textwidth]{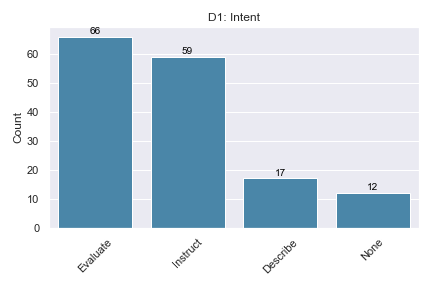}
        \caption{D1: Intent}
    \end{subfigure}
    \hfill
    % D2: Expression
    \begin{subfigure}[b]{0.3\textwidth}
        \centering
        \includegraphics[width=\textwidth]{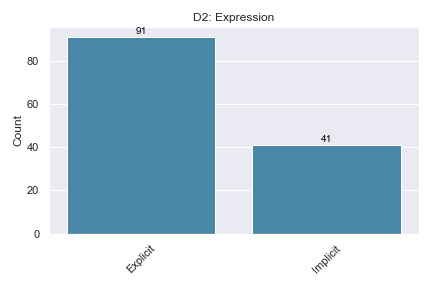}
        \caption{D2: Expression}
    \end{subfigure}
    \hfill
    % D3: Target Actuality
    \begin{subfigure}[b]{0.3\textwidth}
        \centering
        \includegraphics[width=\textwidth]{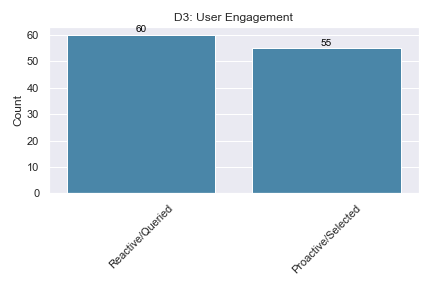}
        \caption{D3: User Engagement}
    \end{subfigure}

    \vspace{0.5cm}
    
    % D4: Target Relation
    \begin{subfigure}[b]{0.3\textwidth}
        \centering
        \includegraphics[width=\textwidth]{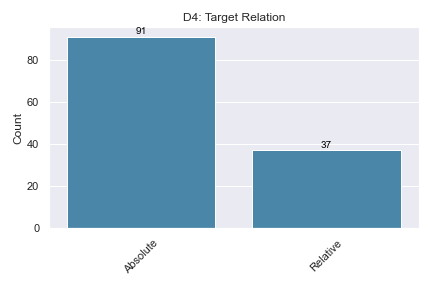}
        \caption{D4: Target Relation}
    \end{subfigure}
    \hfill
    % D5: Content Level
    \begin{subfigure}[b]{0.3\textwidth}
        \centering
        \includegraphics[width=\textwidth]{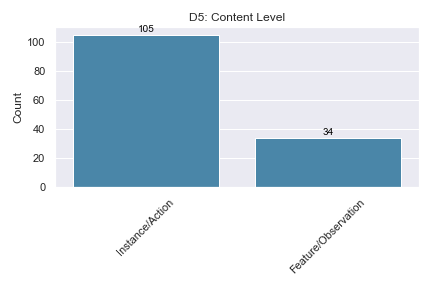}
        \caption{D5: Content Level}
    \end{subfigure}
    \hfill
    % D6: Time Granularity
    \begin{subfigure}[b]{0.3\textwidth}
        \centering
        \includegraphics[width=\textwidth]{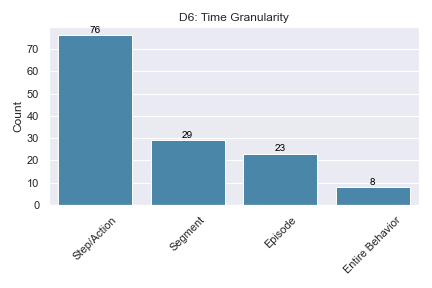}
        \caption{D6: Target Actuality}
    \end{subfigure}

    \vspace{0.5cm}

    % D7: Value Granularity
    \begin{subfigure}[b]{0.3\textwidth}
        \centering
        \includegraphics[width=\textwidth]{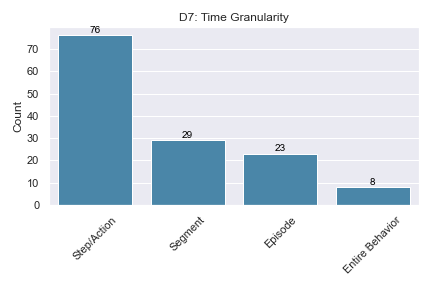}
        \caption{D7: Time Granularity}
    \end{subfigure}
    \hfill
    % D8: User Engagement
    \begin{subfigure}[b]{0.3\textwidth}
        \centering
        \includegraphics[width=\textwidth]{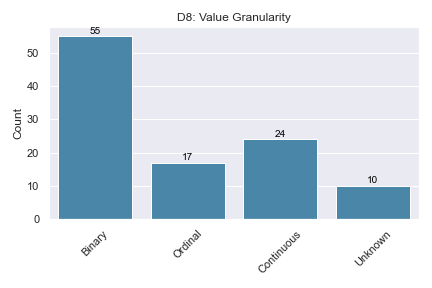}
        \caption{D8: Choice Set Size}
    \end{subfigure}
    \hfill
    % D10: Exclusivity
    \begin{subfigure}[b]{0.3\textwidth}
        \centering
        \includegraphics[width=\textwidth]{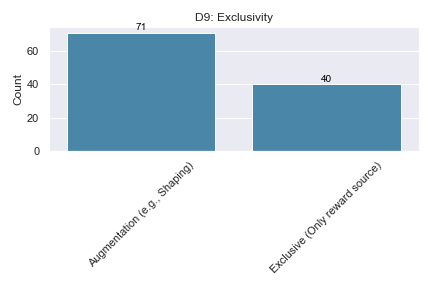}
        \caption{D9: Exclusivity}
    \end{subfigure}

    \caption{Counts of surveyed papers for each attribute in Dimensions D1-D9}
\end{figure}

\begin{landscape}
\begin{table}[]
\centering
\resizebox{1.28\textwidth}{!}{%\
\definecolor{blueish}{HTML}{F4F7F8}
\rowcolors{2}{white}{blueish}
\setlength{\tabcolsep}{1.5pt}
\renewcommand\arraystretch{1.20}
\begin{tabular}{r | *{4}{c} | *{2}{c} | *{2}{c} | *{2}{c} | *{2}{c} | *{2}{c} | *{4}{c} | *{3}{c} | *{2}{c} | *{1}{c} | r | r | r }
    & \multicolumn{4}{|c|}{\textcolor{DarkHumanColor}{\textbf{D1}}} 
    & \multicolumn{2}{c|}{\textcolor{DarkHumanColor}{\textbf{D2}}} 
    & \multicolumn{2}{c|}{\textcolor{DarkHumanColor}{\textbf{D3}}} 
    & \multicolumn{2}{c|}{\textcolor{DarkCombinedColor}{\textbf{D4}}} 
    & \multicolumn{2}{c|}{\textcolor{DarkCombinedColor}{\textbf{D5}}} 
    & \multicolumn{2}{c|}{\textcolor{DarkCombinedColor}{\textbf{D6}}}
    & \multicolumn{4}{c|}{\textcolor{DarkModelColor}{\textbf{D7}}}
    & \multicolumn{3}{c|}{\textcolor{DarkModelColor}{\textbf{D8}}}
    & \multicolumn{2}{c|}{\textcolor{DarkModelColor}{\textbf{D9}}}\\
    & \multicolumn{4}{|c|}{\textcolor{DarkHumanColor}{\textbf{Intent}}} 
    & \multicolumn{2}{c|}{\textcolor{DarkHumanColor}{\textbf{Expres.}}} 
    & \multicolumn{2}{c|}{\textcolor{DarkHumanColor}{\textbf{Engag.}}} 
    & \multicolumn{2}{c|}{\textcolor{DarkCombinedColor}{\textbf{Rela.}}} 
    & \multicolumn{2}{c|}{\textcolor{DarkCombinedColor}{\textbf{Content}}} 
    & \multicolumn{2}{c|}{\textcolor{DarkCombinedColor}{\textbf{Tgt.Act.}}} 
    & \multicolumn{4}{c|}{\textcolor{DarkModelColor}{\textbf{Tmp.Gra.}}} 
    & \multicolumn{3}{c|}{\textcolor{DarkModelColor}{\textbf{Choi.Set}}} 
    & \multicolumn{2}{c|}{\textcolor{DarkModelColor}{\textbf{Excls.}}} 
    \\
    \cmidrule{2-24}
    Publication 
    & \rot{Evaluate} & \rot{Instruct} & \rot{Describe} & \rot{None} 
    & \rot{Explicit} & \rot{Implicit} 
    & \rot{Proactive} & \rot{Reactive} 
    & \rot{Absolute} & \rot{Relative} 
    & \rot{Instance} & \rot{Feature} 
    & \rot{Actual} & \rot{Hypothetical} 
    & \rot{Step/Action} & \rot{Segment} & \rot{Episode} & \rot{Ent.Beh.}
    & \rot{Binary} & \rot{Discrete} & \rot{Continuous} 
    & \rot{Exclusive} & \rot{Augmenting} & Description & Year & Venue \\
% Your complex header structure goes here
\midrule
Maclin et al.~\cite{maclin1996creating} &  & \OK &  &  & \OKGREY & \OKGREY & \OK &  & \OK &  &  & \OK & \OKGREY & \OKGREY &  &  &  & \OK &  &  &  &  & \OK & Suggestions via Instructions & 1996 & Machine Learning \\
Kuhlmann et al.~\cite{kuhlmann2004guiding} &  &  & \OK &  & \OK &  & \OK &  & \OK &  &  & \OK &  & \OK & \OK &  &  & \OKGREY &  &  &  &  & \OK & If-Then-Rules & 2004 & AAAI Workshop \\
Ferrez et al.~\cite{ferrez2005you} &  &  &  & \OK &  & \OK &  & \OK & \OK &  & \OK &  & \OK &  & \OK &  &  &  &  &  & \OK & \OK &  & Interaction error-related potentials & 2005 & IJCAI \\
Maclin et al.~\cite{maclin2005giving} &  & \OK &  &  & \OK &  &  & \OK &  & \OK & \OK &  & \OK &  & \OK &  &  &  &  & \OK &  &  & \OK & Preference over actions & 2005 & AAAI 2005 \\
Thomaz et al.~\cite{thomaz2005real} & \OK &  &  &  & \OK &  & \OK &  & \OK &  & \OK & \OK & \OK &  & \OK &  &  &  &  &  & \OK &  & \OK & Natural interaction & 2005 & AAAI \\
Thomaz et al.~\cite{thomaz2006reinforcement} & \OK &  &  &  & \OK &  &  &  &  & \OK & \OK &  & \OK &  & \OK &  &  &  & \OKGREY & \OKGREY &  &  &  & Action Preferece (Wirth 2018) & 2006 &  \\
Thomaz et al.~\cite{thomaz2007asymmetric} & \OK &  &  &  & \OK &  & \OK &  &  &  & \OK &  & \OK &  & \OKGREY & \OKGREY &  &  &  &  &  &  &  & Positive/Negative Scalar Feedback & 2007 & IEEE RO-MAN \\
Bradley Knox et al.~\cite{Knox2008} & \OK &  &  &  & \OK &  &  & \OK & \OK &  & \OK &  & \OK &  & \OK &  &  &  &  &  & \OK & \OK &  & Scalar Reward & 2008 & IEEE ICDL \\
Branavan et al.~\cite{branavan2009reinforcement} &  & \OKGREY & \OKGREY &  & \OK &  & \OK &  & \OK &  & \OK &  &  & \OK & \OK &  & \OKGREY &  &  &  &  & \OK &  & Action plans & 2009 & ACL \\
Zucker et al.~\cite{zucker2010optimization} & \OK &  &  &  & \OK &  &  &  &  & \OK &  & \OK & \OK &  & \OK &  &  &  & \OKGREY & \OKGREY &  &  &  & State Preference (Wirth 2018) & 2010 &  \\
Knox et al.~\cite{knox2010combining} & \OK &  &  &  & \OK &  &  & \OK &  &  & \OK &  & \OK &  & \OK &  &  &  &  &  & \OK &  &  & Shaping & 2010 &  \\
Tenorio-Gonzalez et al.~\cite{tenorio2010dynamic} & \OK & \OK &  &  &  & \OK & \OK &  & \OK &  & \OK &  & \OK & \OKGREY & \OK &  &  &  & \OK &  &  &  & \OK & Voice Commands (Good/Bad, Action Adv.) & 2010 & IBERAMIA \\
Judah et al.~\cite{judah2010reinforcement} & \OK &  &  &  & \OK &  &  & \OK & \OK &  & \OK &  & \OK &  & \OK &  &  &  & \OKGREY &  &  &  & \OK & Critique Advice & 2010 & AAAI \\
Cakmak et al.~\cite{cakmak2011human} &  & \OK &  &  &  & \OK & \OK & \OK &  & \OKGREY & \OK & \OK &  & \OK &  &  & \OK &  & \OK &  &  &  & \OK & Examples for Bad and Good Hand-Overs & 2011 & IROS 2011 \\
Taylor et al.~\cite{taylor2011integrating} &  & \OK &  &  & \OK &  & \OK &  & \OK &  & \OK &  & \OK &  & \OK &  &  &  & \OK &  &  &  & \OK & Demonstrations, Interactive Action-Advice & 2011 & AAMAS \\
Cakmak et al.~\cite{cakmak2011human} & \OK &  &  &  & \OK &  &  & \OK &  & \OK &  & \OK & \OK &  &  &  & \OK &  & \OK &  &  &  & \OK & Preferences over Configurations & 2011 & IROS 2011 \\
Knox et al.~\cite{knox2012reinforcement} & \OK &  &  &  & \OK &  &  & \OK & \OK &  & \OK &  & \OK &  & \OK & \OKGREY &  &  &  &  &  &  & \OK & Shaping & 2012 & IEEE RO-MAN \\
Akrour et al.~\cite{akrour2012april} & \OK &  &  &  & \OK &  & \OK &  &  & \OK & \OK &  & \OK &  &  & \OK & \OK &  & \OK &  &  &  &  & Preference & 2012 & ECML/PKDD \\
Hayes-Roth et al.~\cite{hayes2013advice} &  &  & \OK &  & \OK &  & \OK &  & \OK &  &  & \OK &  & \OK &  &  & \OK &  &  &  &  & \OK &  & General constraints & 2013 &  \\
Li et al.~\cite{li2013using} & \OK &  &  &  & \OK &  & \OK &  & \OK &  & \OK &  & \OK &  & \OK &  &  &  & \OK &  &  &  & \OK & Feedback & 2013 & AAMAS \\
Griffith et al.~\cite{griffith2013policy} &  &  &  &  &  &  &  &  &  &  &  &  &  &  &  &  &  &  &  &  &  &  &  &  & 2013 &  \\
Griffith et al.~\cite{griffith2013policy} & \OK &  &  &  & \OK &  &  & \OK &  &  & \OK &  & \OK &  & \OK &  &  &  & \OK &  &  &  &  & Shaping & 2013 & NeurIPS \\
Loftin et al.~\cite{loftin2014learning} & \OK &  &  &  & \OK &  & \OK &  & \OK &  & \OK &  & \OK &  & \OK &  &  &  & \OK &  &  & \OKGREY &  & Feedback signals & 2014 & IEEE ROMAN \\
Odom et al.~\cite{odom2015active} &  & \OK &  &  & \OK &  &  & \OK & \OK &  &  & \OK & \OK & \OKGREY &  &  &  & \OK &  &  &  &  &  & Advice & 2015 & AAAI \\
Amir et al.~\cite{Amir2016} &  & \OK &  &  & \OK &  & \OKGREY & \OKGREY & \OK &  & \OK &  & \OK &  & \OK &  &  &  & \OK &  &  &  & \OK & (Action) Advice & 2016 & IJCAI 2016 \\
Subramanian et al.~\cite{subramanian2016exploration} &  & \OK &  &  & \OK &  & \OK &  & \OK &  & \OK &  &  & \OK &  &  & \OK &  &  &  &  &  &  & Demonstrations & 2016 &  \\
Raza et al.~\cite{raza2016reward} &  & \OK &  &  & \OK &  &  & \OK & \OK &  & \OK &  & \OK &  & \OK &  &  &  & \OK &  &  &  & \OK & Demonstrations/action advice & 2016 & AAAI \\
Loftin et al.~\cite{loftin2016learning} & \OK &  &  &  & \OK & \OKGREY & \OK &  &  &  & \OK &  & \OK &  & \OK &  &  &  & \OK &  &  & \OKGREY & \OKGREY & Discrete Feedb. with diff. training strategies & 2016 &             
AAMAS \\
\end{tabular}%
    }
    \caption{Summary table of the literature survey. Surveyed work is classified according to the conceptual framework (\OK) Black checkmarks refer to exclusive attributes, (\OKGREY) Grey checkmarks indicate that feedback can have characteristics across different work, (\OKBLUE) Blue checkmarks  indicate necessary simultaneous presence of two attributes. We can classify existing types of feedback according to our dimensions.}
    \label{tab:surveyed_papers_dimensions}
    \end{table}
    \end{landscape}

\begin{landscape}
\begin{table}[]
\centering
\resizebox{1.28\textwidth}{!}{%\
\definecolor{blueish}{HTML}{F4F7F8}
\rowcolors{2}{white}{blueish}
\setlength{\tabcolsep}{1.5pt}
\renewcommand\arraystretch{1.20}
\begin{tabular}{r | *{4}{c} | *{2}{c} | *{2}{c} | *{2}{c} | *{2}{c} | *{2}{c} | *{4}{c} | *{3}{c} | *{2}{c} | *{1}{c} | c | c | c }
    & \multicolumn{4}{|c|}{\textcolor{DarkHumanColor}{\textbf{D1}}} 
    & \multicolumn{2}{c|}{\textcolor{DarkHumanColor}{\textbf{D2}}} 
    & \multicolumn{2}{c|}{\textcolor{DarkHumanColor}{\textbf{D3}}} 
    & \multicolumn{2}{c|}{\textcolor{DarkCombinedColor}{\textbf{D4}}} 
    & \multicolumn{2}{c|}{\textcolor{DarkCombinedColor}{\textbf{D5}}} 
    & \multicolumn{2}{c|}{\textcolor{DarkCombinedColor}{\textbf{D6}}}
    & \multicolumn{4}{c|}{\textcolor{DarkModelColor}{\textbf{D7}}}
    & \multicolumn{3}{c|}{\textcolor{DarkModelColor}{\textbf{D8}}}
    & \multicolumn{2}{c|}{\textcolor{DarkModelColor}{\textbf{D9}}}\\
    & \multicolumn{4}{|c|}{\textcolor{DarkHumanColor}{\textbf{Intent}}} 
    & \multicolumn{2}{c|}{\textcolor{DarkHumanColor}{\textbf{Expres.}}} 
    & \multicolumn{2}{c|}{\textcolor{DarkHumanColor}{\textbf{Engag.}}} 
    & \multicolumn{2}{c|}{\textcolor{DarkCombinedColor}{\textbf{Rela.}}} 
    & \multicolumn{2}{c|}{\textcolor{DarkCombinedColor}{\textbf{Content}}} 
    & \multicolumn{2}{c|}{\textcolor{DarkCombinedColor}{\textbf{Tgt.Act.}}} 
    & \multicolumn{4}{c|}{\textcolor{DarkModelColor}{\textbf{Tmp.Gra.}}} 
    & \multicolumn{3}{c|}{\textcolor{DarkModelColor}{\textbf{Choi.Set}}} 
    & \multicolumn{2}{c|}{\textcolor{DarkModelColor}{\textbf{Excls.}}} 
    \\
    \cmidrule{2-24}
    Publication 
    & \rot{Evaluate} & \rot{Instruct} & \rot{Describe} & \rot{None} 
    & \rot{Explicit} & \rot{Implicit} 
    & \rot{Proactive} & \rot{Reactive} 
    & \rot{Absolute} & \rot{Relative} 
    & \rot{Instance} & \rot{Feature} 
    & \rot{Actual} & \rot{Hypothetical} 
    & \rot{Step/Action} & \rot{Segment} & \rot{Episode} & \rot{Ent.Beh.}
    & \rot{Binary} & \rot{Discrete} & \rot{Continuous} 
    & \rot{Exclusive} & \rot{Augmenting} & Description & Year & Venue \\
% Your complex header structure goes here
\midrule
El Asri et al.~\cite{elAsri2016} & \OK &  &  &  & \OK &  &  & \OK &  & \OK & \OK &  & \OK &  &  &  & \OK &  &  & \OK &  & \OK &  & Scores & 2016 & AAMAS \\
S{\o}rensen et al.~\cite{sorensen2016breeding} & \OK &  &  &  & \OK &  &  & \OK &  & \OK & \OK &  & \OK &  &  & \OK &  & \OKGREY &  &  &  & \OKGREY &  & Interactive Evolution via Pref. Selection & 2016 & IEEE CIG \\
Veeriah et al.~\cite{veeriah2016face} & \OK &  &  &  &  & \OK &  & \OK & \OK &  & \OK &  & \OK &  & \OKGREY & \OKGREY &  &  &  &  & \OKGREY &  & \OK & Facial Landmark Detect. as Observations & 2016 & IJCAI 2016 - IML Workshop \\
Fachantidis et al.~\cite{fachantidis2017learning} &  & \OK &  &  & \OK &  & \OK &  & \OK &  & \OK &  & \OK &  & \OK &  &  &  & \OK &  &  &  & \OK & Action Advice in Teaching Task & 2017 & ML and Knowledge Extr. \\
MacGlashan et al.~\cite{macglashan2017interactive} & \OK &  &  &  & \OK &  & \OK &  & \OK &  & \OK &  & \OK &  & \OK &  &  &  &  &  &  &  &  & Policy-Pependent Feedback & 2017 & ICML \\
Kim et al.~\cite{kim2017intrinsic} & \OKGREY &  &  & \OK &  & \OK & \OKGREY & \OKGREY & \OK &  & \OK &  & \OK &  & \OK &  &  &  &  &  & \OK & \OK &  & Error-related potentials & 2017 & Scientific reports \\
Christiano et al.~\cite{christiano2017deep} & \OK &  &  &  & \OK &  &  & \OK &  & \OK & \OK &  & \OK &  &  & \OK &  &  & \OKGREY & \OKGREY &  &  &  & Trajectory Preference (Wirth 2018) & 2017 &  \\
Lin et al.~\cite{lin2017explore} &  & \OK &  &  & \OK &  &  & \OKGREY & \OK &  & \OK &  & \OK &  & \OK &  &  &  & \OK &  &  &  & \OK & Action Advice & 2017 & ArXiv \\
Gao et al.~\cite{gao2018april} & \OK &  &  &  & \OK &  &  &  &  &  & \OK &  & \OK &  & \OK &  & \OK &  &  &  &  &  &  & Preferences & 2018 &  \\
Li et al.~\cite{li2018interactive} & \OK & \OK &  &  & \OK &  &  & \OK & \OK &  & \OK &  & \OK & \OK & \OK &  &  &  & \OK &  &  &  & \OK & Demonstrations, Scalar (Binary) Feedback & 2018 & IEEE ROMAN  \\
Krening et al.~\cite{krening2018newtonian} &  & \OK &  &  &  & \OK & \OKGREY & \OKGREY & \OK &  & \OK &  & \OK &  & \OK &  &  &  & \OK &  &  &  &  & Newtonian Action Advice & 2018 & AAMAS'19 \\
Torabi et al.~\cite{Torabi2018} &  & \OK &  & \OK &  & \OK &  & \OKGREY & \OK &  & \OKGREY & \OKGREY & \OK &  & \OK &  &  &  &  &  & \OK & \OK &  & Demonstrations & 2018 & IJCAI 2018 \\
Krening et al.~\cite{krening2018interaction} & \OK & \OK &  &  & \OK &  &  & \OK & \OK &  & \OK &  & \OK &  & \OK &  &  &  & \OK &  &  &  & \OK & Active Advice/Binary Critique & 2018 & ACM THRI \\
Ibarz et al.~\cite{ibarz2018reward} & \OK & \OK &  &  & \OK &  & \OKGREY & \OKGREY & \OK & \OK & \OK &  & \OK &  &  & \OK & \OKGREY &  & \OK & \OK &  & \OK &  & Demos and Pairwise Pref. & 2018 & NeurIPS \\
Arakawa et al.~\cite{arakawa2018dqn} &  &  &  & \OK &  & \OK &  & \OK & \OKGREY & \OKGREY & \OKGREY & \OKGREY & \OK &  &  & \OK &  &  &  &  & \OK &  & \OK & Facial Expressions & 2018 &  \\
Mindermann et al.~\cite{mindermann2018active} & \OKGREY &  & \OKGREY &  & \OK &  &  & \OK &  & \OK &  & \OK &  & \OKGREY &  &  &  & \OK &  &  &  & \OKGREY &  & Discrete Queries: Reward Function Selection. & 2018 & NeurIPS \\
Brown et al.~\cite{brown2018risk} &  & \OK &  &  & \OK &  &  & \OK & \OK &  & \OK &  &  & \OK &  &  & \OK &  &  &  &  &  &  & Demonstrations & 2018 & ICML \\
Losey et al.~\cite{losey2018including} & \OK &  &  &  & \OKGREY & \OKGREY &  & \OK &  & \OK & \OK &  & \OK &  & \OK &  &  &  & \OKGREY &  & \OKGREY & \OK &  & Corrections & 2018 & CoRL 2021 \\
Yeh et al.~\cite{yeh2018bridging} &  & \OK &  &  &  & \OK & \OK &  & \OK &  & \OKGREY &  & \OK &  &  &  & \OK &  &  &  & \OK &  & \OK & (Eval.) Binary Advice, (Informative) Advice & 2018 & Adv. in Cognitive Sys. \\
Cruz et al.~\cite{cruz2018multi} &  & \OK &  &  &  & \OK & \OK &  & \OK &  &  & \OK & \OK &  & \OK &  &  &  &  &  &  & \OK &  & Multi-Modal/Multi-Sensory Advice & 2018 & IJCNN \\
Basu et al.~\cite{basu2018learning} & \OKGREY &  & \OKGREY &  & \OK &  &  & \OK &  & \OK &  & \OK & \OKGREY & \OKGREY &  &  & \OK &  &  &  &  & \OKGREY &  & Feature Queries & 2018 & HRI \\
Lon{\v{c}}arevi{\'c} et al.~\cite{lonvcarevic2018user} & \OK &  &  &  & \OK &  &  & \OK & \OK &  & \OK &  & \OK &  &  &  & \OK &  &  & \OK &  & \OK &  & Exact, Unsigned and Signed Reward & 2018 & IEEE-RAS \\
Ibarz et al.~\cite{ibarz2018reward} & \OK &  &  &  & \OK &  &  & \OK & \OK & \OK & \OK &  & \OK & \OKGREY & \OK & \OK &  &  & \OK &  &  & \OK &  & Preferences and demonstrations & 2018 & NeurIPS \\
Warnell et al.~\cite{warnell2018deep} & \OK &  &  &  & \OK &  &  &  & \OK &  & \OK &  & \OK &  & \OK &  &  &  &  & \OK &  &  & \OK & Scalar Feedback (positive/negative)  & 2018 & AAAI \\
Mill{\'a}n et al.~\cite{millan2019human} & \OK &  &  &  & \OK &  & \OKGREY &  & \OK &  & \OK &  & \OK &  & \OK &  &  &  &  &  &  &  & \OK & Boolean-Evalution for Shaping & 2019 & Biomimetics \\
Basu et al.~\cite{basu2019active} & \OK &  &  &  & \OK &  &  & \OK &  & \OK & \OK &  & \OK &  &  & \OK &  &  & \OK &  &  & \OK &  & Hierarch. Pref. Queries, Pairwise comparisons & 2019 & IEEE/RJS IROS \\
De Winter et al.~\cite{de2019accelerating} &  & \OKGREY & \OK &  &  & \OK & \OK &  & \OK &  &  & \OK & \OKGREY & \OKGREY &  &  &  & \OKGREY &  &  &  &  & \OK & Constraints & 2019 & Robotics (MDPI) \\
Akkaladevi et al.~\cite{akkaladevi2019towards} &  & \OK &  &  & \OK &  &  & \OK &  & \OK &  &  & \OK &  & \OK &  &  &  & \OK &  &  & \OK &  & Action Choice & 2019 & Procedia Manufacturing \\
Wang et al.~\cite{Wang2019interactive} &  & \OK &  &  & \OKGREY &  & \OK &  &  &  & \OK &  &  & \OK & \OK &  &  &  &  &  &  & \OK &  & Prior Demonstrations & 2019 & IJCAI \\
Frazier et al.~\cite{frazier2019improving} &  & \OK &  &  & \OK &  &  & \OK & \OK &  & \OK &  & \OK &  & \OK &  &  &  & \OK &  &  &  & \OK & Action Advice & 2019 & AAAI AIIDE-19 \\
\end{tabular}%
    }
    \caption{Summary table of the literature survey. Surveyed work is classified according to the conceptual framework (\OK) Black checkmarks refer to exclusive attributes, (\OKGREY) Grey checkmarks indicate that feedback can have characteristics across different work, (\OKBLUE) Blue checkmarks  indicate necessary simultaneous presence of two attributes. We can classify existing types of feedback according to our dimensions.}
    \label{tab:surveyed_papers_dimensions}
    \end{table}
    \end{landscape}

\begin{landscape}
\begin{table}[]
\centering
\resizebox{1.25\textwidth}{!}{%\
\definecolor{blueish}{HTML}{F4F7F8}
\rowcolors{2}{white}{blueish}
\setlength{\tabcolsep}{1.5pt}
\renewcommand\arraystretch{1.20}
\begin{tabular}{r | *{4}{c} | *{2}{c} | *{2}{c} | *{2}{c} | *{2}{c} | *{2}{c} | *{4}{c} | *{3}{c} | *{2}{c} | *{1}{c} | c | c | c }
    & \multicolumn{4}{|c|}{\textcolor{DarkHumanColor}{\textbf{D1}}} 
    & \multicolumn{2}{c|}{\textcolor{DarkHumanColor}{\textbf{D2}}} 
    & \multicolumn{2}{c|}{\textcolor{DarkHumanColor}{\textbf{D3}}} 
    & \multicolumn{2}{c|}{\textcolor{DarkCombinedColor}{\textbf{D4}}} 
    & \multicolumn{2}{c|}{\textcolor{DarkCombinedColor}{\textbf{D5}}} 
    & \multicolumn{2}{c|}{\textcolor{DarkCombinedColor}{\textbf{D6}}}
    & \multicolumn{4}{c|}{\textcolor{DarkModelColor}{\textbf{D7}}}
    & \multicolumn{3}{c|}{\textcolor{DarkModelColor}{\textbf{D8}}}
    & \multicolumn{2}{c|}{\textcolor{DarkModelColor}{\textbf{D9}}}\\
    & \multicolumn{4}{|c|}{\textcolor{DarkHumanColor}{\textbf{Intent}}} 
    & \multicolumn{2}{c|}{\textcolor{DarkHumanColor}{\textbf{Expres.}}} 
    & \multicolumn{2}{c|}{\textcolor{DarkHumanColor}{\textbf{Engag.}}} 
    & \multicolumn{2}{c|}{\textcolor{DarkCombinedColor}{\textbf{Rela.}}} 
    & \multicolumn{2}{c|}{\textcolor{DarkCombinedColor}{\textbf{Content}}} 
    & \multicolumn{2}{c|}{\textcolor{DarkCombinedColor}{\textbf{Tgt.Act.}}} 
    & \multicolumn{4}{c|}{\textcolor{DarkModelColor}{\textbf{Tmp.Gra.}}} 
    & \multicolumn{3}{c|}{\textcolor{DarkModelColor}{\textbf{Choi.Set}}} 
    & \multicolumn{2}{c|}{\textcolor{DarkModelColor}{\textbf{Excls.}}} 
    \\
    \cmidrule{2-25}
    Publication 
    & \rot{Evaluate} & \rot{Instruct} & \rot{Describe} & \rot{None} 
    & \rot{Explicit} & \rot{Implicit} 
    & \rot{Proactive} & \rot{Reactive} 
    & \rot{Absolute} & \rot{Relative} 
    & \rot{Instance} & \rot{Feature} 
    & \rot{Actual} & \rot{Hypothetical} 
    & \rot{Step/Action} & \rot{Segment} & \rot{Episode} & \rot{Ent.Beh.}
    & \rot{Binary} & \rot{Discrete} & \rot{Continuous} 
    & \rot{Exclusive} & \rot{Augmenting} & Description & Year & Venue \\
% Your complex header structure goes here
\midrule
Zhang et al.~\cite{zhang2019leveraging} &  &  &  & \OK &  & \OK &  & \OK & \OK &  &  & \OK & \OK &  & \OK &  &  &  &  &  & \OK &  & \OK & Learning from Attention & 2019 & IJCAI \\
Arzate Cruz et al.~\cite{arzate_cruz_survey_2020} & \OK &  &  &  & \OK &  & \OKGREY & \OK & \OK & \OKGREY & \OK &  & \OK &  & \OK &  &  &  & \OK &  &  & \OKGREY & \OKGREY & Action-Advice & 2020 & 2020 ACM DIS \\
Arzate Cruz et al.~\cite{arzate_cruz_survey_2020} & \OK &  &  &  & \OK &  &  & \OK & \OK &  & \OK &  & \OK &  & \OK & \OKGREY & \OKGREY &  &  &  & \OK & \OKGREY & \OKGREY & Scalar-Valued Rating & 2020 & 2020 ACM DIS \\
Wilde et al.~\cite{wilde2020improving} & \OK &  &  &  & \OK &  &  & \OK &  & \OK & \OK &  & \OK &  &  &  & \OK &  & \OK &  &  &  & \OKGREY & Active Preference Learning & 2020 & The Int. Journ. of Robotics Res \\
Faulkner et al.~\cite{faulkner2020interactive} & \OK &  &  &  & \OK &  &  &  & \OK &  & \OK &  & \OK &  & \OK &  &  &  &  &  & \OK &  & \OK & Noisy feedback & 2020 & IEEE ICRA \\
Arzate Cruz et al.~\cite{arzate2020mariomix} & \OK &  &  &  & \OK &  & \OK &  &  & \OK & \OK &  & \OK &  &  & \OK &  &  &  &  &  & \OK &  & Assignment of prefered behavior to segment & 2020 & CHI PLAY '20 \\
Koert et al.~\cite{koert2020multi} & \OK & \OK & \OKGREY &  & \OK &  & \OK &  & \OK &  & \OK & \OKGREY & \OK &  & \OK &  &  & \OKGREY & \OK &  &  &  & \OK & Action Advice, State Mod. and Subgoal Reward Def. & 2020 & Frontiers Robotics/AI \\
Arzate Cruz et al.~\cite{arzate_cruz_survey_2020} & \OK &  &  &  & \OK &  &  & \OK & \OK &  & \OK &  & \OK &  & \OK & \OKGREY & \OKGREY &  & \OK &  &  & \OKGREY & \OKGREY & Critique & 2020 & 2020 ACM DIS \\
Li et al.~\cite{li2020facial} & \OKGREY &  &  & \OK & \OKGREY & \OK &  & \OK & \OK &  & \OK &  & \OK &  & \OKGREY & \OKGREY &  &  & \OK &  &  &  & \OK & Keypress feedback and facial expressions & 2020 & Auton. Agents and M-A Sys. \\
Moreira et al.~\cite{moreira2020deep} &  & \OK &  &  & \OK &  &  & \OK & \OK &  & \OK &  & \OK &  & \OK &  &  &  & \OK &  &  & \OKGREY & \OKGREY & Corrective Feedback/Advice & 2020 & Applied Sciences \\
Zhang et al.~\cite{zhang2020atari} &  &  &  & \OK &  & \OK &  & \OK & \OK &  &  & \OK & \OK &  & \OK &  &  &  &  &  & \OK &  & \OK & Gaze for attention guidance & 2020 & AAAI 2020 \\
Arzate Cruz et al.~\cite{arzate_cruz_survey_2020} &  &  & \OK &  & \OK &  & \OKGREY & \OK & \OK &  &  & \OK & \OK &  & \OKGREY &  & \OK &  & \OK & \OKGREY &  & \OKGREY & \OKGREY & Guidance & 2020 & 2020 ACM DIS \\
Raza et al.~\cite{raza2020human} &  & \OK &  &  & \OK &  & \OK &  & \OK &  & \OK &  &  & \OK & \OK &  &  &  & \OK &  &  &  & \OK & Action Assignment/Pseudo Demonstrations & 2020 & ACM TAAS \\
Menner et al.~\cite{menner2020using} & \OK &  &  &  & \OK &  &  & \OK &  &  & \OK &  & \OK &  &  &  & \OK &  &  & \OK &  & \OK &  & Therapist Ratings & 2020 & IEEE Transactions on robotics \\
Wilde et al.~\cite{wilde2020improving} &  & \OKGREY & \OK &  & \OK &  & \OK &  & \OK &  &  & \OK &  & \OK &  &  &  & \OK &  &  &  &  & \OKGREY & Constraints (on robot movement) & 2020 & The Int. Journ. of Robotics Res \\
Bignold et al.~\cite{bignold2021persistent} & \OKGREY &  & \OK &  & \OK &  & \OK &  & \OK &  & \OKGREY & \OK & \OKGREY & \OK & \OK &  &  & \OK &  &  &  &  & \OK & Persistent rule-based advice & 2021 & Neural Comp. and App. \\
Chetouani et al.~\cite{chetouani2021interactive} & \OKGREY & \OK & \OKGREY &  & \OKGREY & \OKGREY & \OKGREY & \OKGREY & \OK &  & \OK & \OKGREY & \OK & \OK & \OK &  &  &  &  &  &  & \OKGREY & \OKGREY & Teaching Signals: Instructions & 2021 &  \\
Najar et al.~\cite{najar2021reinforcement} & \OK & \OK &  &  &  &  & \OKGREY & \OK & \OK &  & \OK & \OK & \OK & \OK & \OK & \OK & \OK &  &  &  &  &  & \OK & Contextual advice & 2021 & Frontiers Robotics/AI \\
Li et al.~\cite{li2021learning} &  & \OK &  &  &  & \OK & \OK &  &  & \OK & \OK &  & \OK &  &  &  & \OK &  &  &  & \OK &  & \OK & Physical Correction & 2021 & ICRA 2021 \\
Najar et al.~\cite{najar2021reinforcement} & \OK &  &  &  & \OK &  & \OKGREY & \OKGREY & \OK & \OKGREY & \OK &  & \OKGREY &  & \OKGREY & \OKGREY & \OKGREY &  &  &  &  &  & \OK & Evaluative Feedback/Critique & 2021 & Frontiers Robotics/AI \\
Li et al.~\cite{li2021roial} & \OK &  &  &  & \OK &  &  & \OK & \OK & \OKGREY & \OK &  & \OK &  &  &  & \OKGREY & \OKGREY & \OKGREY & \OK &  & \OK &  & Ordinal feedback and preferences & 2021 & ICRA 2021 \\
Najar et al.~\cite{najar2021reinforcement} &  &  & \OK &  & \OK &  & \OK &  & \OK &  &  & \OK &  & \OK &  &  &  & \OK &  &  &  &  & \OK & General advice & 2021 & Frontiers Robotics/AI \\
Sumers et al.~\cite{sumers2021learning} & \OK & \OK & \OK &  &  & \OK & \OK &  & \OK & \OKGREY & \OK & \OK & \OK & \OK &  &  & \OK &  & \OK &  &  & \OK &  & Evaluative, Imperative and Descriptive Feedback & 2021 & AAAI 2021 \\
Najar et al.~\cite{najar2021reinforcement} & \OK &  &  &  & \OK &  & \OKGREY & \OKGREY & \OK &  & \OK &  & \OK &  & \OKGREY & \OKGREY & \OKGREY &  &  &  &  &  & \OK & Feedback & 2021 & Frontiers Robotics/AI \\
Chetouani et al.~\cite{chetouani2021interactive} & \OK &  &  &  & \OKGREY & \OKGREY & \OKGREY & \OKGREY & \OK &  & \OK &  & \OK &  & \OK &  &  &  &  &  &  & \OKGREY & \OKGREY & Teaching Signals: Feedback & 2021 &  \\
Xu et al.~\cite{xu2021accelerating} &  &  &  & \OK &  & \OK &  & \OK & \OK &  & \OK &  & \OK &  & \OK &  &  &  &  &  & \OK & \OKGREY & \OKGREY & Implicit (and natural) feedback & 2021 & Neurocomputing \\
Law et al.~\cite{law2021hammers} &  &  & \OKGREY &  & \OK &  & \OK &  & \OK &  &  & \OK &  & \OK &  &  &  & \OK &  &  &  &  & \OK & Redesigning the agent’s tool. & 2021 & DIS21 \\
Cui et al.~\cite{cui2021empathic} &  &  &  & \OK &  & \OK & \OKGREY & \OKGREY & \OKGREY &  & \OK &  & \OK &  & \OKGREY & \OKGREY &  &  &  & \OKGREY & \OKGREY &  & \OK & Facial reactions & 2021 & CoRL \\
Millan-Arias et al.~\cite{millan2021robust} &  & \OK &  &  & \OK &  & \OK &  & \OK &  & \OK &  & \OK &  & \OK &  &  &  & \OK &  &  &  & \OK & Advice & 2021 & HAI/IEEE Access \\
Najar et al.~\cite{najar2021reinforcement} &  & \OKGREY & \OK &  & \OK &  & \OKGREY & \OKGREY & \OK &  & \OK &  &  & \OK & \OK &  &  &  &  &  &  &  & \OK & Guidance & 2021 & Frontiers Robotics/AI \\
Najar et al.~\cite{najar2021reinforcement} &  & \OK &  &  & \OK &  & \OKGREY & \OKGREY & \OK &  & \OK &  & \OKGREY & \OKGREY & \OK &  &  &  &  &  &  &  & \OK & Contextual instructions & 2021 & Frontiers Robotics/AI \\
Wilde et al.~\cite{wilde2021learning} & \OK &  &  &  & \OK &  &  & \OK &  & \OK & \OK &  & \OK &  &  & \OKGREY & \OKGREY &  &  &  & \OK & \OK &  & Scale Feedback & 2021 & CoRL \\
\end{tabular}%
    }
    \caption{Summary table of the literature survey. Surveyed work is classified according to the conceptual framework (\OK) Black checkmarks refer to exclusive attributes, (\OKGREY) Grey checkmarks indicate that feedback can have characteristics across different work, (\OKBLUE) Blue checkmarks  indicate necessary simultaneous presence of two attributes. We can classify existing types of feedback according to our dimensions.}
    \label{tab:surveyed_papers_dimensions}
    \end{table}
    \end{landscape}

\begin{landscape}
\begin{table}[]
\centering
\resizebox{1.25\textwidth}{!}{%\
\definecolor{blueish}{HTML}{F4F7F8}
\rowcolors{2}{white}{blueish}
\setlength{\tabcolsep}{1.5pt}
\renewcommand\arraystretch{1.20}
\begin{tabular}{r | *{4}{c} | *{2}{c} | *{2}{c} | *{2}{c} | *{2}{c} | *{2}{c} | *{4}{c} | *{3}{c} | *{2}{c} | *{1}{c} | c | c | c }
    & \multicolumn{4}{|c|}{\textcolor{DarkHumanColor}{\textbf{D1}}} 
    & \multicolumn{2}{c|}{\textcolor{DarkHumanColor}{\textbf{D2}}} 
    & \multicolumn{2}{c|}{\textcolor{DarkHumanColor}{\textbf{D3}}} 
    & \multicolumn{2}{c|}{\textcolor{DarkCombinedColor}{\textbf{D4}}} 
    & \multicolumn{2}{c|}{\textcolor{DarkCombinedColor}{\textbf{D5}}} 
    & \multicolumn{2}{c|}{\textcolor{DarkCombinedColor}{\textbf{D6}}}
    & \multicolumn{4}{c|}{\textcolor{DarkModelColor}{\textbf{D7}}}
    & \multicolumn{3}{c|}{\textcolor{DarkModelColor}{\textbf{D8}}}
    & \multicolumn{2}{c|}{\textcolor{DarkModelColor}{\textbf{D9}}}\\
    & \multicolumn{4}{|c|}{\textcolor{DarkHumanColor}{\textbf{Intent}}} 
    & \multicolumn{2}{c|}{\textcolor{DarkHumanColor}{\textbf{Expres.}}} 
    & \multicolumn{2}{c|}{\textcolor{DarkHumanColor}{\textbf{Engag.}}} 
    & \multicolumn{2}{c|}{\textcolor{DarkCombinedColor}{\textbf{Rela.}}} 
    & \multicolumn{2}{c|}{\textcolor{DarkCombinedColor}{\textbf{Content}}} 
    & \multicolumn{2}{c|}{\textcolor{DarkCombinedColor}{\textbf{Tgt.Act.}}} 
    & \multicolumn{4}{c|}{\textcolor{DarkModelColor}{\textbf{Tmp.Gra.}}} 
    & \multicolumn{3}{c|}{\textcolor{DarkModelColor}{\textbf{Choi.Set}}} 
    & \multicolumn{2}{c|}{\textcolor{DarkModelColor}{\textbf{Excls.}}} 
    \\
    \cmidrule{2-25}
    Publication 
    & \rot{Evaluate} & \rot{Instruct} & \rot{Describe} & \rot{None} 
    & \rot{Explicit} & \rot{Implicit} 
    & \rot{Proactive} & \rot{Reactive} 
    & \rot{Absolute} & \rot{Relative} 
    & \rot{Instance} & \rot{Feature} 
    & \rot{Actual} & \rot{Hypothetical} 
    & \rot{Step/Action} & \rot{Segment} & \rot{Episode} & \rot{Ent.Beh.}
    & \rot{Binary} & \rot{Discrete} & \rot{Continuous} 
    & \rot{Exclusive} & \rot{Augmenting} & Description & Year & Venue \\
% Your complex header structure goes here
\midrule
Okudo et al.~\cite{okudo2021subgoal} &  &  & \OK &  & \OK & \OKGREY & \OKGREY & \OKGREY & \OKGREY & \OKGREY &  & \OK &  & \OK &  &  & \OK &  & \OK &  &  &  & \OK & Subgoal Generation & 2021 & IEEE Access \\
Bobu et al.~\cite{bobu2021feature} &  &  & \OK &  &  & \OK &  & \OK & \OK &  &  & \OK &  & \OKGREY &  & \OK &  &  &  &  & \OK &  & \OK & Feature Traces & 2021 & HRI '21 \\
Chetouani et al.~\cite{chetouani2021interactive} &  & \OK &  &  & \OKGREY & \OKGREY & \OKGREY & \OKGREY & \OK &  & \OK &  &  & \OK & \OK &  &  &  &  &  &  & \OKGREY & \OKGREY & Teaching Signals: Demonstration & 2021 &  \\
Lee et al.~\cite{lee2021b} & \OK &  &  &  & \OK &  &  & \OK &  & \OK & \OK &  & \OK &  &  & \OK &  &  &  &  &  & \OK &  & Preferences & 2021 &  \\
Kwon et al.~\cite{kwon2021targeted} &  & \OK &  &  &  & \OKGREY &  & \OK & \OK &  & \OK &  & \OKGREY & \OKGREY &  & \OK &  &  & \OK &  &  & \OK &  & Expert Annotations & 2021 & ICML 2021 \\
Najar et al.~\cite{najar2021reinforcement} &  & \OK &  &  & \OK &  & \OKGREY & \OKGREY & \OK &  & \OK &  & \OKGREY & \OKGREY & \OK &  &  &  &  &  &  &  & \OK & Corrective Feedback & 2021 & Frontiers Robotics/AI \\
Cruz et al.~\cite{cruz2021interactive} &  & \OKGREY & \OK &  & \OK &  & \OK &  & \OKGREY & \OKGREY & \OKGREY & \OK & \OK & \OKGREY & \OK &  &  &  &  &  &  &  & \OK & Repairing via interactive explanations & 2021 & 2021 IEEE CoG \\
Punzi et al.~\cite{punzi2022imcasting} &  & \OK &  &  &  & \OK &  & \OK & \OK &  & \OK &  &  & \OKGREY &  &  & \OK &  & \OK &  &  & \OK &  & Nonverbal behaviour training & 2022 & IUI \\
Bai et al.~\cite{bai2022training} &  & \OK &  &  & \OK &  & \OK &  &  & \OK & \OK &  & \OKGREY & \OKGREY & \OK &  &  &  &  & \OK &  & \OK &  & Correcting one response out of a set reponses & 2022 & ArXiv \\
Sharma et al.~\cite{sharma2022correcting} &  & \OKGREY & \OK &  &  & \OK & \OK &  & \OK &  & \OKGREY & \OK & \OK &  &  &  & \OK &  & \OK &  &  &  & \OK & Corrections: adding constraints/specifying sub-goals & 2022 & Robotics: Science and Systems \\
Hsiung et al.~\cite{hsiung2022learning} & \OK & \OK &  &  & \OK &  & \OK &  & \OK &  & \OK &  & \OK &  & \OK &  &  &  & \OKGREY &  & \OK & \OK &  & Action Instructions and Evaluative Scanar Feedback & 2022 & ACM/IEEE HRI '22 \\
Cruz et al.~\cite{cruz2022broad} &  & \OK &  &  & \OK &  & \OK &  & \OK &  & \OK &  & \OK &  & \OK &  &  &  & \OK &  &  &  & \OK & Broad Persistent Advice & 2022 & IROS2022 Workshop \\
Crochepierre et al.~\cite{crochepierre2022interactive} & \OK & \OK &  &  & \OK &  & \OK & \OK & \OKGREY & \OK & \OKGREY & \OKGREY & \OK &  & \OK &  &  &  &  & \OK &  &  & \OK & Preference cat./pairs,  Expression Sugg. & 2022 & IJCAI 22 \\
Liu et al.~\cite{liu2022robot} &  & \OK &  &  & \OK &  & \OK &  &  & \OK & \OK &  & \OK & \OKGREY & \OKGREY & \OKGREY &  &  &  &  & \OK &  & \OK & Interventions via Shared Control & 2022 & RSS \\
Chi et al.~\cite{chi2022instruct} & \OK & \OK &  &  & \OK &  & \OKGREY & \OKGREY & \OK &  & \OK &  & \OK &  & \OK &  &  &  & \OK & \OK &  & \OK &  & Teaching by instruction or evaluation & 2022 & IEEE RO-MAN \\
Van Waveren et al.~\cite{van2022correct} &  & \OK & \OKGREY &  &  & \OK & \OK &  & \OKGREY & \OK & \OK & \OKGREY & \OK &  & \OK &  &  &  & \OK &  &  &  & \OK & Open-Language, Block-Based Non-Expert Feedb. & 2022 & ACM/IEEE HRI '22 \\
Mehta et al.~\cite{mehta2022unified} & \OK & \OK &  &  &  & \OK & \OK & \OK & \OK & \OK & \OK &  & \OK &  & \OK & \OK &  &  & \OK & \OK &  & \OK &  & Demonstrations, Corrections, and Preferences & 2022 & ACM THRI \\
Aronson et al.~\cite{aronson2022gaze} &  & \OK &  & \OK &  & \OK & \OKGREY &  & \OK &  & \OKGREY & \OK & \OK &  & \OK &  &  &  &  &  & \OK & \OK &  & Natural Eye Gaze and Joystick Input & 2022 & Robotics Science and Systems \\
Wang et al.~\cite{wang2022skill} & \OK &  &  &  & \OK &  & \OKGREY & \OKGREY & \OK & \OK & \OK &  & \OK &  &  & \OK &  &  & \OK &  &  &  & \OK & Skill preferences, mark. of positive trajectory seg. & 2022 & CoRL \\
Scherf et al.~\cite{scherf2022learning} &  & \OK &  &  & \OK & \OKGREY &  & \OK & \OK &  & \OK &  & \OK &  & \OK &  &  &  &  &  & \OK &  & \OK & Action Advice & 2022 & IEEE-RAS \\
Schrum et al.~\cite{schrum2022mind} &  & \OK &  &  &  & \OK & \OK &  &  & \OK & \OK &  & \OK &  & \OK &  &  &  & \OK &  &  &  & \OK & Corrective Actions & 2022 & HRI '22 \\
Trick et al.~\cite{trick2022interactive} &  & \OK &  &  &  & \OK & \OK &  & \OK &  & \OK &  & \OK &  & \OK &  &  &  &  & \OKGREY & \OKGREY &  & \OK & Multimodal Advice & 2022 & IEEE RA-L \\
Sheidlower et al.~\cite{sheidlower2022keeping} & \OK &  &  &  & \OK &  &  & \OKGREY & \OK &  & \OK &  & \OK &  & \OK &  &  &  & \OK &  &  &  & \OK &  & 2022 & IROS2022 \\
Lin et al.~\cite{lin2022inferring} &  & \OKGREY & \OK &  &  & \OK & \OK & \OKGREY & \OK & \OK & \OKGREY & \OK & \OKGREY & \OK &  & \OK &  & \OK & \OK & \OK &  & \OK &  & Natural language utterance (in context) & 2022 & ACL 2022 \\
Losey et al.~\cite{losey2022physical} &  & \OK &  &  &  & \OK & \OK &  &  & \OKGREY & \OK &  & \OK &  &  & \OK &  &  &  &  & \OK &  & \OK & Intentional Physical Corrections & 2022 & The Int. Journ. of Robotics Res. \\
Punzi et al.~\cite{punzi2022imcasting} & \OK &  &  &  &  & \OK &  & \OK &  & \OK & \OK &  & \OK &  &  &  & \OK &  &  &  &  & \OK &  & Voting & 2022 & IUI \\
Myers et al.~\cite{myers2022learning} & \OK &  &  &  & \OK &  &  & \OK &  & \OK & \OK &  & \OK &  &  &  & \OK &  &  & \OK &  & \OK &  & Rankings & 2022 & CoRL 2021 \\
Wilson-Small et al.~\cite{wilson2023drone} &  & \OK &  &  &  & \OK & \OK &  & \OK &  & \OKGREY & \OKGREY & \OK &  &  & \OKGREY &  &  &  &  & \OK &  & \OK & Drones-Physical Instructive Feedback & 2023 & HRI '23 \\
Zhang et al.~\cite{zhang2023self} & \OK &  &  & \OKGREY & \OK & \OK & \OK &  & \OK &  & \OK &  & \OK &  &  &  & \OKGREY &  &  & \OK & \OKGREY & \OK &  & Survey feedb., self-annot. via facial express./label val. & 2023 & ACM/IEEE HRI '23 \\
Zhang et al.~\cite{zhang2023dual} & \OK &  &  &  & \OK &  &  & \OK & \OK & \OK & \OK &  & \OK &  &  & \OK &  &  & \OK &  &  &  & \OK & Human Guidance via evaluation or prefernece & 2023 & CoRL \\
Thierauf et al.~\cite{thierauf2023instead} &  & \OK &  &  &  & \OK & \OK &  &  & \OK & \OK &  &  & \OK &  & \OKGREY & \OKGREY &  &  &  &  & \OKGREY & \OKGREY & Corrected Instructions & 2023 & ACM THRI \\
B{\i}y{\i}k et al.~\cite{biyik2023active} & \OK &  &  &  & \OK &  &  & \OK &  & \OK & \OK &  & \OK &  &  & \OK &  &  & \OK &  &  & \OK &  & Preference Queries & 2023 & CoRL \\
Zhang et al.~\cite{zhang2023good} &  &  & \OK &  & \OK &  &  & \OK & \OK &  &  & \OK & \OK &  & \OK &  &  &  & \OKGREY &  & \OKGREY &  & \OK & Object-in-view feedback & 2023 & UR 2023 \\
\end{tabular}%
    }
    \caption{Summary table of the literature survey. Surveyed work is classified according to the conceptual framework (\OK) Black checkmarks refer to exclusive attributes, (\OKGREY) Grey checkmarks indicate that feedback can have characteristics across different work, (\OKBLUE) Blue checkmarks  indicate necessary simultaneous presence of two attributes. We can classify existing types of feedback according to our dimensions.}
    \label{tab:surveyed_papers_dimensions}
    \end{table}
    \end{landscape}

\begin{landscape}
\begin{table}[]
\centering
\resizebox{1.25\textwidth}{!}{%\
\definecolor{blueish}{HTML}{F4F7F8}
\rowcolors{2}{white}{blueish}
\setlength{\tabcolsep}{1.5pt}
\renewcommand\arraystretch{1.20}
\begin{tabular}{r | *{4}{c} | *{2}{c} | *{2}{c} | *{2}{c} | *{2}{c} | *{2}{c} | *{4}{c} | *{3}{c} | *{2}{c} | *{1}{c} | c | c | c }
    & \multicolumn{4}{|c|}{\textcolor{DarkHumanColor}{\textbf{D1}}} 
    & \multicolumn{2}{c|}{\textcolor{DarkHumanColor}{\textbf{D2}}} 
    & \multicolumn{2}{c|}{\textcolor{DarkHumanColor}{\textbf{D3}}} 
    & \multicolumn{2}{c|}{\textcolor{DarkCombinedColor}{\textbf{D4}}} 
    & \multicolumn{2}{c|}{\textcolor{DarkCombinedColor}{\textbf{D5}}} 
    & \multicolumn{2}{c|}{\textcolor{DarkCombinedColor}{\textbf{D6}}}
    & \multicolumn{4}{c|}{\textcolor{DarkModelColor}{\textbf{D7}}}
    & \multicolumn{3}{c|}{\textcolor{DarkModelColor}{\textbf{D8}}}
    & \multicolumn{2}{c|}{\textcolor{DarkModelColor}{\textbf{D9}}}\\
    & \multicolumn{4}{|c|}{\textcolor{DarkHumanColor}{\textbf{Intent}}} 
    & \multicolumn{2}{c|}{\textcolor{DarkHumanColor}{\textbf{Expres.}}} 
    & \multicolumn{2}{c|}{\textcolor{DarkHumanColor}{\textbf{Engag.}}} 
    & \multicolumn{2}{c|}{\textcolor{DarkCombinedColor}{\textbf{Rela.}}} 
    & \multicolumn{2}{c|}{\textcolor{DarkCombinedColor}{\textbf{Content}}} 
    & \multicolumn{2}{c|}{\textcolor{DarkCombinedColor}{\textbf{Tgt.Act.}}} 
    & \multicolumn{4}{c|}{\textcolor{DarkModelColor}{\textbf{Tmp.Gra.}}} 
    & \multicolumn{3}{c|}{\textcolor{DarkModelColor}{\textbf{Choi.Set}}} 
    & \multicolumn{2}{c|}{\textcolor{DarkModelColor}{\textbf{Excls.}}} 
    \\
    \cmidrule{2-24}
    Publication 
    & \rot{Evaluate} & \rot{Instruct} & \rot{Describe} & \rot{None} 
    & \rot{Explicit} & \rot{Implicit} 
    & \rot{Proactive} & \rot{Reactive} 
    & \rot{Absolute} & \rot{Relative} 
    & \rot{Instance} & \rot{Feature} 
    & \rot{Actual} & \rot{Hypothetical} 
    & \rot{Step/Action} & \rot{Segment} & \rot{Episode} & \rot{Ent.Beh.}
    & \rot{Binary} & \rot{Discrete} & \rot{Continuous} 
    & \rot{Exclusive} & \rot{Augmenting} & Description & Year & Venue \\
% Your complex header structure goes here
\midrule
Newman et al.~\cite{newman2023towards} &  & \OK &  &  &  & \OK & \OK &  &  & \OK &  & \OK & \OK &  &  &  & \OK &  & \OK &  &  &  & \OK & Corrections & 2023 & HRI '23 \\
Torne et al.~\cite{torne2023breadcrumbs} & \OK &  &  &  & \OK &  &  & \OK &  & \OK & \OK &  & \OK &  &  & \OK &  &  &  & \OK &  &  & \OK & Binary Preferences & 2023 & ArXiv \\
Bignold et al.~\cite{bignold2023human} & \OK & \OK &  &  & \OK &  & \OK &  & \OK &  & \OK &  & \OK &  & \OK &  &  &  & \OK &  &  &  & \OK & (Eval.) Binary Advice, (Informative) Advice & 2023 & Neural Comp. and Appl. \\
Yu et al.~\cite{yu2023thumbs} & \OK &  &  &  & \OK &  &  & \OK & \OK &  & \OK &  & \OK &  &  & \OK &  &  &  & \OKGREY &  & \OKGREY & \OKGREY & Scalar Feedback & 2023 & IROS \\
Ying et al.~\cite{ying2023inferring} &  & \OK &  & \OK &  & \OKGREY & \OKGREY & \OKGREY & \OK &  & \OK & \OKGREY & \OK & \OKGREY & \OK &  &  &  & \OK &  &  & \OK &  & Instructions and Actions for Goal Inference & 2023 & AAAI 2023 Fall Symposium \\
Xu et al.~\cite{xu2023dexterous} &  & \OKGREY & \OK &  &  & \OK & \OK &  & \OK &  & \OKGREY & \OK &  & \OK & \OKGREY & \OK &  &  &  &  & \OKGREY &  & \OK & Visual Milestones/Transitions & 2023 & ICRA 2023 \\
Shek et al.~\cite{shek2023learning} &  & \OKGREY & \OK &  &  & \OK & \OK &  &  & \OK & \OKGREY & \OK &  & \OK & \OKGREY &  &  &  &  &  &  &  & \OK & Object Preferences & 2023 & ICRA 2022 \\
Lukowicz et al.~\cite{lukowicz2023sa} & \OK &  &  &  &  & \OK & \OK &  & \OK &  & \OK & \OKGREY & \OK & \OKGREY & \OK &  &  &  & \OK &  &  & \OKGREY &  & Reinforcement & 2023 & 
Frontiers in AI and App. \\
Fitzgerald et al.~\cite{fitzgerald2023inquire} & \OK & \OK &  &  & \OKGREY &  &  & \OK & \OKGREY & \OKGREY & \OK &  & \OKGREY & \OKGREY & \OK & \OK &  &  & \OK & \OK &  & \OK &  & Demos, Prefs., Corrections/Binary Feedb. & 2023 & CoRL 2023 \\
Liu et al.~\cite{liu2023interactive} &  & \OK & \OKGREY &  &  & \OK &  &  & \OKGREY & \OKGREY & \OK &  &  & \OK &  & \OK &  &  & \OK &  &  &  & \OK & Verbal correction & 2023 & CoRL 2023 Workshop \\
Yu et al.~\cite{yu2023language} & \OKGREY & \OK & \OKGREY &  &  & \OK & \OK &  & \OKGREY & \OKGREY & \OK & \OK & \OK &  &  & \OK & \OKGREY &  & \OK &  &  & \OKGREY &  & Language instructions/corrections & 2023 & ArXiv \\
Wang et al.~\cite{wang2023voyager} &  & \OK &  &  & \OK &  &  & \OK & \OK &  &  & \OK & \OK &  &  & \OK &  &  &  &  &  &  & \OK & Gameplay Interact. for intermediate goal states & 2023 & ArXiv \\
Wu et al.~\cite{wu2023fine} & \OK &  &  &  & \OK &  & \OK & \OKGREY & \OK &  & \OK & \OKGREY & \OK &  &  & \OK &  &  & \OK &  &  & \OK &  & Fine-grained annotations & 2023 & ArXiv \\
Candon et al.~\cite{candon2023nonverbal} &  &  &  & \OK &  & \OK & \OKGREY &  &  & \OKGREY & \OK &  & \OK &  &  & \OK &  &  &  &  & \OK & \OK &  & Nonverbal reactions & 2023 & AAMAS 2023 \\
Zhao et al.~\cite{zhao2023learning} &  & \OK &  &  & \OK &  &  & \OK & \OK &  & \OK &  & \OK &  & \OK &  &  &  & \OK &  &  &  & \OK & Actions & 2023 & CoRL 2023 \\
Wang et al.~\cite{wang2023voyager} & \OK &  &  &  & \OK &  &  & \OK & \OK &  &  & \OK & \OK &  &  & \OK &  &  &  &  &  &  & \OK & Visual Inspection of Intermediate Results & 2023 & TMLR \\
Mehta et al.~\cite{mehta2024strol} &  & \OK &  &  &  & \OK & \OK &  &  & \OKGREY & \OK &  & \OK &  & \OK &  &  &  &  &  & \OK & \OK &  & Corrections & 2024 & IEEE RA-L \\
Yow et al.~\cite{yow2024extract} &  & \OK &  &  &  & \OK & \OK &  &  & \OKGREY & \OKGREY & \OKGREY & \OK &  &  & \OK &  &  &  &  & \OKGREY &  & \OK & Language correction & 2024 & ArXiv \\
\end{tabular}%
    }
    \caption{Summary table of the literature survey. Surveyed work is classified according to the conceptual framework (\OK) Black checkmarks refer to exclusive attributes, (\OKGREY) Grey checkmarks indicate that feedback can have characteristics across different work, (\OKBLUE) Blue checkmarks  indicate necessary simultaneous presence of two attributes. We can classify existing types of feedback according to our dimensions.}
    \label{tab:surveyed_papers_dimensions}
    \end{table}
    \end{landscape}

% User Interface Tables

\section{Feedback User Interfaces}
In this appendix, we provide a collection of proposed user interfaces for the collection of feedback.

\subsection{User Interface: Language}
The following works mention language-based user interactions for feedback collection:
\begin{table}[h]
\centering
\definecolor{blueish}{HTML}{F4F7F8}
\rowcolors{2}{white}{blueish}
\setlength{\tabcolsep}{1.5pt}
\renewcommand\arraystretch{1.20}
\begin{tabular}{r | p{10cm}}
\toprule
\textbf{Publication} & \textbf{Description} \\
\midrule
Kuhlmann et al.\cite{kuhlmann2004guiding} & State-dependent if-then-else rules to guide agent \\
Tenorio-Gonzalez et al.\cite{tenorio2010dynamic} & Using language commands (e.g., action to take, evaluation, confirmation of reached goal state) \\
Krening et al.\cite{krening2018newtonian} & Language instructions mapped to actions in a grid world domain \\
Krening et al.\cite{krening2018interaction} & Language instructions/binary critique for grid world domain \\
De Winter et al.~\cite{de2019accelerating} & Verbally telling constraints to the system in natural language, then translated into first-order logic; input can be temporal action sequences \\
Bignold et al.~\cite{bignold2021persistent} & When giving a rating for a state, the user can also formulate a general rule; multiple rules are processed with probabilistic policy reuse \\
Sumers et al.\cite{sumers2021learning} & Simple game-like object collecting example; linguistic feedback is grouped into three categories and produced for observed episodes \\
Okudo et al.\cite{okudo2021subgoal} & Generate subgoals for a more complex overall behavior in a grid-world domain (points to reach via a UI drawing style); for a robot (verbal description of subgoals to reach) \\
Millan-Arias et al.~\cite{millan2021robust} & No interactive experiment discussed; using simulated oracles for cartpole language-based actions \\
Chi et al.\cite{chi2022instruct} & Humans are queried to either teach via instruction/action advice or evaluation on a five-point scale for predicted action; simplified medical care setting \\
Van Waveren et al.\cite{van2022correct} & From natural language and "block-based" (i.e., programming language) feedback, infer action refinement, alternative actions, and forbidden actions \\
Bai et al.\cite{bai2022training} & Chat interface; ranking of two entries with confidence (one or the other is better) \\
Trick et al.\cite{trick2022interactive} & Giving action advice (encoding by mentioning an object for the robot to grip) via language (with keyword spotting) or by interpreting pointing gestures; robotics grasping task \\
Lin et al.~\cite{lin2022inferring} & Flight booking use case; users can give natural language feedback to an agent; reward function is inferred from free-form feedback \\
Ying et al.\cite{ying2023inferring} & First, humans utter an instruction, then perform a series of actions; goal is inferred from both; grid world domain \\
Lukowicz et al.\cite{lukowicz2023sa} & Speech and visual cues to penalize/reinforce (anticipated) behavior \\
Liu et al.\cite{liu2023interactive} & User interrupts robot behavior to utter verbal corrections \\
\bottomrule
\end{tabular}
\caption{User Interface: Language}
\end{table}

\begin{table}[h]
\centering
\definecolor{blueish}{HTML}{F4F7F8}
\rowcolors{2}{white}{blueish}
\setlength{\tabcolsep}{1.5pt}
\renewcommand\arraystretch{1.20}
\begin{tabular}{r | p{11cm}}
\toprule
\textbf{Publication} & \textbf{Description} \\
\midrule
Yu et al.\cite{yu2023language} & Robotics domain; language is translated into rewards; demonstrated as a chat-like interface where humans can give multiple text feedback utterances, e.g., comments on quality, what to do instead/focus on \\
Wu et al.\cite{wu2023fine} & Feedback for text; interactive text annotations on sentence-granularity, separate annotations for missing information, relevance, and factuality \\
Yow et al.~\cite{yow2024extract} & Robot object grasping scenarios; correct trajectories of robots in space via language corrections (e.g., move closer to, go down, etc.) \\
\bottomrule
\end{tabular}
\caption{User Interface: Language (continuation)}
\end{table}

\subsection{User Interface: Visual}
The following works mention visual user interactions for feedback collection:
\begin{table}[h]
\centering
\definecolor{blueish}{HTML}{F4F7F8}
\rowcolors{2}{white}{blueish}
\setlength{\tabcolsep}{1.5pt}
\renewcommand\arraystretch{1.20}
\begin{tabular}{r | p{11cm}}
\toprule
\textbf{Publication} & \textbf{Description} \\
\midrule
Veeriah et al.~\cite{veeriah2016face} & Using facial features as input in the observations and human-generated explicit rewards. Facial values are used to optimize rewards \\
Torabi et al.~\cite{Torabi2018} & Instead of giving demonstrations via an agent-compatible interface, agent actions are observed and inferred with an inverse dynamics model \\
Arakawa et al.\cite{arakawa2018dqn} & Using facial gestures to infer reward/preferences \\
Li et al.\cite{li2020facial} & Simultaneous binary feedback and facial feedback (classified as good and bad) \\
Zhang et al.\cite{zhang2020atari} & Eye-tracking for Atari games; gaze can be used as additional information (e.g., attention guidance) during human play \\
Cui et al.\cite{cui2021empathic} & Mapping facial gestures (without clear intention) to a reward distribution; using calibration phase to learn mapping \\
Okudo et al.\cite{okudo2021subgoal} & Generate subgoals for a more complex overall behavior in a grid-world domain (points to reach via a UI drawing style); for a robot (verbal description of subgoals to reach) \\
Aronson et al.\cite{aronson2022gaze} & Combine joystick inputs with gaze tracking in a robot teleoperation setting (picking objects) to identify goals \\
Trick et al.\cite{trick2022interactive} & Giving action advice (encoding by mentioning an object for the robot to grip) via language (with keyword spotting) or by interpreting pointing gestures; robotics grasping task \\
Zhang et al.\cite{zhang2023self} & Photo-taking task (optimal frame); explicit evaluative survey feedback, followed by a review of the feedback with both facial tracking and interaction changing of values \\
Xu et al.\cite{xu2023dexterous} & Users provide visual milestones (successful subgoal states as image observations) and move the robot to new positions \\
Candon et al.~\cite{candon2023nonverbal} & Measure multiple features of nonverbal human behavior during co-execution of an autonomous agent in a video game setting \\
\bottomrule
\end{tabular}
\caption{User Interface: Visual}
\end{table}

\newpage

\subsection{User Interface: Buttons/Controls}
The following works mention visual buttons/other types of controls for feedback collection:
\begin{table}[h]
\centering
\definecolor{blueish}{HTML}{F4F7F8}
\rowcolors{2}{white}{blueish}
\setlength{\tabcolsep}{1.5pt}
\renewcommand\arraystretch{1.20}
\begin{tabular}{r | p{11cm}}
\toprule
\textbf{Publication} & \textbf{Description} \\
\midrule
Thomaz et al.\cite{thomaz2005real, thomaz2007asymmetric} & Sophie's Kitchen (interactive interface) with on-screen scales \\
Judah et al.\cite{judah2010reinforcement} & Critique session where the end-user can analyze trajectories of the current policy and label actions as good or bad \\
Cakmak et al.\cite{cakmak2011human} & Four good and four bad examples for hand-over; robot controlled by human via GUI sliders \\
Taylor et al.\cite{taylor2011integrating} & Interactive football simulation with multiple players; human provides "action/pass" command at appropriate times \\
Knox et al.\cite{knox2012reinforcement} & Buttons for positive/negative feedback \\
Li et al.\cite{li2013using} & Button presses to give positive/negative feedback for an action \\
Loftin et al.\cite{loftin2014learning} & Investigating human feedback strategies \\
Amir et al.\cite{Amir2016} & In student-teacher learning, the teacher can give action advice\\
Raza et al.\cite{raza2016reward} & Action advice for reward shaping instead of evaluative reward, Gridworld domain, small user study (2 participants) \\
Loftin et al.\cite{loftin2016learning} & Simple video game environment with buttons for feedback \\
El Asri et al.\cite{elAsri2016} & Users rate trajectories (dialogs) with scores from 1 to 10 \\
Sørensen et al.~\cite{sorensen2016breeding} & Super Mario; select most promising behavior to guide evolution \\
Fachantidis et al.\cite{fachantidis2017learning} & Teaching task with finite advice budget; teacher can choose advice as action or no advice \\
MacGlashan et al.\cite{macglashan2017interactive} & Slider with simple environment rendering \\
Christiano et al.~\cite{christiano2017deep} &  Two videos side-by-side with feedback buttons \\
Li et al.\cite{li2018interactive} & Grid world domain; demonstration, then evaluative feedback during learning \\
Basu et al.\cite{basu2018learning} & Using preferences between observed episodes induced by different reward functions and specifying different salient features \\
Lončarević et al.\cite{lonvcarevic2018user} & Robotics; basketball throw, users can rate throws on different scales (ordinal scale: further than goal, hit, closer than goal, etc.) \\
Ibarz et al.\cite{ibarz2018reward} & Atari games; two phases: initial demo generation, then iterative updates with preference reward model \\
Warnell et al.\cite{warnell2018deep} & Atari games; Button-based UI for feedback \\
Akkaladevi et al.\cite{akkaladevi2019towards} & Robotics co-assembly task; feedback via GUI, users can choose the best action from a list of possible actions \\
Wang et al.\cite{Wang2019interactive} & Train a supervised policy based on human-generated data to support learning in interactive RL; feedback not specified \\
Frazier et al.\cite{frazier2019improving} & Minecraft domain, goal-seeking task; simulated action advice (with low frequency) from oracle, no actual user studies \\
\bottomrule
\end{tabular}
\caption{User Interface: Button/Controls}
\end{table}

\textcolor{white}{cont.}

\begin{table}[h]
\centering
\definecolor{blueish}{HTML}{F4F7F8}
\rowcolors{2}{white}{blueish}
\setlength{\tabcolsep}{1.5pt}
\renewcommand\arraystretch{1.20}
\begin{tabular}{r | p{11cm}}
\toprule
\textbf{Publication} & \textbf{Description} \\
\midrule
Arzate Cruz et al.\cite{arzate2020survey} & Provide agent with action they believe is optimal + Scalar-valued rating \\
Wilde et al.\cite{wilde2020improving} & Visual interface showing 2D trajectories; users asked for preferred trajectories, list of violations and speed, refinement of annotated zones \\
Arzate Cruz et al.\cite{arzate2020mariomix} & Super Mario; segmented level view \\
Koert et al.\cite{koert2020multi} & Multiple ways for human interaction with robot learning systems; initial reward subgoals, evaluative feedback or action advice after performed actions, state manipulation helping the robot \\
Arzate Cruz et al.\cite{arzate2020survey} & Gridworld; give +1/-1 rating for a single action \\
Moreira et al.\cite{moreira2020deep} & User interface showing the available actions with associated predicted probabilities; users must select best actions during a consecutive 100-step phase \\
Arzate Cruz et al.\cite{arzate2020survey} & The human user specifies the goal object(s) in the environment \\
Raza et al.\cite{raza2020human} & Human teacher "relabels" the selection actions, i.e., chooses best action but agents act according to learned policy \\
Menner et al.\cite{menner2020using} & Therapists rate gait of subjects on a scale from 1-3 in four categories to train a reward model in a supervised fashion \\
Wilde et al.\cite{wilde2020improving} & GUI to specify zones indicating traffic rules in a 2D warehouse; zones represent avoidance, speed limits, desired roads, etc. \\
Law et al.\cite{law2021hammers} & Redesign agents in a loop based on performance \\
Wilde et al.\cite{wilde2021learning} & Combine preference-based and scalar reward by using a slider to show preference on a spectrum \\
Hsiung et al.\cite{hsiung2022learning} & Users can choose to use evaluative or action advice to teach robot agents \\
Cruz et al.\cite{cruz2022broad} & Humans can give "action advice" for steps if they want; do not give for each step; discusses efficient use of samples \\
Crochepierre et al.\cite{crochepierre2022interactive} & Symbolic regression domain; first group expressions into three groups, then generate preferences for pairs; optionally create "corrected" expression \\
Chi et al.\cite{chi2022instruct} & Humans are queried to either teach via instruction/action advice or evaluation on a five-point scale for predicted action; simplified medical care setting \\
Wang et al.\cite{wang2022skill} & Two phases: label trajectories from an offline robotics dataset into preferable and non-preferable; label segments of trajectories having made more progress towards the goal \\
Bai et al.\cite{bai2022training} & Chat interface; ranking of two entries with confidence (one or the other is better) \\
Scherf et al.\cite{scherf2022learning} & During action advice giving, users are recorded and behavioral cues like response time and facial expressions are recorded \\
Sheidlower et al.\cite{sheidlower2022keeping} & Binary feedback with "toggle" approach (i.e., states are rewarded as long as the feedback direction is not actively changed) \\
Punzi et al.\cite{punzi2022imcasting} & Immersive VR environment; voting for the best actor out of a pool of actors via button presses \\
\bottomrule
\end{tabular}
\caption{User Interface: Button/Controls (continuation)}
\end{table}

\textcolor{white}{cont.}

\begin{table}[h]
\centering
\definecolor{blueish}{HTML}{F4F7F8}
\rowcolors{2}{white}{blueish}
\setlength{\tabcolsep}{1.5pt}
\renewcommand\arraystretch{1.20}
\begin{tabular}{r | p{11cm}}
\toprule
\textbf{Publication} & \textbf{Description} \\
\midrule
Myers et al.\cite{myers2022learning} & Robotics; Lunar Lander; rankings over observed trajectories \\
Zhang et al.\cite{zhang2023dual} & Robotics/simulated robotics interfaces; robotics task \\
Bıyık et al.\cite{biyik2023active} & Preferences between segments in driving, tossing, and robotics tasks \\
Torne et al.\cite{torne2023breadcrumbs} & Robotics domain; use binary preferences to guide exploration; asynchronous feedback \\
Bignold et al.\cite{bignold2023human} & Cartpole; view steps and actions, possibility to overwrite/rate given action if disagreement \\
Yu et al.\cite{yu2023thumbs} & Showing that scalar feedback can be better than binary feedback if considering the labelers \\
Ying et al.\cite{ying2023inferring} & First, humans utter an instruction, then perform a series of actions; goal is inferred from both; grid world domain \\
Fitzgerald et al.\cite{fitzgerald2023inquire} & Rare example of a multi-type feedback approach; studies how to query; no actual implementation with human feedback, just with oracles \\
Zhao et al.~\cite{zhao2023learning} & Selection of an object according to a personal reward function and an overall reward function; users select one of several objects assigned a weight and color \\
Li et al.\cite{li2021learning} & Robotics application; applies corrections to robot movements to avoid goal misspecification; effect from multiple corrections is accumulated \\
\bottomrule
\end{tabular}
\caption{User Interface: Button/Controls (continuation)}
\end{table}

\clearpage

\subsection{User Interface: Physical}
The following works mention using physical user interfaces for feedback collection, especially relevant in the domain of human-robot interaction:
\begin{table}[tbh]
\centering
\definecolor{blueish}{HTML}{F4F7F8}
\rowcolors{2}{white}{blueish}
\setlength{\tabcolsep}{1.5pt}
\renewcommand\arraystretch{1.20}
\begin{tabular}{r | p{11cm}}
\toprule
\textbf{Publication} & \textbf{Description} \\
\midrule
Li et al.\cite{li2021roial} & Gait/walking support; requests ordinal feedback (good, very good, etc.) and preference compared to previous setup \\
Bobu et al.~\cite{bobu2021feature} & Sequence of states achieved by human manipulation of robot; representing decreasing value of a requested feature (e.g., distance to an object) \\
Liu et al.\cite{liu2022robot} & Human operations can intervene in robot behavior based on tele-operation equipment \\
Mehta et al.\cite{mehta2022unified} & Robotics application; multi-stage approach starting with demonstrations, then corrections, and finally preferences \\
Aronson et al.\cite{aronson2022gaze} & Combine joystick inputs with gaze tracking in a robot teleoperation setting (picking objects) to identify goals \\
Schrum et al.~\cite{schrum2022mind} & Users can choose to provide correction to race car driving towards a goal; learned to be interpreted as action labels \\
Losey et al.\cite{losey2022physical} & In a human-robot co-working scenario, intentional physical corrections to the robot actions (like pushing/pulling, etc.) can be interpreted as feedback, changing the current trajectory online \\
Wilson-Small et al.\cite{wilson2023drone} & Humans can guide drone behavior by physical touch/interactions with the drones \\
Shek et al.~\cite{shek2023learning} & Users can correct the position of a specific object that the robot handles with its actuators (so the robot arm can be repositioned, but just indirectly by adjusting the robot position) \\
Mehta et al.~\cite{mehta2024strol} & Robotics; correction for moving robot trajectories via physical guidance \\
\bottomrule
\end{tabular}
\caption{User Interface: Physical}
\end{table}

\clearpage

\subsection{User Interface: Programming}
The following works mention using programming-like interfaces (i.e. directly using programming, formal or domain-specific languages) for feedback collection:
\begin{table}[h]
\centering
\definecolor{blueish}{HTML}{F4F7F8}
\rowcolors{2}{white}{blueish}
\setlength{\tabcolsep}{1.5pt}
\renewcommand\arraystretch{1.20}
\begin{tabular}{r | p{11cm}}
\toprule
\textbf{Publication} & \textbf{Description} \\
Maclin et al.~\cite{maclin1996creating} & Give instructions/rules in inductional logic/programming language (PROLOG) \\
Hayes-Roth et al.\cite{hayes2013advice} & General constraints as LISP expressions to specify rules, heuristics, etc. \\
Yeh et al.\cite{yeh2018bridging} & Minecraft; humans can give template-based advice, then generates more fine-grained reward via a buffer \\
Van Waveren et al.\cite{van2022correct} & From natural language and "block-based" (i.e., programming language) feedback, infer action refinement, alternative actions, and forbidden actions \\
Lin et al.~\cite{lin2022inferring} & Flight booking use case; users can give natural language feedback to an agent; reward function is inferred from free-form feedback \\
\midrule
\bottomrule
\end{tabular}
\caption{User Interface: Other}
\end{table}

\newpage

\subsection{User Interface: Other}
The following works mention using other, miscellaneous types of user input for feedback collection:
\begin{table}[h]
\centering
\definecolor{blueish}{HTML}{F4F7F8}
\rowcolors{2}{white}{blueish}
\setlength{\tabcolsep}{1.5pt}
\renewcommand\arraystretch{1.20}
\begin{tabular}{r | p{11cm}}
\toprule
\textbf{Publication} & \textbf{Description} \\
\midrule
Ferrez et al.\cite{ferrez2005you} & Using EEG error potentials to measure error in command interpretation (if the command was correctly understood by the interface) \\
Cakmak et al.\cite{cakmak2011human} & Preferences over hand-overs executed by a robot (hand-over of objects) \\
Odom et al.\cite{odom2015active} & Group features and request feedback for entire groups of features \\
Subramanian et al.\cite{subramanian2016exploration} & Demo generation interface \\
Branavan et al.~\cite{branavan2009reinforcement} & Provide action plans for reward function evaluation \\
Kim et al.\cite{kim2017intrinsic} & Measure EEG for a robot imitating human gestures \\
Lin et al.\cite{lin2017explore} & Only simulated; noisy oracle for action advice \\
Losey et al.\cite{losey2018including} & Not implemented, but potentially drawing of new trajectory/leading a robot \\
Cruz et al.\cite{cruz2018multi} & Simulated robotics environment \\
Millán et al.~\cite{millan2019human} & Simulated user giving "advice", agreeing/disagreeing with a particular action \\
Zhang et al.\cite{zhang2019leveraging} & Using human gaze as an additional feedback signal \\
Faulkner et al.\cite{faulkner2020interactive} & Robot learning scenario; same value range as ground-truth reward function, i.e., scalar reward function, given via oracle \\
Najar et al.\cite{najar2021reinforcement} & Context-dependent with respect to the current state of the task; examples: evaluative feedback, corrective feedback, guidance, contextual instructions \\
Najar et al.\cite{najar2021reinforcement} & Communicated non-interactively prior to learning; not context-dependent, e.g., general constraints or general instructions \\
Najar et al.\cite{najar2021reinforcement} & Feedback informs about past actions \\
Xu et al.\cite{xu2021accelerating} & Measure error potentials from EEG to detect suboptimal behavior \\
Najar et al.\cite{najar2021reinforcement} & Guidance informs about future actions \\
Najar et al.\cite{najar2021reinforcement} & Action advice given before an action; differentiates between high-level and low-level instructions (e.g., a single action or a goal similar to options) \\
Chetouani et al.\cite{chetouani2021interactive} & Not specified, but assumed to be through the "social channel" \\
Chetouani et al.\cite{chetouani2021interactive} & Through the "task channel," e.g., via kinesthetic, teleoperation, and observation \\
Harnack et al.\cite{harnack2023quantifying} & Fully simulated feedback; binary signal based on whether state has improved based on action or not \\
Thierauf et al.\cite{thierauf2023instead} & Correction of natural language instructions during training and inference \\
Newman et al.~\cite{newman2023towards} & Correcting the goal position of household items in a (simulated) dishwasher \\
\bottomrule
\end{tabular}
\caption{User Interface: Other}
\end{table}

\end{document}